\documentclass[10pt,twocolumn,letterpaper]{article}

\usepackage{iccv}
\usepackage{times}
\usepackage{epsfig}
\usepackage{graphicx}
\usepackage{amsmath}
\usepackage{amssymb}
\usepackage{xcolor}
\usepackage{capt-of}
\usepackage{multirow}
\usepackage{booktabs}
\usepackage{tabularx}
\usepackage{rotating}
\usepackage[accsupp]{axessibility}  

\usepackage[pagebackref=true,breaklinks=true,colorlinks,letterpaper=true, bookmarks=false]{hyperref}

\iccvfinalcopy 

\newbox\jsavebox
\newcommand{\jsubfig}[2]{%
	\sbox\jsavebox{#1}%
	\parbox[t]{\wd\jsavebox}{\centering\usebox\jsavebox\\#2}%
	}

\newcommand{\esc}[1]{{\color{red}[\textbf{ES:} #1]}}

\newcommand{\hec}[1]{{\color{teal}[\textbf{HE:} #1]}}
\newcommand{\phc}[1]{{\color{violet}[\textbf{PH:} #1]}}

\newcommand{\rev}[1]{{\color{red}#1}}

\long\def\ignorethis#1{}

\usepackage[T1]{fontenc} 
\usepackage{iftex} 

\ifLuaTeX
\protected\def\pdfmapline {\pdfextension mapline }
\fi

\newcommand{\whitetxt}[1]{{\color{white}#1}\normalfont}

\pdfmapline{+delphine < Delphine.ttf <T1-WGL4.enc}

\begin{document}

\title{Vox-E: Text-guided Voxel Editing of 3D Objects }

\author{
Etai Sella$^{1}$ \ \ \
Gal Fiebelman$^{1}$ \ \ \
Peter Hedman$^{2}$ \ \ \
 Hadar Averbuch-Elor$^{1}$ 
\\[2mm]
\vspace{1em}
$^1$Tel Aviv University \ \ \
$^2$Google Research
}


\twocolumn[{%
	\renewcommand\twocolumn[1][]{#1}%
	\maketitle
 \vspace{-25pt}
	\begin{center}
        \setlength{\tabcolsep}{2pt}
        \begin{tabular}{cc ccc cc cc ccc}
        \vspace{-10pt}
        \includegraphics[width=0.12\textwidth,trim={1.35cm 1.35cm 1.35cm 0cm},clip]{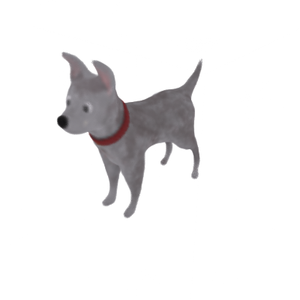} &
        { } &
        \includegraphics[width=0.12\textwidth,trim={1.35cm 1.35cm 1.35cm 0cm},clip]{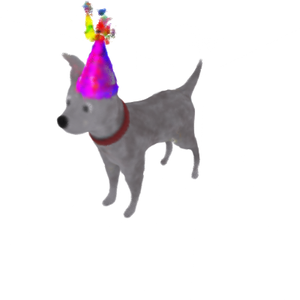} & 
        \includegraphics[width=0.12\textwidth,trim={1.35cm 1.35cm 1.35cm 0cm},clip]{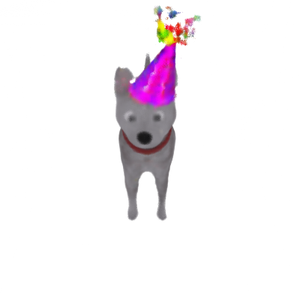} & 
        \includegraphics[width=0.12\textwidth,trim={1.35cm 1.35cm 1.35cm 0cm},clip]{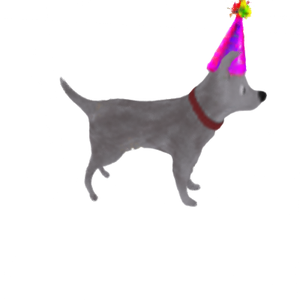} &
        {  } &

        { } &
        \includegraphics[width=0.12\textwidth,trim={1.35cm 1.35cm 1.35cm 0cm},clip]{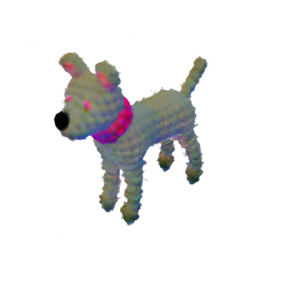} & 
        \includegraphics[width=0.12\textwidth,trim={1.35cm 1.35cm 1.35cm 0cm},clip]{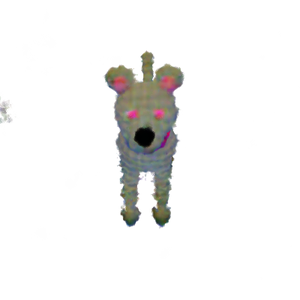} & 
        \includegraphics[width=0.12\textwidth,trim={1.35cm 1.35cm 1.35cm 0cm},clip]{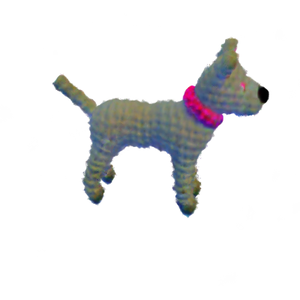} \\ 
        %
        %
        \includegraphics[width=0.12\textwidth,trim={1.35cm 1.35cm 1.35cm 0cm},clip]{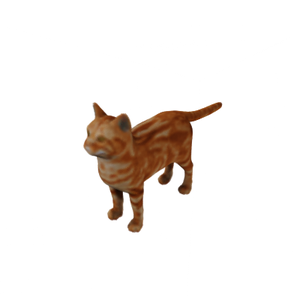} &
        { } &
        \includegraphics[width=0.12\textwidth,trim={1.35cm 1.35cm 1.35cm 0cm},clip]{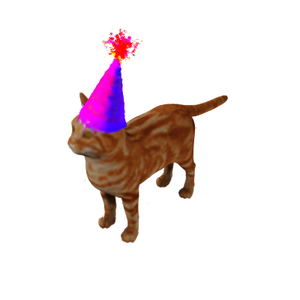} & 
        \includegraphics[width=0.12\textwidth,trim={1.35cm 1.35cm 1.35cm 0cm},clip]{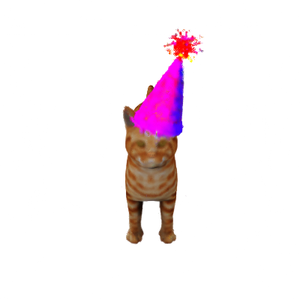} & 
        \includegraphics[width=0.12\textwidth,trim={1.35cm 1.35cm 1.35cm 0cm},clip]{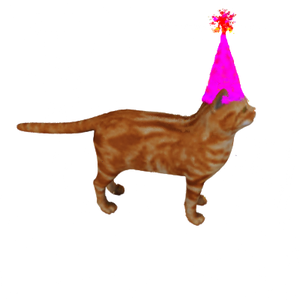} &
        { } &

        { } &
        \includegraphics[width=0.12\textwidth,trim={1.35cm 1.35cm 1.35cm 0cm},clip]{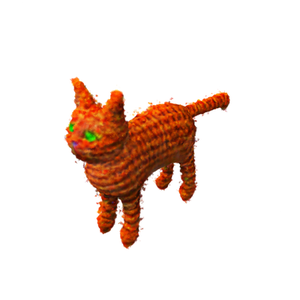} & 
        \includegraphics[width=0.12\textwidth,trim={0.75cm 1.35cm 1.35cm 0cm},clip]{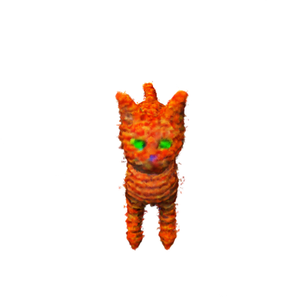} & 
        \includegraphics[width=0.12\textwidth,trim={0.375cm 1.35cm 1.35cm 0cm},clip]{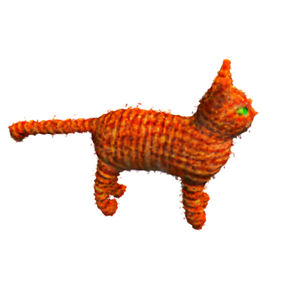} \\ 
        \includegraphics[width=0.12\textwidth,trim={0.6cm 1.35cm 0.525cm 0cm},clip]{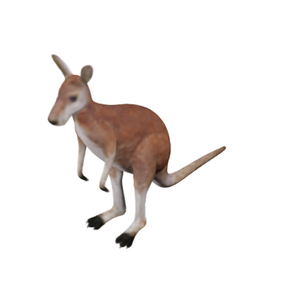} &
        { } &
        \includegraphics[width=0.12\textwidth,trim={0.6cm 1.35cm 0.525cm 0cm},clip]{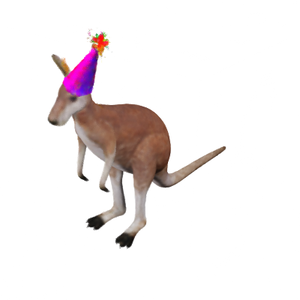} & 
        \includegraphics[width=0.12\textwidth,trim={0.6cm 1.35cm 0.525cm 0cm},clip]{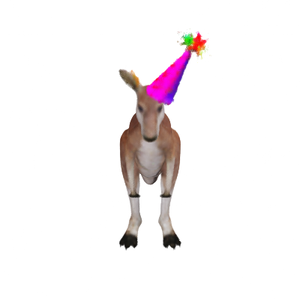} & 
        \includegraphics[width=0.12\textwidth,trim={0.6cm 1.35cm 0.525cm 0cm},clip]{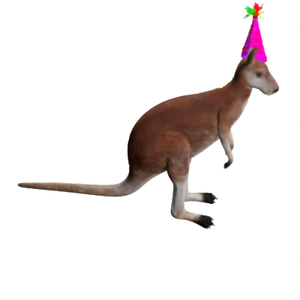} &
        { } &

        { } &
        \includegraphics[width=0.12\textwidth,trim={0.6cm 1.35cm 0.525cm 0cm},clip]{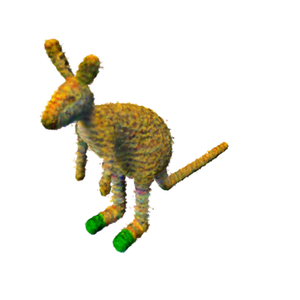} & 
        \includegraphics[width=0.12\textwidth,trim={0.6cm 1.35cm 0.525cm 0cm},clip]{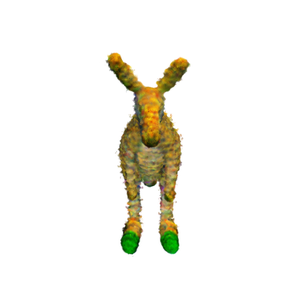} & 
        \includegraphics[width=0.12\textwidth,trim={0.6cm 1.35cm 0.525cm 0cm},clip]{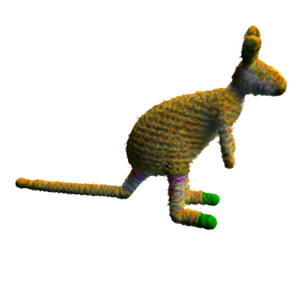} \\
        Original
        && \multicolumn{3}{c}{``A $\left<object\right>$ \emph{with a birthday hat}''} &
        && \multicolumn{3}{c}{``A \emph{yarn doll} of a $\left<object\right>$''} \\
    \end{tabular}
    \vspace{1mm}
    \captionof{figure}{Given multiview images of an object (left), our technique generates volumetric edits from target text prompts, allowing for significant geometric and appearance changes, while faithfully preserving the input object.  
    The objects can be edited either \emph{locally} (center) or \emph{globally} (right), depending on the nature of the user-provided text prompt. 
    }
    \label{fig:teaser}
	\end{center}
}]

\begin{abstract}
Large scale text-guided diffusion models have garnered significant attention due to their ability to synthesize diverse images that convey complex visual concepts. 
This generative power has more recently been leveraged to perform text-to-3D synthesis. 
In this work, we present a technique that harnesses the power of latent diffusion models for editing existing 3D objects.
Our method takes oriented 2D images of a 3D object as input and learns a grid-based volumetric representation of it.
To guide the volumetric representation to conform to a target text prompt, we follow unconditional text-to-3D methods and optimize a Score Distillation Sampling (SDS) loss. 
However, we observe that combining this diffusion-guided loss with an image-based regularization loss that encourages the representation not to deviate too strongly from the input object
is challenging, as it requires achieving two conflicting goals while viewing only structure-and-appearance coupled 2D projections. 
Thus, we introduce a novel volumetric regularization loss that operates directly in 3D space, utilizing the explicit nature of our 3D representation to enforce correlation between the global structure of the original and edited object. 
Furthermore, we present a technique that optimizes cross-attention volumetric grids to refine the spatial extent of the edits.
Extensive experiments and comparisons demonstrate the effectiveness of our approach in creating a myriad of edits  which cannot be achieved by prior works\footnote{Our code can be reached through our project page at \url{http://vox-e.github.io/}}.

\ignorethis{
However, the problem of performing text-driven edits of existing 3D objects with these powerful models  poses new challenges \hec{what challenges?} \phc{preserve the identity of the original object? taking the diffusion model out from its training distribution?} and has largely been unexplored. In this work we explore this task and offer a novel method for performing both local and global edits on 3D objects given textual descriptions by leveraging latent diffusion models as guidance. 

and edits this volumetric representation using a textual description of the edited object as guidance. 
We base our method on two core components: a generative component and a regularizing component. 

}   
\end{abstract}

\section{Introduction}

Creating and editing 3D models is a cumbersome task. While template models are readily available from online databases, tailoring one to a specific artistic vision often requires extensive knowledge of specialized 3D editing software.
In recent years, neural field-based representations (e.g., NeRF~\cite{mildenhall2021nerf}) demonstrated expressive power in faithfully capturing fine details, while offering effective optimization schemes through differentiable rendering. Their applicability has recently expanded also for a variety of editing tasks. However, research in this area has mostly focused on either appearance-only manipulations, which change the object's texture~\cite{xiang2021neutex,yang2022neumesh} and style~\cite{zhang2022arf,wang2022nerf}, or geometric editing via correspondences with an explicit mesh representation~\cite{garbin2022voltemorph,yuan2022nerf,xu2022deforming}---linking these representations to the rich literature on mesh deformations~\cite{igarashi2005rigid,sorkine2007rigid}. Unfortunately, these methods still require placing user-defined control points on the explicit mesh representation, and cannot allow for adding new structures or significantly adjusting the geometry of the object.

In this work, we are interested in enabling more flexible and localized object edits, guided only by textual prompts, which can be expressed through \emph{both} appearance and geometry modifications. To do so, we leverage the incredible competence of pretrained 2D diffusion models in editing images to conform with target textual descriptions. We carefully apply a \emph{score distillation} loss, as recently proposed in the unconditional text-driven 3D generation setting~\cite{poole2022dreamfusion}. Our key idea is to regularize the optimization in 3D space. We achieve this by coupling two volumetric fields, providing the system with more freedom to comply with the text guidance, on the one hand, while preserving the input structure, on the other hand. 

\ignorethis{Rather than using neural fields, we base our method on \emph{lighter} ReLU Fields~\cite{karnewar2022relu} which do not require any neural networks and instead model the scene as a voxel grid where each voxel contains learned features. }

Rather than using neural fields, we base our method on \emph{lighter} voxel-based representations 
which learn scene features over a sparse voxel grid.
This explicit grid structure not only allows for faster reconstruction and rendering times, but also for achieving a tight \emph{volumetric} coupling between volumetric fields representing the 3D object before and after applying the desired edit using a novel \emph{volumetric correlation loss} over the density features. 
%
%
To further refine the spatial extent of the edits, we utilize 2D cross-attention maps which roughly capture regions associated with the target edit, and lift them to volumetric grids. This approach is built on the premise that, while independent 2D internal features of generative models can be noisy, unifying them into a single 3D representation allows for better distilling the semantic knowledge.  
We then use these 3D cross-attention grids as a signal for a binary volumetric segmentation algorithm that splits the reconstructed volume into edited and non-edited regions, allowing for merging the features of the volumetric grids to better preserve regions that should not be affected by the textual edit.

Our approach, coined \emph{Vox-E}, provides an intuitive voxel editing interface, where the user only provides a simple target text prompt (see Figure~\ref{fig:teaser}). 
We compare our method to existing 3D object editing techniques, and demonstrate that our approach can facilitate local and global edits involving appearance and geometry changes over a variety of objects and text prompts, which are extremely challenging for current methods.

Explicitly stated, our contributions are:
\begin{itemize}
    \item A coupled volumetric representation tied using 3D regularization, allowing for editing 3D objects using diffusion models as guidance while preserving the appearance and geometry of the input object.
    \item A 3D cross-attention based volumetric segmentation technique that defines the spatial extent of textual edits.
    \item Results that demonstrate that our proposed framework can perform a wide array of editing tasks, which cannot be previously achieved.
\end{itemize}




\section{Related Work}

\noindent \textbf{Text-driven Object Editing.} 
Computational methods targeting text-driven image generation and manipulation have seen tremendous progress with the emergence of CLIP~\cite{radford2021learning} and diffusion models~\cite{ho2020denoising}, advancing from specific domains~\cite{styleflow, shen2020interpreting,patashnik2021styleclip,gal2021stylegannada} to more generic ones~\cite{nichol2021glide,avrahami2022blended,couairon2022diffedit,kawar2022imagic}. Several recent methods allow for performing convincing localized edits on real images without requiring mask guidance~\cite{bar2022text2live,hertz2022prompt,brooks2022instructpix2pix,pnpDiffusion2022, Parmar2023ZeroshotIT}. 
However, these methods all operate on single images and cannot facilitate a consistent editing of 3D objects.

While less common, methods for manipulating 3D objects are also gaining increasing interests.
Methods such as LADIS~\cite{huang2022ladis} and ChangeIt3D~\cite{achlioptas2022changeit3d} aim at learning the relations between 3D shape parts and text directly using datasets composed of edit descriptions and shape pairs. These works allow for geometric edits but fail to generalize to out of distribution shapes and cannot modify appearance. 

Alternatively, several methods have proposed leveraging 2D image projections, matching these to a driving text. Text2Mesh~\cite{michel2022text2mesh} uses CLIP for stylizing 3D meshes based on textual prompts. Tango~\cite{chen2022tango} also styles meshes with CLIP, enabling additionally stylization of lighting conditions, reflectance properties and local geometric variations. TEXTure~\cite{richardson2023texture} use a depth-to-image diffusion model for texturing 3D meshes. Unlike our work, these methods focus mostly on texturing meshes, and cannot be used for generating significant geometric modifications, such as adding glasses or other types of accessories.

\begin{figure*}
    \centering
    \includegraphics[width=\textwidth]{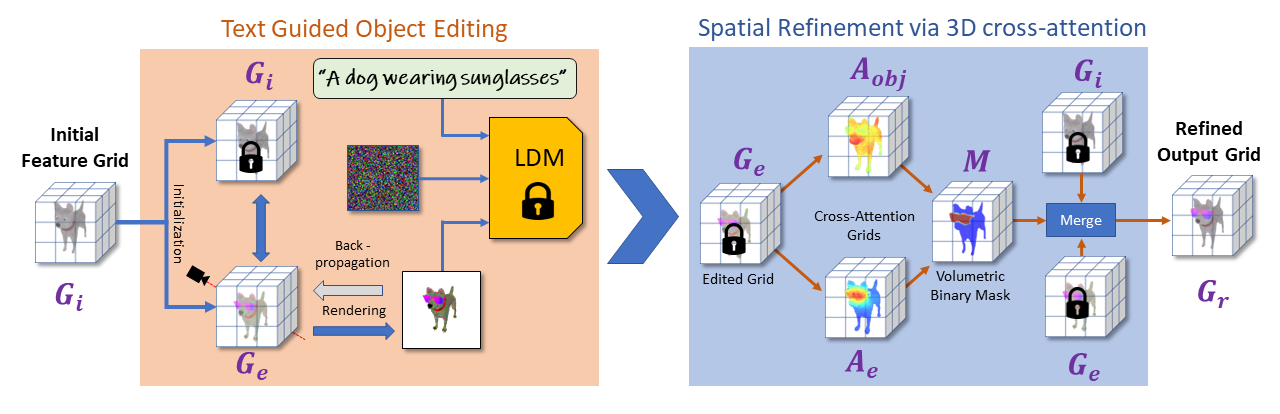}
    \caption{\textbf{An overview of our approach}. Given a set of posed images depicting an object, we optimize an initial feature grid (left). We then perform text-guided object editing using a generative SDS loss and a volumetric regularization, optimizing an edited grid $G_e$. To localize the edits, we optimize 3D cross-attention grids which define probability distributions over the object and the edit regions. We obtain a volumetric mask from these grids using an energy minimization problem over all the voxels. Finally, we merge the initial and edited grid to obtain a refined volumetric grid (right). }
    \label{fig:overview}
\end{figure*}

\medskip
\noindent \textbf{Neural Field Editing.} 
Neural fields (e.g., NeRF~\cite{mildenhall2021nerf}), which can be effectively learned from multi-view images through differentiable rendering, have recently shown great promise for representing object and scenes. Prior works have demonstrated that these fields can be adapted to express different forms of manipulations. ARF~\cite{zhang2022arf} transfers the style of an exemplar image to a NeRF. NeRF-Art~\cite{wang2022nerf} performs a text-driven style transfer. Distilled Feature Fields~\cite{kobayashi2022decomposing}  distill the knowledge of 2D image feature extractors into a 3D feature field and use this feature field to localize edits performed by CLIP-NeRF~\cite{wang2022clip}, which optimizes a radiance field so that its rendered images match with a text prompt via CLIP.

Several works have shown that neural fields can be edited by editing selected 2D images~\cite{liu2021editing,yang2022neumesh}. NeuTex~\cite{xiang2021neutex} uses 2D texture maps, which can be edited directly, to represent the surface appearance. 
Other works demonstrated geometric editing of shapes represented with neural fields via correspondences with an explicit mesh representation~\cite{garbin2022voltemorph,yuan2022nerf,xu2022deforming}, that can be edited using as-rigid-as-possible deformations~\cite{sorkine2007rigid}. However, these cannot easily allow for modifying the 3D mesh to incorporate additional parts, according to the user's provided description. Concurrently to our work, Instruct-NeRF2NeRF ~\cite{haque2023instruct} uses an image editing model to iteratively edit multi-pose images from which an edited 3D scene is reconstructed. Unlike our work which optimizes the underlying 3D representation, they optimize the input images directly. \ignorethis{This approach differs greatly from ours, as the driving optimization occurs on the input images as apposed to the weights of the underlying neural 3D representation.}
Furthermore, our method is based on grid-based representations rather than neural fields, in particular ReLU Fields~\cite{karnewar2022relu}, which do not require any neural networks and instead model the scene as a voxel grid where each voxel contains learned features. We show that having an explicit grid structure is beneficial for editing 3D objects as it enables fast reconstruction and rendering times as well as powerful volumetric regularization. 

\medskip
\noindent \textbf{Text-to-3D.} Following the great success of text-to-image generation, we are witnessing increasing interests in unconditional text-driven generation of 3D objects and scenes. CLIP-Forge~\cite{sanghi2022clip} uses CLIP guidance to generate coarse object shapes from text. Dream Fields~\cite{jain2022zero}, DreamFusion~\cite{poole2022dreamfusion}, Score Jacobian Chaining~\cite{haochen2022score} and Latent-NeRF~\cite{metzer2022latent} optimize radiance fields to generate the geometry and color of objects driven by the text. While DreamFields relies on CLIP, the other three methods instead use a score distillation loss, which enables the use of a pretrained 2D diffusion model. Magic3D~\cite{lin2022magic3d} proposes a two-stage optimization technique to overcome DreamFusion's slow optimization. Unlike these works, we focus on the conditional setting. In our case, a 3D object is provided, and the desired edit should preserve the object's geometry and appearance. Still, we compare with Latent-NeRF in the experiments, as it can use rough 3D shapes as guidance.

\begin{figure*} %
\centering
\jsubfig{\includegraphics[height=2.78cm,trim={0.5cm 0.5cm 0.5cm 0.5cm},clip]{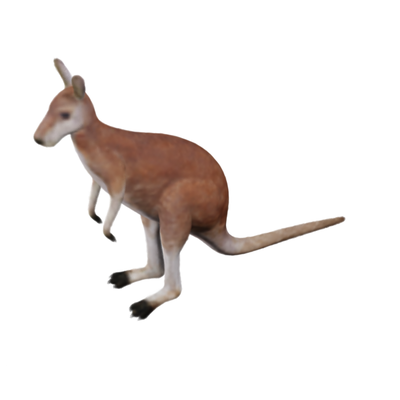}}{}\hfill
\jsubfig{\includegraphics[height=2.78cm,trim={0.5cm 0.5cm 0.5cm 0.5cm},clip]{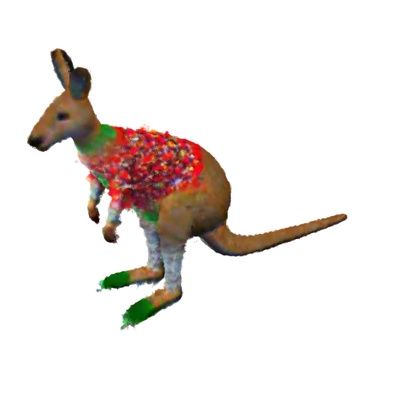}}{} \hfill
\jsubfig{\includegraphics[height=2.78cm]{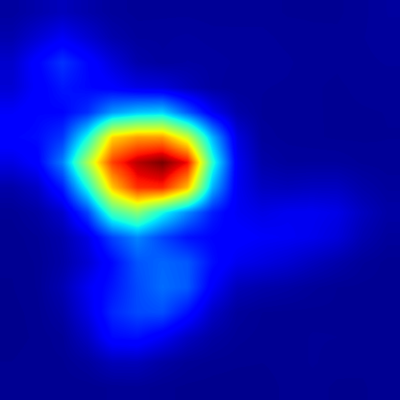}}{} \hfil
\jsubfig{\includegraphics[height=2.78cm]{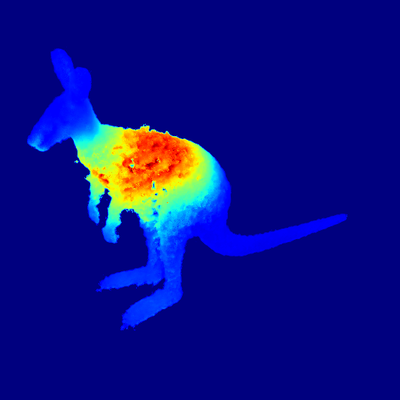}}{}\hfill
\jsubfig{\includegraphics[height=2.78cm]{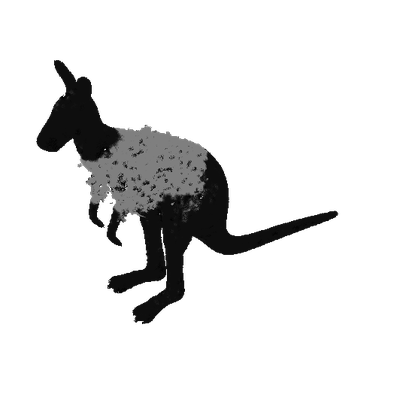}}{}\hfill
\jsubfig{\includegraphics[height=2.78cm,trim={0.5cm 0.5cm 0.5cm 0.5cm},clip]{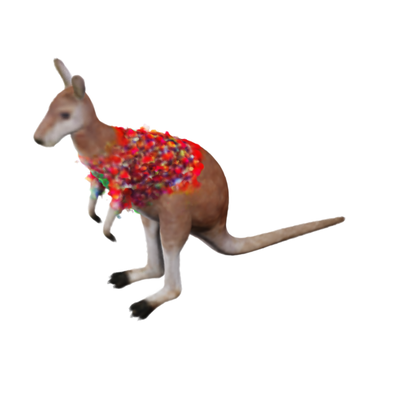}}{}
\vspace{2.2pt} 
\\
\jsubfig{\includegraphics[height=2.78cm,trim={0.5cm 0.5cm 0.5cm 0.5cm},clip]{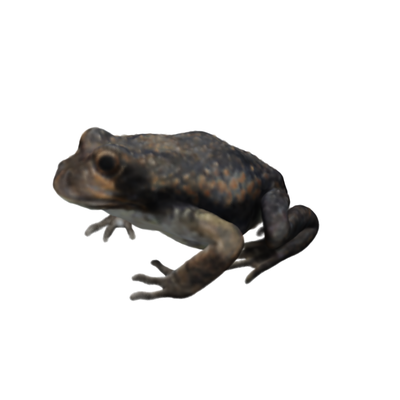}}{\footnotesize {Input Image}} \hfill
\jsubfig{\includegraphics[height=2.78cm,trim={0.5cm 0.5cm 0.5cm 0.5cm},clip]{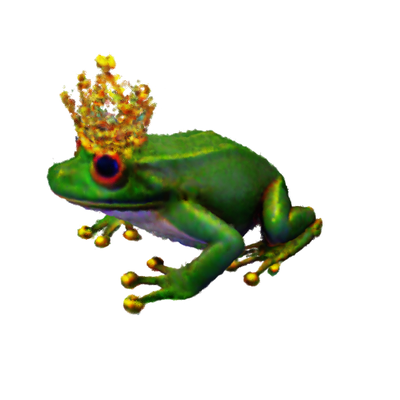}}{\footnotesize {Edited grid}} \hspace{0.1pt}
\jsubfig{\includegraphics[height=2.78cm]{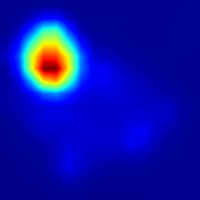}}{\footnotesize {2D Cross Attention Map}} 
\jsubfig{\includegraphics[height=2.78cm]{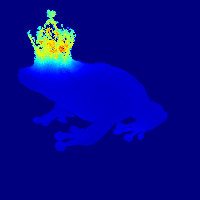}}{\footnotesize {3D Cross Attention Grid}}\hfill
\jsubfig{\includegraphics[height=2.78cm]{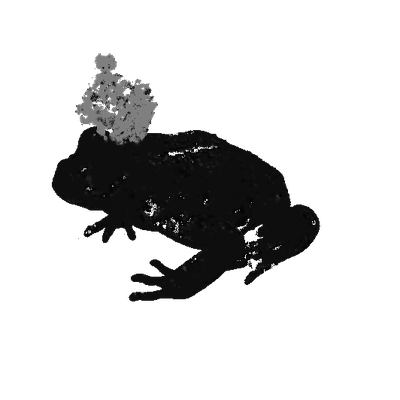}}{\footnotesize {Segmentation Mask}}\hfill
\jsubfig{\includegraphics[height=2.78cm,trim={0.5cm 0.5cm 0.5cm 0.5cm},clip]{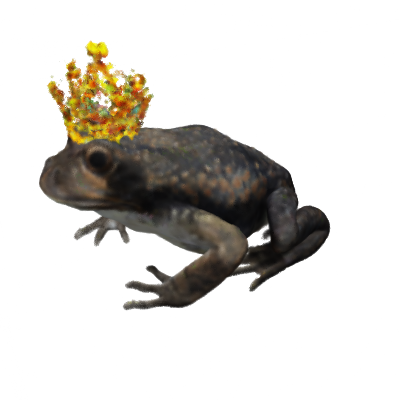}}{\footnotesize {Output}}
\vspace{5pt} 
\caption{\textbf{Optimizing 3D cross-attention grids for edit localization}. 
We leverage rough 2D cross-attention maps (third column) for supervising the training of 3D cross-attention grids (fourth column). Provided with cross-attention grids associated with the edit (as demonstrated above for ``christmas sweater" and ``crown") and object regions, we formulate an energy minimization problem, which outputs a volumetric binary segmentation mask (fifth column). We then merge the features of the input (first column) and edited (second column) grids using this volumetric mask to obtain our final output (rightmost column). Note that warmer colors correspond to higher activations in the cross-attention maps and edited regions are colored in gray in the binary segmentation mask. 
}
\label{fig:attention}
\end{figure*}

\section{Method}
In this work, we consider the problem of editing 3D objects given a captured set of posed multiview images describing this object and a text prompt expressing the desired edit. We first represent the input object with a grid-based volumetric representation (Section \ref{sec:rep}). We then optimize a \emph{coupled} voxel grid, such that it resembles the input grid on the one hand while conforming to the target text on the other hand (Section \ref{sec:editing}). To further refine the spatial extent of the edits, we perform an (optional) refinement step (Section \ref{sec:refine}).
Figure \ref{fig:overview} provides an overview of our approach.

\subsection{Grid-Based Volumetric Representation}
\label{sec:rep}
Our volumetric representation is based on the voxel grid model first introduced in DVGO~\cite{sun2022direct} and later simplified in ReLU Fields~\cite{karnewar2022relu}. We use a 3D grid $G$, where each voxel holds a 4D feature vector. We model the object’s geometry using a single feature channel which represents spatial density values when passed through a ReLU nonlinearity. The three additional feature channels represent the object’s appearance, and are mapped to RGB colors when passed through a sigmoid function. Note that in contrast to most recent neural 3D scene representations (including ReLU Fields)
\ignorethis{where each voxel holds \esc{an $N$ dimensional} feature vector. We model the object's geometry using a single feature channel which represents spatial density values when passed through a \esc{non-linearity (ReLU in the ReLU fields setting, SoftPlus in the DVGO setting)}. The additional feature channels represent the object's appearance, and are mapped to RGB colors when passed through \esc{an activation function (sigmoid in the ReLU fields setting, an MLP in DVGO)}. Note that in contrast to most recent neural 3D scene representations\ignorethis{( including ReLU Fields)}}, we do not model view dependent appearance effects, as we found it leads to undesirable artifacts when guided with 2D diffusion-based models. 

To represent the input object with our grid-based representation, we use images and associated camera poses to perform volumetric rendering as described in NeRF \cite{mildenhall2021nerf}. However, in contrast to NeRF, we do not use any positional encoding and instead sample our grid at each location query to obtain interpolated density and color values, which are then accumulated along each ray. We use a simple L1 loss between our rendered outputs and the input images to learn a grid-based volume $G_i$ that represents the input object.

\subsection{Text-guided Object Editing}
\label{sec:editing}
Equipped with the initial voxel grid $G_i$ described in the previous section, we perform text-guided object editing by optimizing $G_e$, a grid representing the edited object which is initialized from $G_i$. Our optimization scheme combines a generative component, guided by the target text prompt, and a pullback term that encourages the new grid not to deviate too strongly from its initial values. As we later show, our coupled volumetric representation provides added flexibility to our system, allowing for better balancing between the two objectives by regularizing directly in 3D space. Next we describe these two optimization objectives.

\subsubsection*{Generative Text-guided Objective}
To encourage our feature grid to respect the desired edit provided via a textual prompt, we use a Score Distillation Sampling (SDS) loss applied over Latent Diffusion Models (LDMs). SDS was first introduced in DreamFusion \cite{poole2022dreamfusion}, and consists of minimizing the difference between noise injected to a generator's output and noise predicted by a pre-trained Denoising Diffusion Probabilistic Model (DDPM). Formally, at each optimization iteration, noise is added to a generated image $x$ using a random time-step $t$,
\begin{equation}
    x_t = x + \epsilon_t,
\end{equation}
where $\epsilon_t$ is the output of a noising function $Q(t)$ at time-step $t$. The score distillation gradients (computed per pixel) can be expressed as:
\begin{equation}
  \nabla_x\mathcal{L}_{SDS} = w(t)\left( \epsilon_t-\epsilon_\phi(x_t, t, s) \right),
\end{equation}
where $w(t)$ is a weighting function, $s$ is an input guidance text, and $\epsilon_\phi(z_t, t, s)$ is the noise predicted by a pre-trained DDPM with weights $\phi$  given $x_t$, $t$ and $s$.
As suggested by Lin et al.~\cite{lin2022magic3d}, we use an annealed SDS loss which gradually decreases the maximal time-step we draw $t$ from, allowing SDS to focus on high frequency information after the outline of the edit has formed. We empirically found that this often leads to higher quality outputs. 

\ignorethis{
In practice, frameworks that utilize SDS \hec{(such as XX,YY,ZZ)} often select the time-step $t$ randomly at each iteration by drawing from a uniform distribution:
\begin{equation}
    t \sim U[t_0 + \varepsilon, t_{final} + \varepsilon],
\end{equation}
with $\varepsilon$ being some small number, and $t_0$ and $t_{final}$ being the first and final time-steps, respectively. This implies that the noisy image we feed the DDPM with is uniformly distributed between nearly pure noise (at the largest time-step) and and the original un-noised output (at the smallest time-step). 

We observe that as noise levels increase, higher frequency information in the image becomes less and less visible, which in turn means that SDS will struggle in facilitating meaningful changes in the higher frequency information of the generated outputs at higher time-steps. Additionally, at lower noise levels SDS will struggle with driving meaningful low frequency changes, as that is ill-posed in regards to what the model was trained to do at these time-steps. Using this observation, we show that gradually decreasing the maximal time-step we draw $t$ from allows SDS to focus on high frequency information after the outline of the edit has formed, and in turn leads to higher quality outputs. 
}

\subsubsection*{Volumetric Regularization}
\label{sec:volume_reg}
Regularization is key in our problem setting, as we want to avoid over-fitting to specific views and also not to deviate too far from the original 3D representation. Therefore, we propose a volumetric regularization term, which couples our edited grid $G_e$ with the initial grid $G_i$. Specifically, we incorporate a loss term which encourages correlation between the density features of the input grid $f^\sigma_i$ and the density features of the edited grid $f^\sigma_e$:

\begin{equation}
    \mathcal{L}_{reg3D} = 1 - \frac{Cov(f^\sigma_i, f^\sigma_e)}
    {\sqrt{Var(f^\sigma_i)Var(f^\sigma_e)}}
\end{equation}

This volumetric loss has a significant edge over image space losses as it allows for decoupling the appearance of the scene from its structure, thereby connecting the volumetric representations in 3D space rather than treating it as a multiview optimization problem.

\begin{figure} %
\centering 
\rotatebox{90}{\whitetxt{x}2D maps}
\jsubfig{\includegraphics[height=1.9cm] {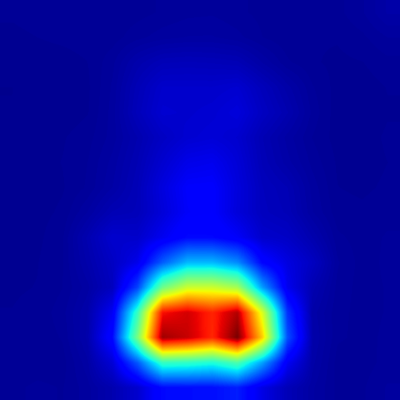}}{}
\jsubfig{\includegraphics[height=1.9cm]{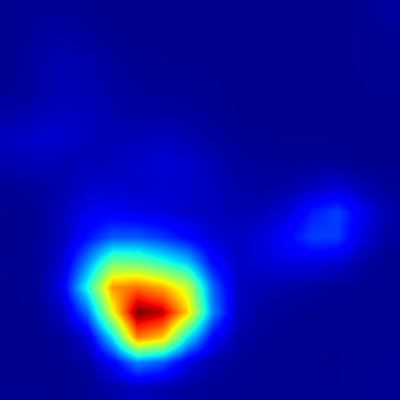}}{}  
\jsubfig{\includegraphics[height=1.9cm] {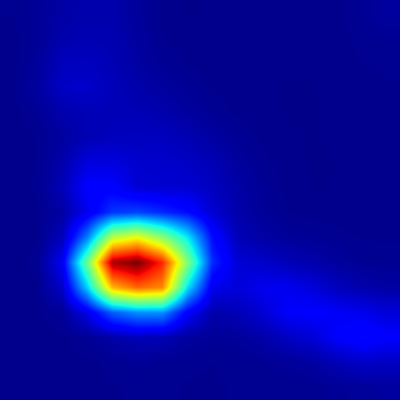}}{}
\jsubfig{\includegraphics[height=1.9cm]{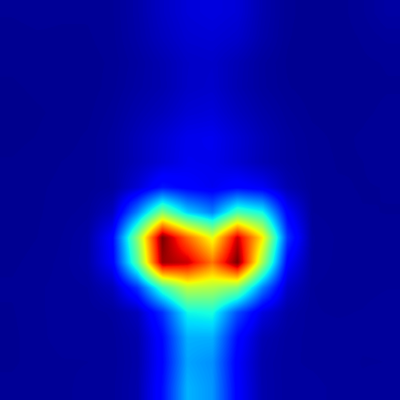}}{}
\\[4pt]
\rotatebox{90}{3D grid ($A_e$)}
\jsubfig{\includegraphics[height=1.9cm]{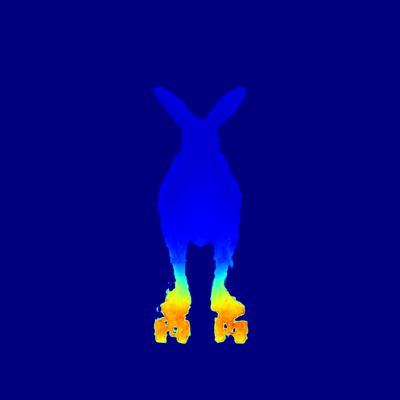}}{} 
\jsubfig{\includegraphics[height=1.9cm]{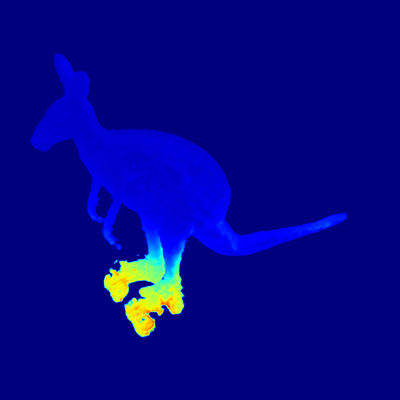}}{}  
\jsubfig{\includegraphics[height=1.9cm] {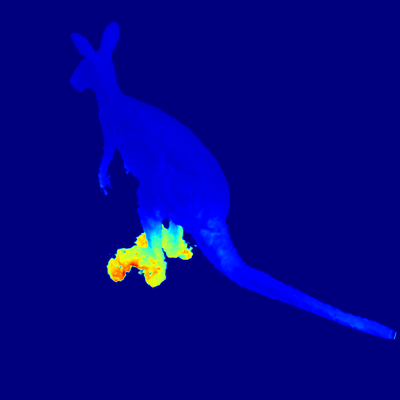}}{}
\jsubfig{\includegraphics[height=1.9cm]{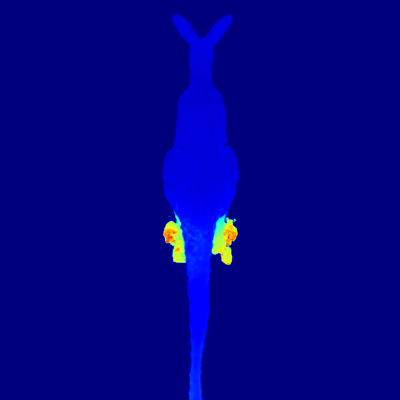}}{}

\vspace{5pt}

\caption{\textbf{Cross-attention 2D maps and rendered 3D grids over multiple viewpoints}, obtained for the token associated with the word ''rollerskates" (from the ''kangaroo on rollerskates" text prompt). While 2D cross-attention may yield inconsistent observations, such as high probabilities over the tail region in the rightmost column, our 3D grids can more accurately localize the region of interest (effectively smoothing out such inconsistencies).
}

\label{fig:extra_attn}
\end{figure}

\subsection{Spatial Refinement via 3D Cross-Attention}
\label{sec:refine}
While our optimization framework described in the previous section can mostly preserve the shape and the identity of a 3D object, for local edits, it is usually desirable to only change specific \emph{local} regions, while keeping other regions completely fixed.
Therefore, we add an (optional) refinement step which leverages the signal from cross-attention layers to produce a volumetric binary mask $M$ that marks the voxels which should be edited. We then obtain the refined grid $G_r$ by merging the input grid $G_i$ and edited $G_r$ grid as
\begin{equation}
   G_r =  M \cdot G_e + (1 - M) \cdot G_i.
\end{equation}

\ignorethis{
Formally, we define the volumetric binary mask $M$ as follows:
\begin{equation}
    M[i, j, k] = 
    \begin{cases}
           1 &  [i, j, k]\in E\\
           0 &\text{otherwise}, \\ 
         \end{cases}
\end{equation}
where $E$ is the latent set of voxels that should be edited.
}

In the context of 2D image editing with diffusion models, the outputs of the cross-attention layers roughly capture the spatial regions associated with each word (or token) in the text. More concretely, these cross attention maps can be interpreted as probability distributions over tokens for each image patch~\cite{hertz2022prompt,brooks2022instructpix2pix}. 
%
We elevate these 2D probability maps to a 3D grid by using them as supervision for training a ReLU field. We initialize the density values from the ReLU field trained in Section~\ref{sec:rep} and keep these fixed, while using probability maps in place of color images and optimizing for the probability values in the grid using an L1 loss.
As shown in Figures~\ref{fig:attention} and ~\ref{fig:extra_attn}, optimizing for a volumetric representation allows for ultimately refining the 2D probability maps, for instance by resolving over inconsistent 2D observations (as illustrated in Figure~\ref{fig:extra_attn}).

We then convert these 3D probability fields to our binary mask $M$ using a seam-hiding segmentation algorithm based on energy minimization~\cite{agarwala2004interactive}. Specifically, we extract a segmentation grid that minimizes an energy function composed of two terms: A \emph{unary} term, which penalizes disagreements with the label probabilities, and a \emph{smoothness} term, which penalizes large pairwise color differences within similarly-labeled voxels. We define the label probabilities for voxel cell as the element-wise softmax of two cross-attention grids $A_{e}$ and $A_{obj}$, where
\begin{itemize}
\item $A_{e}$ is the cross-attention grid associated with the token describing the edit (e.g.\ \emph{sunglasses}), and
\item $A_{obj}$ is the grid associated with the object, defined as the maximum probability over all other tokens in the prompt.
\end{itemize}
We compute the smoothness term from local color differences in the edited grid $G_e$. That is, we sum
\begin{equation}
    w_{pq}=\text{exp}\left(\frac{-(c_p - c_q)^2}{2\sigma^2}\right)
\end{equation}
for each pair of same-labeled neighboring voxels $p$ and $q$, where $c_p$ and $c_q$ are RGB colors from $G_e$. In our experiments, we use $\sigma=0.1$ and balance the data and smoothness terms with a parameter $\lambda=5$ (strengthening the smoothness term). Finally, we solve this energy minimization problem via graph cuts~\cite{boykov2001fast}, resulting in the high quality segmentation masks shown in Figure~\ref{fig:attention}.

\begin{figure} %
\centering 
\ignorethis{
\jsubfig{\includegraphics[height=1.9cm,trim={2.0cm 3.0cm 1.0cm 3.0cm},clip]{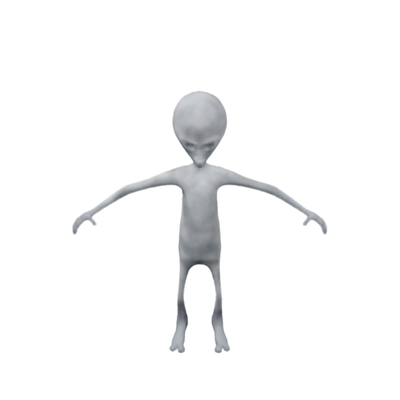}\includegraphics[height=1.9cm,trim={2.0cm 3.0cm 1.0cm 3.0cm},clip]{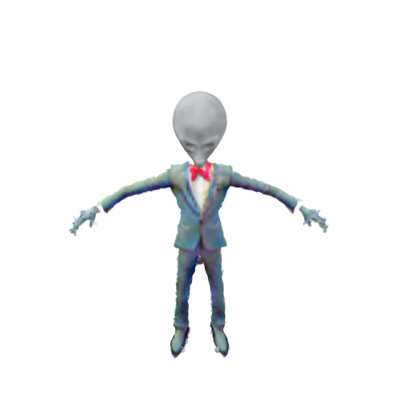}}{\footnotesize {``An alien wearing a tuxedo"}}
}
\jsubfig{\includegraphics[height=1.9cm,trim={0.6cm 1.2cm 0.4cm 1.2cm},clip]{images/results/input_alien.png}\includegraphics[height=1.9cm,trim={0.8cm 1.2cm 0.4cm 1.2cm},clip]{images/results/our_alien.png}}{\footnotesize {``An alien wearing a tuxedo"}}
\hfill
\jsubfig{\includegraphics[height=1.9cm]{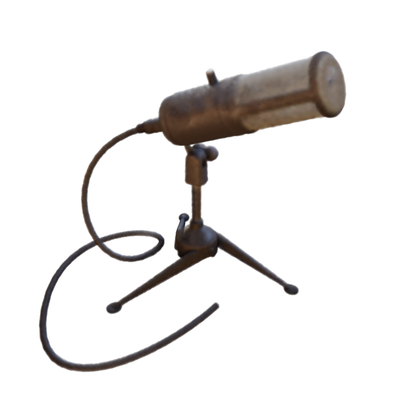}\includegraphics[height=1.9cm]{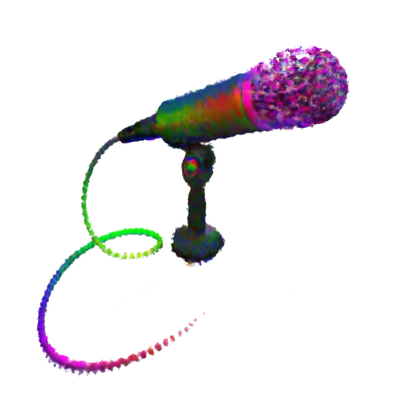}}{\footnotesize {``A rainbow colored microphone"}}
\\
\ignorethis{
\jsubfig{\includegraphics[height=1.9cm, trim={1.0cm 3.0cm 3.0cm 1.0cm}, clip]{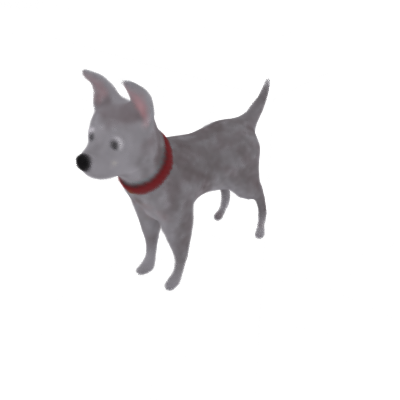}\includegraphics[height=1.9cm, trim={3.0cm 10.0cm 5.0cm 2.0cm}, clip]{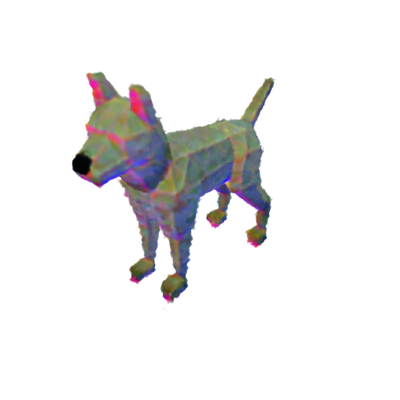}}{\footnotesize {``A dog in low-poly video game style"}}
}
\jsubfig{\includegraphics[height=1.9cm, trim={0.4cm 2.5cm 1.2cm 0.4cm}, clip]{images/results/reg_dog20.png}\includegraphics[height=1.9cm, trim={0.6cm 2.0cm 1.0cm 0.4cm}, clip]{images/results/low_poly_20.png}}{\footnotesize {``A dog in low-poly video game style"}}
\hfill
\jsubfig{\includegraphics[height=1.9cm]{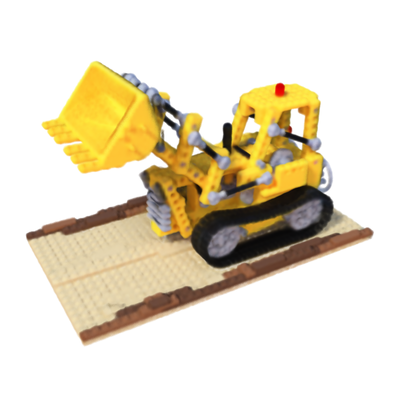}\includegraphics[height=1.9cm]{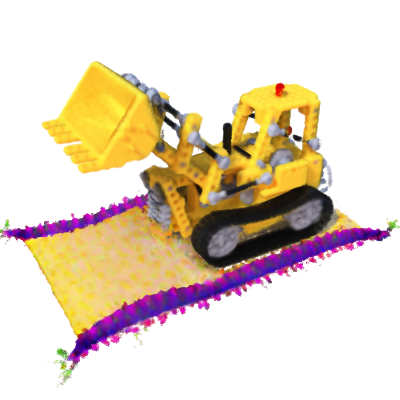}}{\footnotesize {``A bulldozer on a magic carpet"}}
\\
\ignorethis{
\jsubfig{\includegraphics[height=1.9cm,trim={5.0cm 9.0cm 5.0cm 3.0cm},clip]{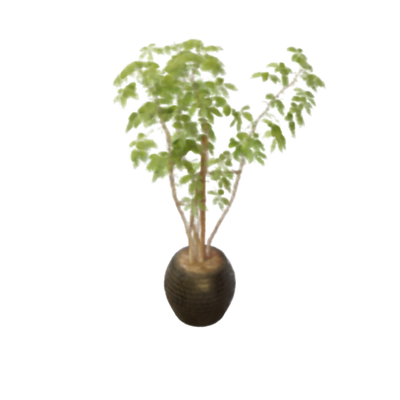}\includegraphics[height=1.9cm,trim={5.0cm 9.0cm 5.0cm 3.0cm},clip]{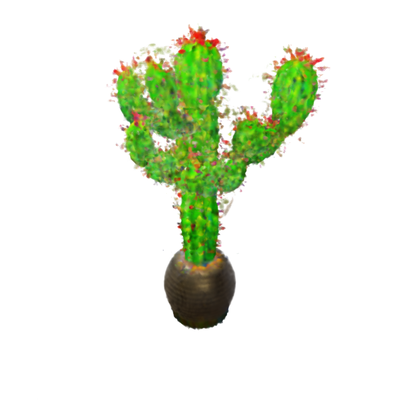}}{\footnotesize {``A cactus in a pot"}}
}
\whitetxt{x}
\jsubfig{\includegraphics[height=1.9cm, trim={1.2cm 0.8cm 1.2cm 0.4cm},clip]{images/results/ficus.png}\includegraphics[height=1.9cm, trim={1.2cm 0.8cm 1.2cm 0.4cm},clip]{images/results/cactus.png}}{\footnotesize {``A cactus in a pot"}}
\hfill
\ignorethis{
\jsubfig{\includegraphics[height=1.9cm,trim={5.0cm 9.0cm 5.0cm 5.0cm},clip]{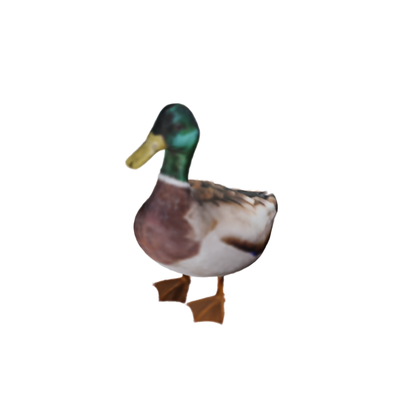}\includegraphics[height=1.9cm,trim={3.0cm 2.0cm 3.0cm 1.0cm},clip]{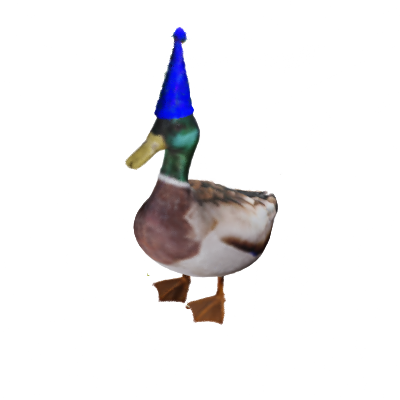}}{\footnotesize {``A duck with a wizard hat"}}
}
\jsubfig{\includegraphics[height=1.9cm,trim={1.0cm 1.8cm 1.7cm 1.0cm},clip]{images/results/input_duck.png}\includegraphics[height=1.9cm,trim={1.7cm 1.8cm 1.0cm 1.0cm},clip]{images/results/ours_duck.png}}{\footnotesize {``A duck with a wizard hat"}}
\\
\jsubfig{
\whitetxt{x}
\includegraphics[height=1.65cm]{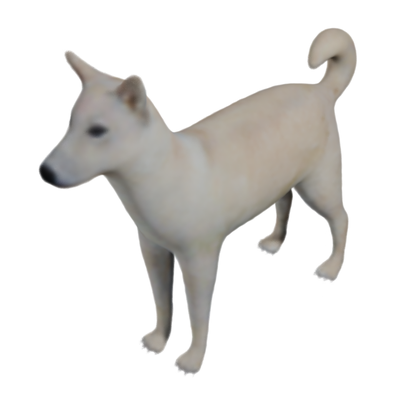}
\whitetxt{x}
\includegraphics[height=1.65cm]{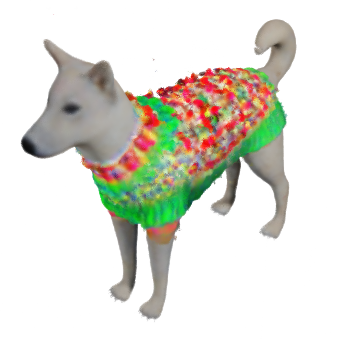}}{\footnotesize {``A dog wearing a christmas sweater"}}
\hfill
\whitetxt{x}
\jsubfig{\includegraphics[height=2.0cm, trim={0.1cm 0.5cm 0.8cm 0.4cm},clip]{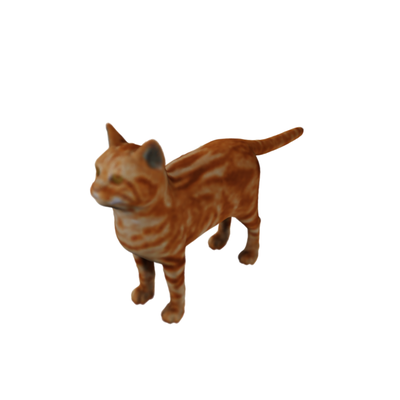}\includegraphics[height=1.9cm, trim={0.4cm 0.4cm 0.1cm 0.4cm},clip]{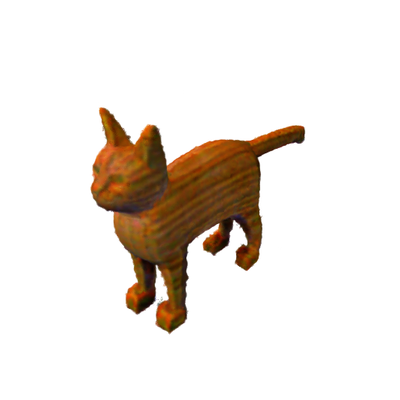}}{\footnotesize {``A cat made of wood"}}
\vspace{2.5pt}
\caption{Results obtained by our method over different objects and prompts (with the inputs displayed on the left). Please refer to the supplementary material for additional qualitative results.}
\label{fig:results}
\end{figure}

\section{Experiments}
\label{sec:results}
\begin{figure} %
\centering
\rotatebox{90}{\whitetxt{x}DFF+CN}
\jsubfig{\includegraphics[height=1.91cm, trim={3.0cm 2.cm 2.7cm 2.5cm}, clip]{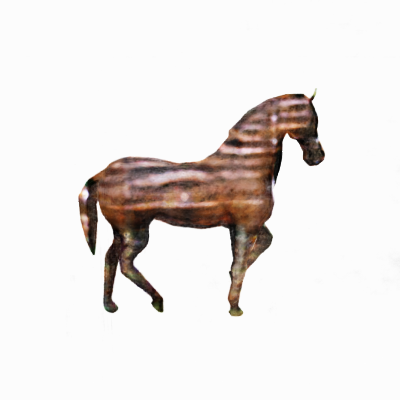}}{}\hfill
\jsubfig{\includegraphics[height=1.91cm, trim={3.0cm 3.0cm 2.7cm 2.5cm}, clip]{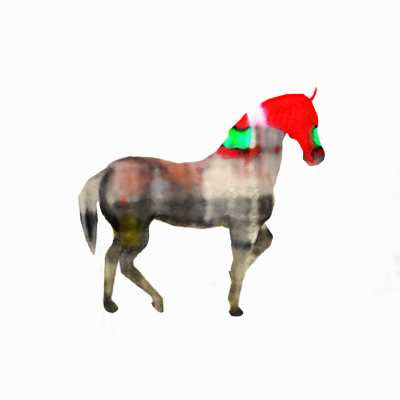}}{} \hfill
\jsubfig{\includegraphics[height=1.91cm, trim={3.0cm 3.0cm 2.7cm 2.5cm}, clip]{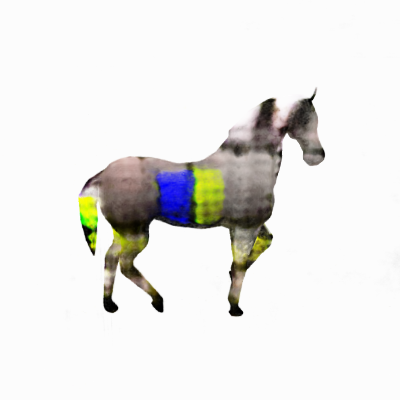}}{} \hfill
\jsubfig{\includegraphics[height=1.91cm, trim={3.0cm 3.0cm 2.7cm 2.5cm}, clip]{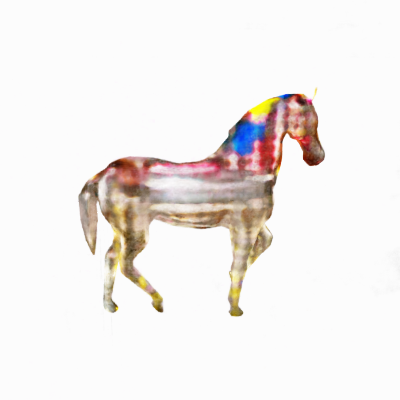}}{} \vspace{1.5pt}
\\ 
\rotatebox{90}{Latent-Paint~}
\jsubfig{\includegraphics[height=1.91cm]{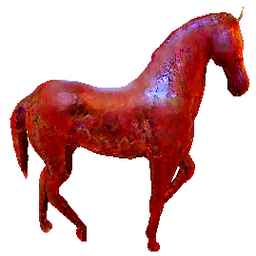}}{}\hfill
\jsubfig{\includegraphics[height=1.91cm]{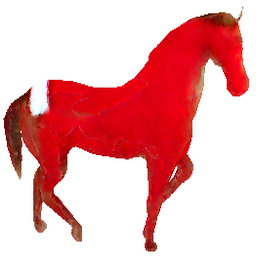}}{} \hfill
\jsubfig{\includegraphics[height=1.91cm]{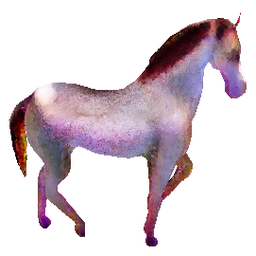}}{}  \hfill
\jsubfig{\includegraphics[height=1.91cm]{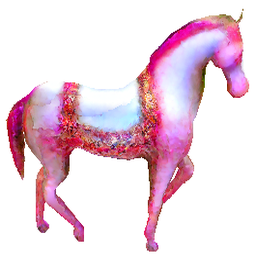}}{} \vspace{1.5pt} 
\\ 
\rotatebox{90}{SketchShape}
\jsubfig{\includegraphics[height=1.91cm]{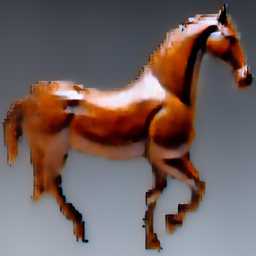}}{} \hfill
\jsubfig{\includegraphics[height=1.91cm]{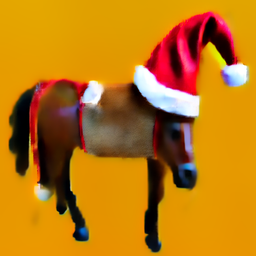}}{} \hfill
\jsubfig{\includegraphics[height=1.91cm]{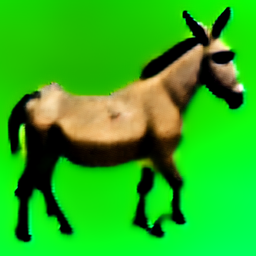}}{} \hfill
\jsubfig{\includegraphics[height=1.91cm]{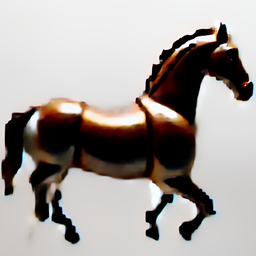}}{} \vspace{1.5pt} 
\\
\rotatebox{90}{\whitetxt{x}Text2Mesh}
\jsubfig{\includegraphics[height=1.91cm]{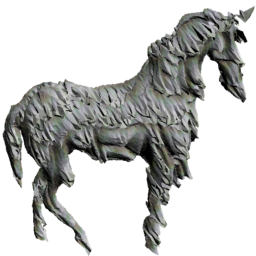}}{}\hfill
\jsubfig{\includegraphics[height=1.91cm]{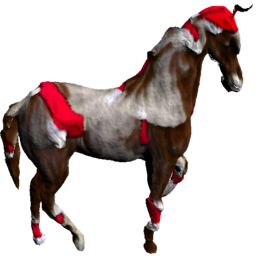}}{} \hfill
\jsubfig{\includegraphics[height=1.91cm]{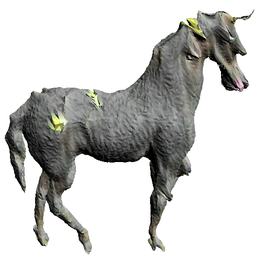}}{} \hfill
\jsubfig{\includegraphics[height=1.91cm]{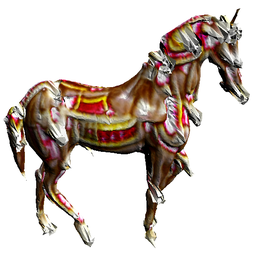}}{} \vspace{1.5pt}
\\
\rotatebox{90}{\whitetxt{xxx}Ours}
\jsubfig{\includegraphics[height=1.91cm]{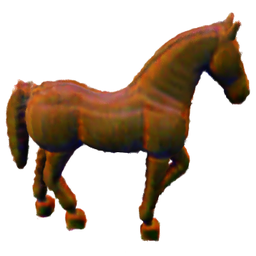}}{\footnotesize {``A wood carving of a horse"}}\hfill
\jsubfig{\includegraphics[height=1.91cm]{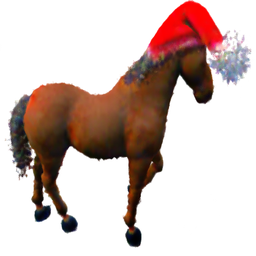}}{\footnotesize {``Horse wearing a santa hat"}} \hfill
\jsubfig{\includegraphics[height=1.91cm]{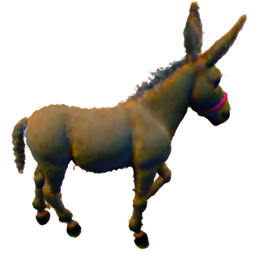}}{\footnotesize {``A donkey"}}\hfill
\jsubfig{\includegraphics[height=1.91cm]{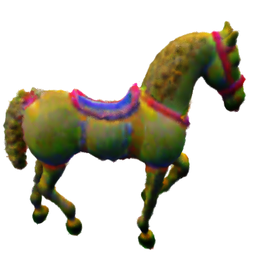}}{\footnotesize {``A carousel horse"}}
\vspace{5pt} 
\caption{\textbf{Comparison to other 3D Object editing techniques}. We show qualitative results obtained using  Text2Mesh~\cite{michel2022text2mesh}, two applications of Latent-NeRF ~\cite{metzer2022latent} (Latent-Paint and SketchShape) and DFF+CN~\cite{kobayashi2022decomposing,wang2022clip} and compare to our method. To accommodate their problem setting, the top three methods are provided with uncolored meshes. Note that the input meshes are visible on the second row from the top \ignorethis{row} (as Latent-Paint does not edit the object's geometry). As illustrated above, prior methods struggle at achieving semantic localized edits. Our method succeeds, while maintaining high fidelity to the input object. 
}
\label{fig:comparisons}
\end{figure}

We show qualitative editing results over diverse 3D objects and various edits in Figures \ref{fig:teaser}, \ref{fig:results},  \ref{fig:comparisons}, 
\ref{fig:supp_realscenes},
Please refer to the supplementary material for many fly-through visualizations demonstrating that our results are indeed consistent across different views.
%

%

To assess the quality of our object editing approach, we conduct several sets of experiments, quantifying the extent of which these rendered images conform to the target text prompt. We provide comparisons with prior 3D object editing methods in Section \ref{sec:comp3D}, and comparisons to 2D editing methods in Section \ref{sec:comp2D}. An additional comparison to an unconditional text-to-3D method is presented in Section \ref{sec:comp_uncond} \ignorethis{Comparisons with prior 3D object editing methods are provided in Section \ref{sec:comp3D}. Comparisons to 2D editing methods are provided in Section \ref{sec:comp2D}.} Results over real scenes are illustrated in Section \ref{sec:realscenes}. We show ablations in Section \ref{sec:ablations}.  Finally, we discuss limitations in Section \ref{sec:limitations}.  Additional results, visualizations, ablations and comparisons can be found in the supplementary material.

\medskip \noindent \textbf{Synthetic Object Dataset.} 
We assembled a dataset using freely available meshes found on the internet. 
Each mesh was rendered from 100 views in Blender. 
For a quantitative evaluation, we paired each object in our dataset with
a number of both local and global edit prompts including:
\begin{itemize}
    \item  ``A $\left<object\right>$ wearing sunglasses''.
    \item ``A $\left<object\right>$ wearing a party hat''. 
    \item ``A $\left<object\right>$ wearing a Christmas sweater''.
    \item ``A  yarn doll of a $\left<object\right>$''.
    \item ``A wood carving of a $\left<object\right>$''.
\end{itemize}

 \noindent We separately evaluate local and global edits, using our spatial refinement step over local edits only. For instance, the first three prompts above are considered local edits (where regions that are not associated with the text prompt should remain unchanged) and the last two as edits that should produce global edits. We provide additional details in the supplementary material.

\medskip \noindent \textbf{Runtime.} All experiments were performed on a single RTX A5000 GPU (24GB VRAM). The training time for our method is approx. 50 minutes for the editing stage and 15 minutes for the optional refinement stage.

\ignorethis{
 We separately evaluate local and global edits. For instance, the first three prompts above are considered local edits and the last two as edits that should produce global edits. We provide additional details in the supplementary material.
}


\subsection{Metrics}
\medskip \noindent \textbf{Edit Fidelity.} We evaluate how well the generated results capture the target text prompt using two metrics:  
 \smallskip \newline \emph{CLIP Similarity} ($\text{CLIP}_{Sim}$) measures the semantic similarity between the output objects and the target text prompts. We encode both the prompt and images rendered from our 3D outputs using CLIP's text and image encoders, respectively, and measure the cosine-distance between these encodings. 
 \smallskip \newline \emph{CLIP Direction Similarity} ($\text{CLIP}_{Dir}$) evaluates the quality of the edit in regards to the input by measuring the directional CLIP similarity first introduced by Gal et al.~\cite{gal2021stylegannada}. This metric measures the cosine distance between the direction of the change from the input and output rendered images and the direction of the change from an input prompt (\emph{i.e.} ``a dog") to the one describing the edit (\emph{i.e.} ``a dog wearing a hat").


\medskip \noindent \textbf{Edit Magnitude.}  For ablating components in our model, we use the Frech\'et Inception Distance (FID) \cite{Heusel2017GANsTB,Seitzer2020FID} to measure the difference in visual appearance between: (i) the output and input images ($\text{FID}_{Input}$) and (ii) the output and images generated by the initial reconstruction grid ($\text{FID}_{Rec}$). We show both to demonstrate to what extent the appearance is affected by the edit versus the expressive power of our framework.

\definecolor{graytext}{RGB}{130,130,130}

\begin{table}[t]
\setlength{\tabcolsep}{3.0pt}
 \def\arraystretch{1.1}
\centering
\resizebox{0.75\linewidth}{!}{
\begin{tabular}{llccc}
\toprule
   & Method  & $\text{CLIP}_{Sim}\uparrow$ & $\text{CLIP}_{Dir} \uparrow$  \\ 
    \midrule 
    \multirow{4}{*}{\rotatebox[origin=c]{90}{Local}} & DFF+CN & \color{graytext}{0.34*} & \color{graytext}{0.05*} \\
    & Text2Mesh & \color{graytext}{0.36*} & \color{graytext}{0.08}* \\
    & Latent-NeRF (Sketch / Paint) & 0.32 / 0.31 & 0.01 / 0.01  \\
    & Ours &\textbf{0.36} & \textbf{0.07} \\
    \midrule 
    \multirow{4}{*}{\rotatebox[origin=c]{90}{Global}} & DFF+CN & \color{graytext}{0.32*} & \color{graytext}{0.01*}\\
    & Text2Mesh & \color{graytext}{0.34*} & \color{graytext}{0.03*}  \\
    & Latent-NeRF (Sketch / Paint) & 0.30 / 0.31 & 0.01 / 0.01 \\
    & Ours & \textbf{0.34} & \textbf{0.02} \\
\bottomrule
\end{tabular}
}
\caption{\textbf{Quantitative Evaluation.} We compare against the 3D object editing techniques Text2Mesh~\cite{michel2022text2mesh}, two variants of Latent-NeRF~\cite{metzer2022latent}: SketchShape (Sketch) and Latent-Paint (Paint) and DFF+CN~\cite{kobayashi2022decomposing,wang2022clip}, over local (top) and global (bottom) edits.  *Note that Text2Mesh and DFF+CN explicitly train to minimize a CLIP loss, and thus directly comparing them is uninformative over these metrics.
}
\label{tab:baseline-stats}
\end{table}

\begin{figure} %
\centering 
 \jsubfig{\includegraphics[height=1.9cm]{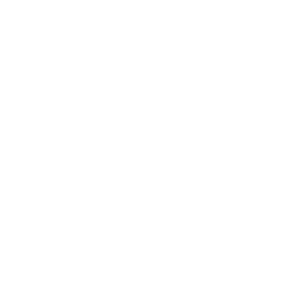}}{}
 \rotatebox{90}{\whitetxt{x}SDEdit\whitetxt{x}\includegraphics[height=0.24cm]{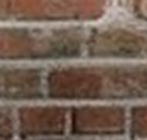}}
 \hfill
\jsubfig{\includegraphics[height=1.9cm]{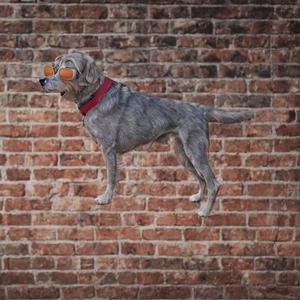}}{} 
\jsubfig{\includegraphics[height=1.9cm] {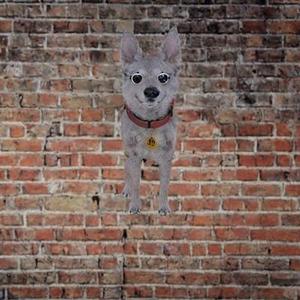}}{} 
\jsubfig{\includegraphics[height=1.9cm]{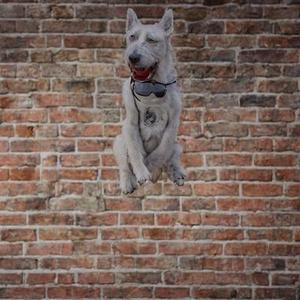}}{} 
\\ 
\jsubfig{\includegraphics[height=1.9cm]{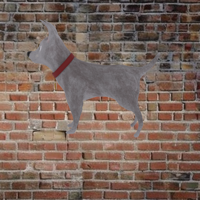}}{}
\rotatebox{90}{\whitetxt{x}IPix2Pix\whitetxt{x}\includegraphics[height=0.24cm]{images/bg.png}}
\hfill 
\hfill 
\jsubfig{\includegraphics[height=1.9cm]{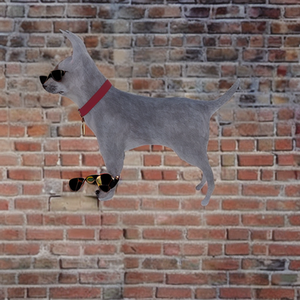}}{} 
\jsubfig{\includegraphics[height=1.9cm] {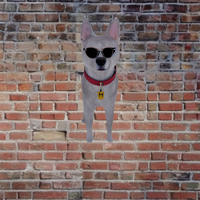}}{} 
\jsubfig{\includegraphics[height=1.9cm]{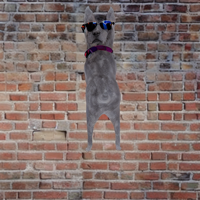}}{} 
\\
\jsubfig{\includegraphics[height=1.9cm] {images/comparisons/image_editing/dog_white.png}}{}
\rotatebox{90}{\whitetxt{xxx}SDEdit}
\hfill
\jsubfig{\includegraphics[height=1.9cm]{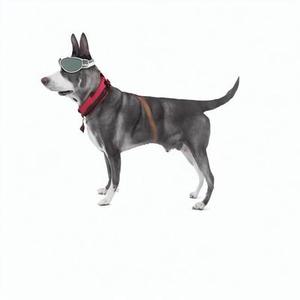}}{} 
\jsubfig{\includegraphics[height=1.9cm] {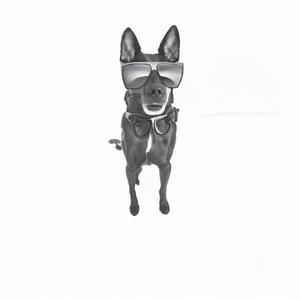}}{} 
\jsubfig{\includegraphics[height=1.9cm]{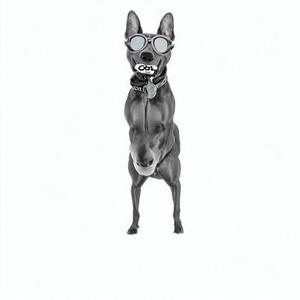}}{} 
\\ 
\jsubfig{\includegraphics[height=1.9cm] {images/comparisons/image_editing/dog_white.png}}{}
\rotatebox{90}{\whitetxt{xx}IPix2Pix}
\hfill
\jsubfig{\includegraphics[height=1.9cm]{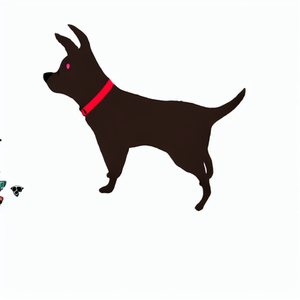}}{} 
\jsubfig{\includegraphics[height=1.9cm] {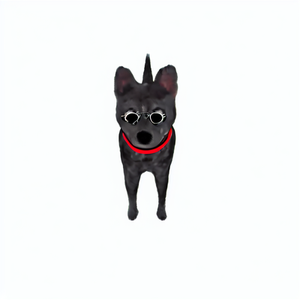}}{} 
\jsubfig{\includegraphics[height=1.9cm]{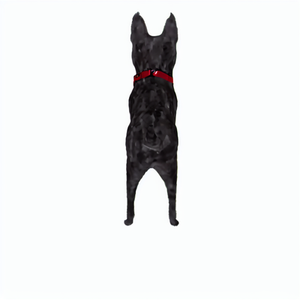}}{} 
\\
\jsubfig{\includegraphics[height=1.9cm]{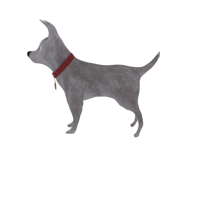}}{}\rotatebox{90}{\whitetxt{xxxxx}Ours}
\hfill
\jsubfig{\includegraphics[height=1.9cm]{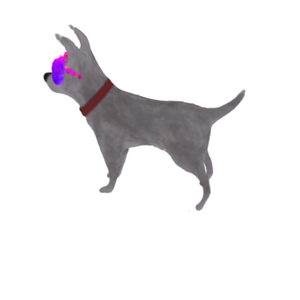}}{} 
\jsubfig{\includegraphics[height=1.9cm] {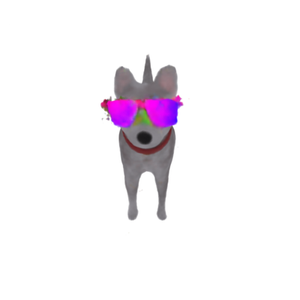}}{} 
\jsubfig{\includegraphics[height=1.9cm] {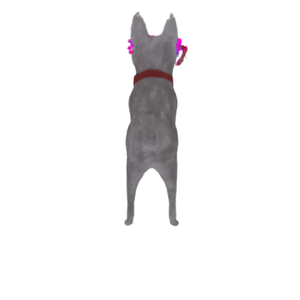}}{} 
\vspace{-12pt} 
\caption{\textbf{Comparison to 2D image editing techniques}. We compare to the text-guided image editing techniques InstructPix2Pix (IPix2Pix)~\cite{brooks2022instructpix2pix} and SDEdit~\cite{meng2022sdedit} by providing it with images from different viewpoints and a target instruction text prompt (``put sunglasses on the dog" for IPix2Pix and ``a dog with sunglasses" for SDEdit and our method). We show one input image on the left, and three outputs on the right (side, front and back views), where the leftmost output corresponds to the input viewpoint. We show two variants, one with added backgrounds (top rows), as we observe that it allows for better preserving the object's appearance. As illustrated above, 2D techniques cannot easily achieve 3D-consistent edit results (illustrated, for instance, by the sunglasses added on the dog's back). 
}
\label{fig:comparison2d}
\end{figure}

\begin{figure*}[ht!] %
\centering
\jsubfig{\includegraphics[height=1.56cm]{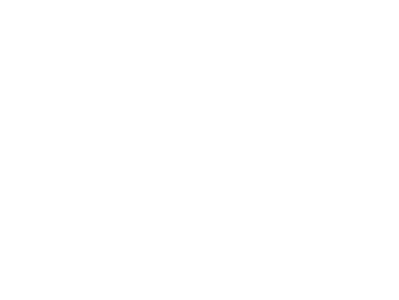}}{}\hfill
\jsubfig{\includegraphics[height=1.53cm]{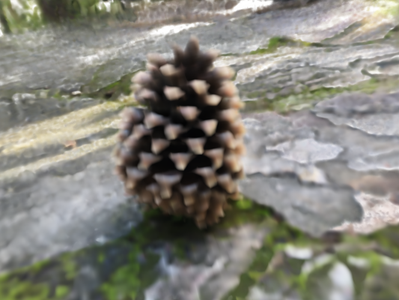}}{} \hfill
\jsubfig{\includegraphics[height=1.53cm]{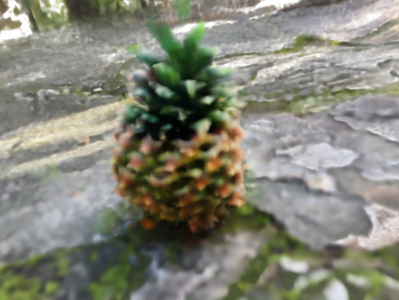}}{} \hfill
\jsubfig{\includegraphics[height=1.53cm]{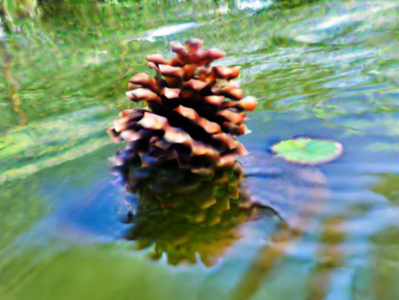}}{}\hfill
\jsubfig{\includegraphics[height=1.53cm]{images/real_scenes/pinecone_white.png}}{} \hfill
\jsubfig{\includegraphics[height=1.53cm]{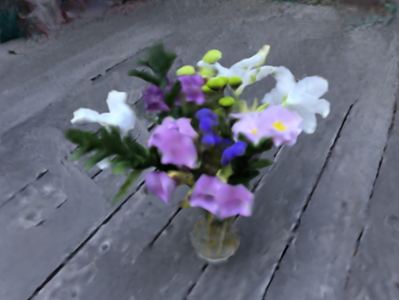}}{} \hfill
\jsubfig{\includegraphics[height=1.53cm]{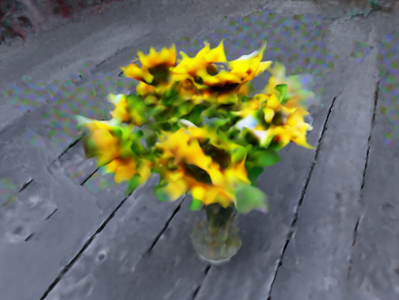}}{} \hfill
\jsubfig{\includegraphics[height=1.53cm]{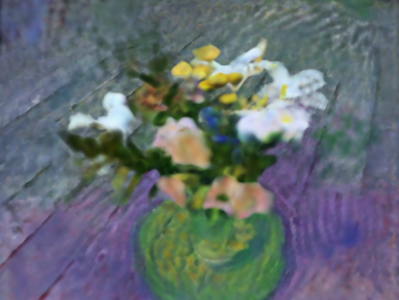}}{} 
\rotatebox[origin=tc]{-90}{RF \whitetxt{xxxx}}
\vspace{-8pt} 
\\ 
\jsubfig{\includegraphics[height=1.53cm]{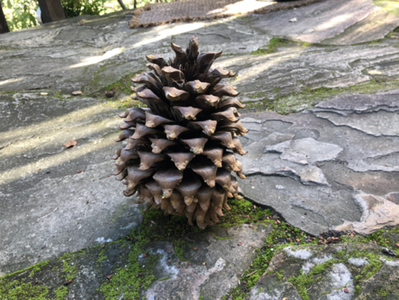}}{\footnotesize {Input}} \hfill
\jsubfig{\includegraphics[height=1.53cm]{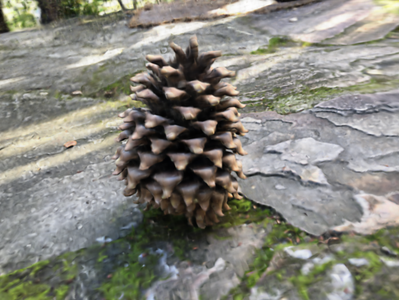}}{\footnotesize {Initial}} \hfill
\jsubfig{\includegraphics[height=1.53cm]{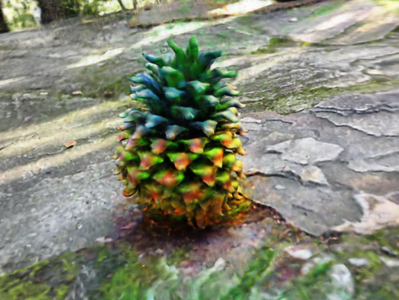}}{\footnotesize {''A pineapple on the ground"}} \hfill
\jsubfig{\includegraphics[height=1.53cm]{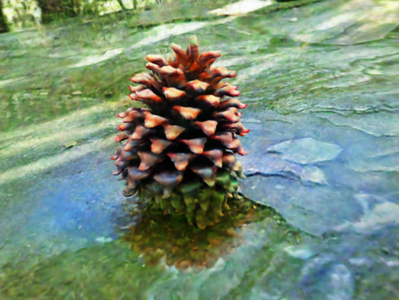}} 
{\footnotesize {''A pinecone floating in a pond"}} \hfill
\jsubfig{\includegraphics[height=1.53cm]{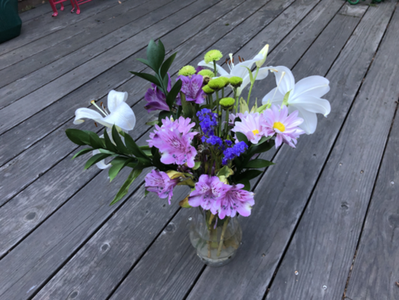}} {\footnotesize {Input}} \hfill
\jsubfig{\includegraphics[height=1.53cm]{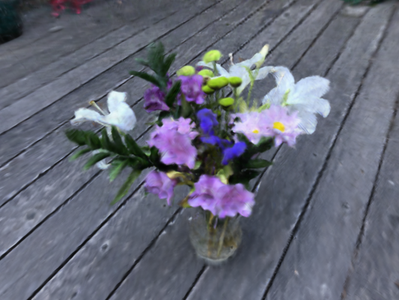}} 
{\footnotesize {Initial}} \hfill
\jsubfig{\includegraphics[height=1.53cm]{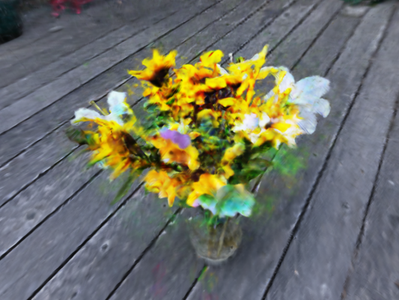}}{\footnotesize {''A vase full of sunflowers"}} \hfill
\jsubfig{\includegraphics[height=1.53cm]{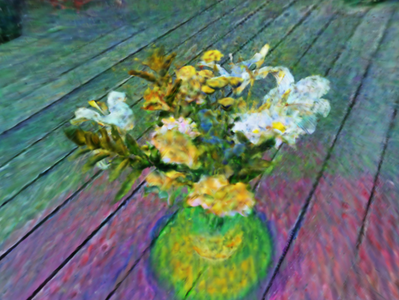}}{\footnotesize {A van-Gogh painting of a flower vase"}}
\rotatebox[origin=tc]{-90}{DVGO\whitetxt{xxxxx}}
\vspace{1.5pt}
\\ 

\caption{
\textbf{Editing real scenes with different underlying 3D representations.} 
We show results obtained when using DVGO \cite{sun2022direct} (bottom row) and ReLU-Fields (RF, top row). We show samples from the input image dataset (leftmost columns), initial scene reconstructions (second columns), results over local edits (third columns) and results over global edits (rightmost columns).}
\label{fig:supp_realscenes}
\end{figure*}
\subsection{3D Object Editing Comparisons}
\label{sec:comp3D}

To the best of our knowledge, there is no prior work that can directly perform our task of text-guided localized edits for 3D objects given a set of posed input images. 
Thus, we consider Distilled Feature Fields~\cite{kobayashi2022decomposing} combined with CLIP-NeRF~\cite{wang2022clip} (DFF+CN), Text2Mesh~\cite{michel2022text2mesh} and Latent-NeRF~\cite{metzer2022latent} which can be applied in a similar setting to ours. These experiments highlight the differences between prior works and our proposed editing technique.

Distilled Feature Fields~\cite{kobayashi2022decomposing} distills 2D image features into a 3D feature field to enable query-based local editing of a 3D scenes. CLIP-NeRF edits a neural radiance field by optimizing the CLIP score of the input query and the rendered image. Combining these two methods allows to edit only the relevant parts of the 3D scene.
Text2Mesh~\cite{michel2022text2mesh} aims at editing the style of a given input mesh to conform to a target prompt with a style transfer network that predicts color and a displacement along the normal direction. As it only predicts displacements along the normal direction, the geometric edits enabled by Text2Mesh are limited mostly to small changes.
Latent-Paint and SketchShape are two applications introduced in Latent-Nerf~\cite{metzer2022latent} which operate on input meshes. 
SketchShape generates shape and appearance from coarse input geometry, while Latent-Paint only edits the appearance of an existing mesh.
%
Note that Text2Mesh and Latent-NeRF are designed for slightly more constrained inputs than our approach. While our focus is on editing 3D models with arbitrary textures (as depicted from associated imagery), they only operate on uncolored meshes. 

We show a qualitative comparison in Figure \ref{fig:comparisons} over an uncolored mesh (its geometry can be observed on the second row from the top\ignorethis{top row} as Latent-Paint keeps the input geometry fixed). As illustrated in the figure, Text2Mesh cannot produce significant geometric edits (\emph{e.g.}, adding a Santa hat to the horse or turning the horse into a donkey). Even SketchShape, which is designed to allow geometric edits, cannot achieve significant localized edits. Furthermore, it fails to preserve the geometry of the input---although, we again note that this method is not intended to preserve the input geometry. 
DFF+CN seems generally less suitable for our problem setting, particularly for prompts that require geometric modifications (i.e. ``A donkey'').
Our method, in contrast to prior works, succeeds in conforming to the target text prompt, while preserving the input geometry, allowing for semantically meaningful changes to both geometry and appearance.

We perform a quantitative evaluation in Table \ref{tab:baseline-stats} on our dataset. To perform a fair comparison where all methods operate within their training domain, we use meshes without texture maps as input for Text2Mesh and Latent-NeRF. As illustrated in the table, our method outperforms all baselines over both local and global edits in terms of CLIP similarity, but Text2Mesh yields slightly higher CLIP direction similarity. We note that Text2Mesh as well as DFF+CN are advantaged in terms of the CLIP metrics as they explicitly optimize on CLIP similarities and thus their scores are not entirely indicative. 

\subsection{2D Image Editing Comparisons}
\label{sec:comp2D}
An underlying assumption in our work is that editing 3D geometry cannot easily be done by reconstructing edited 2D images depicting the scene. To test this hypothesis, we modified images rendered from various viewpoints using the diffusion-based image editing methods InstructPix2Pix \cite{brooks2022instructpix2pix} and SDEdit \cite{meng2022sdedit}. We show two variants of these methods in Figure \ref{fig:comparison2d}, one with added backgrounds, as we observe that it also affects performance. 
In both cases, as illustrated in the figure, 2D methods often struggle to produce meaningful results from less canonical views (e.g., adding sunglasses on the dog’s back) and also produce highly view-inconsistent results. Concurrently to us, Instruct-NeRf2NeRF~\cite{haque2023instruct} explore how to best use these 2D methods to learn view-consistent 3D representations.

\begin{figure} %
\centering 
\jsubfig{\includegraphics[height=1.860cm,trim={1.2cm 0.8cm 1.2cm 1.6cm},clip] {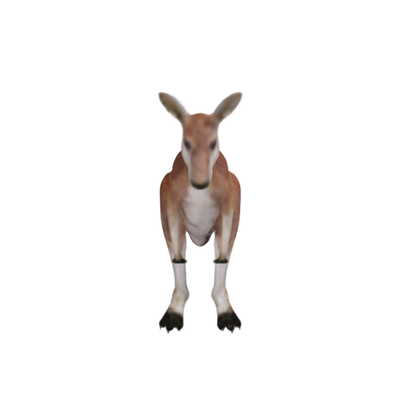}}{\footnotesize {Input}}\hfill
\jsubfig{\includegraphics[height=1.860cm]{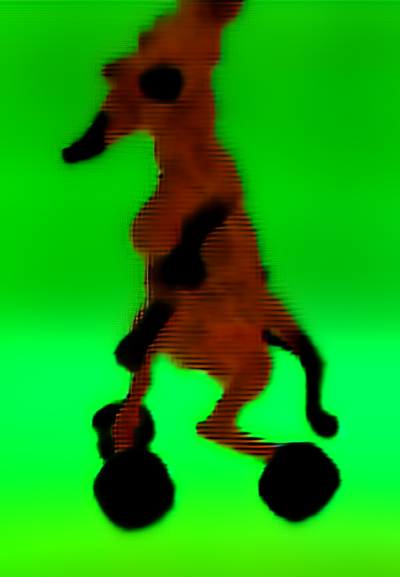}\includegraphics[height=1.860cm,trim={1.2cm 0.8cm 1.2cm 1.6cm},clip]{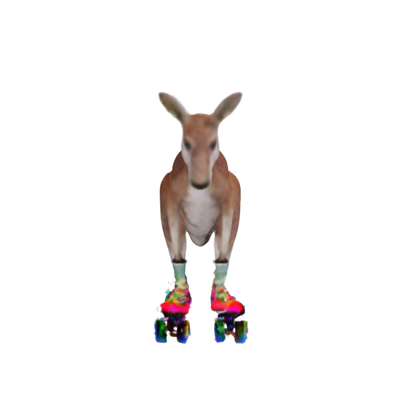}}{\footnotesize {``Kangaroo on rollerskates"}} \hfill
\jsubfig{\includegraphics[height=1.860cm]{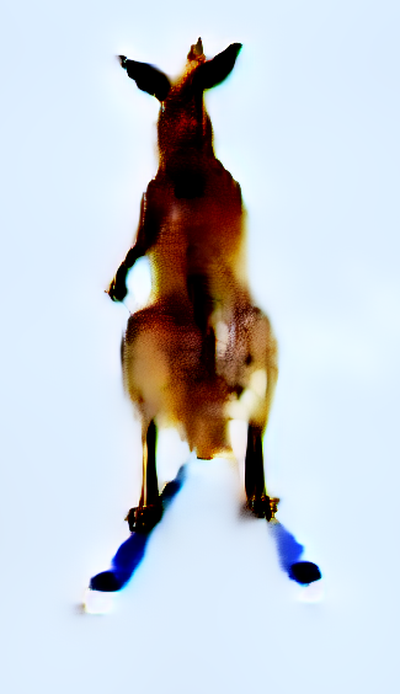}\includegraphics[height=1.860cm,trim={0cm 0cm 0cm 1.75cm},clip]{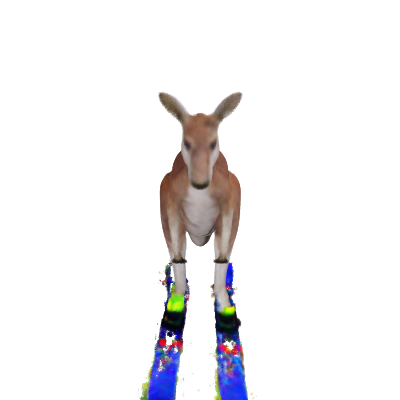}}{\footnotesize {``Kangaroo on skis"}}
\vspace{-1pt} 
\caption{\textbf{Comparison to unconditional text-to-3D generation}. We compare to unconditional text-to-3D methods by comparing to Latent-NeRF ~\cite{metzer2022latent}, providing it with the two target prompts displayed above. We display these alongside our results (LatentNeRF on the left, ours on the right). As illustrated above, unconditional methods cannot easily match an input object, and are also not guaranteed to generate a consistent object over different prompts.  }
\label{fig:t23dablation}
\end{figure}

\subsection{Comparisons to an unconditional text-to-3D model}
\label{sec:comp_uncond}
In Figure \ref{fig:t23dablation} we compare to the unconditional text-to-3D model proposed in Latent-NeRF, to show that such unconditional models are also not guaranteed to generate a consistent object over different prompts. We also note that this result (as well as our edits) would certainly look better if fueled with a proprietary big diffusion model \cite{saharia2022photorealistic}, but nonetheless, these models cannot preserve identity.

\subsection{Real Scenes}
\label{sec:realscenes}
In Figure \ref{fig:supp_realscenes}, we demonstrate that our method also succeeds in modeling and editing real scenes using the  $360^\circ$ \emph{Real Scenes} available by Mildenhall et al.~\cite{mildenhall2021nerf}. As illustrated in the figure, we can locally edit the foreground (e.g., turning the pinecone into a pineapple) as well as globally edit the scene (e.g. turning the scene into a Van-Gogh painting). For these more complex and computationally demanding scenes, we also experiment with implementing our method on top of DVGO \cite{sun2022direct} (bottom row), in addition to ReLU-Fields which we exclusively focus on in all other experiments (top row), as it offers additional features such as scene contraction, a more expressive color feature space and complex ray sampling. These make this underlying representation better suited for editing and reconstructing these real scenes (as illustrated in the columns labeled as 'Initial'). This experiment also demonstrates that our method is agnostic to the underlying 3D representation of the scene and can readily operate over different grid-based representations.
\ignorethis{
As illustrated in the figure, our approach can locally edit the foreground (e.g., turning the flowers into sunflowers) or the background (e.g. turning the ground into a pond). \rev{For these real scenes, our method is applied on top of DVGO~\cite{sun2022direct}, which is better suited for reconstructing more complex scenes (we compare to results obtained with ReLU-Fields in the supplementary material). This experiment also demonstrates that our method is agnostic to the underlying 3D representation of the scene and can readily operate over different grid-based representations.
}
}

\ignorethis{

}

\ignorethis{
\subsection{2D Image Editing Comparisons}

An underlying assumption in our work is that editing 3D geometry cannot easily be done by reconstructing edited 2D images depicting the scene. To test this hypothesis, we modified images rendered from various viewpoints using the diffusion-based image editing methods InstructPix2Pix~\cite{brooks2022instructpix2pix} and SDEdit~\cite{meng2022sdedit}. 
As illustrated in Figure~\ref{fig:comparison2d}, 2D methods often struggle to produce meaningful results from less \emph{canonical} views (\emph{e.g.}, adding sunglasses on the dog's back) and also produce highly view-inconsistent results. 
}

\subsection{Ablations}
\label{sec:ablations}
\begin{figure} %
\centering 
\ignorethis{
\jsubfig{\includegraphics[height=1.8cm,trim={8.5cm 7.0cm 10.0cm 10.0cm},clip]{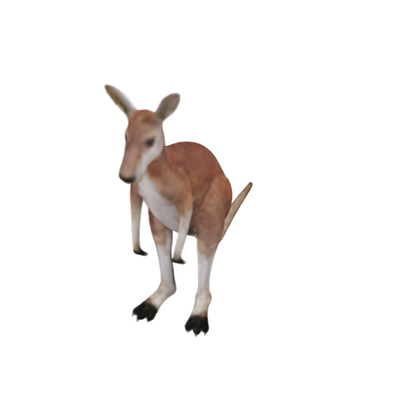}}{}\hspace{9pt}
\jsubfig{\includegraphics[height=1.8cm]{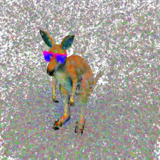}}{} \hfill 
\jsubfig{\includegraphics[height=1.8cm,trim={10.0cm 7.0cm 10.0cm 10.0cm},clip] {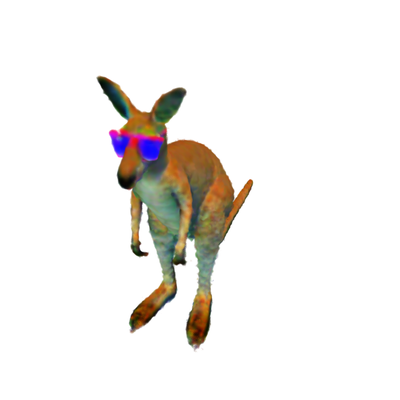}}{} \hfill
\jsubfig{\includegraphics[height=1.8cm,trim={10.0cm 7.0cm 10.0cm 10.0cm},clip]{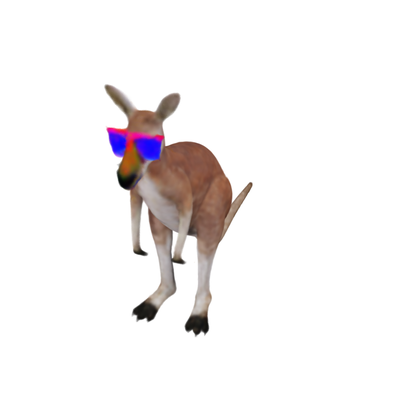}}{}
\\ 
\jsubfig{\includegraphics[height=1.8cm,trim={10.0cm 9.0cm 10.0cm 11.0cm},clip] {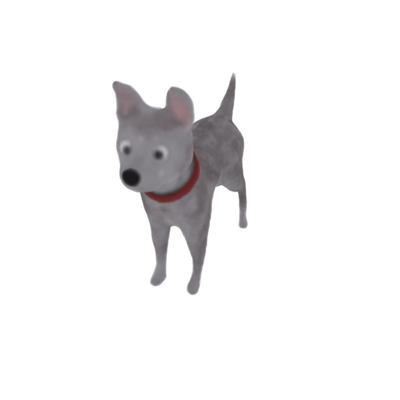}}{}\hfill
\jsubfig{\includegraphics[height=1.8cm]{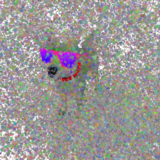}}{} \hfill 
\jsubfig{\includegraphics[height=1.8cm,trim={10.0cm 9.0cm 10.0cm 11.0cm},clip] {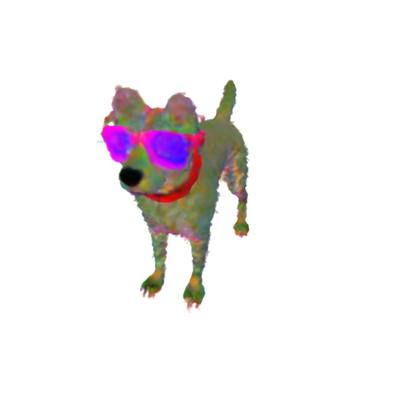}}{} \hfill
\jsubfig{\includegraphics[height=1.8cm,trim={10.0cm 9.0cm 10.0cm 11.0cm},clip]{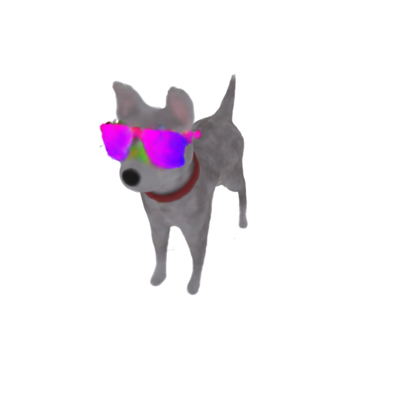}}{}
\\
\jsubfig{\includegraphics[height=1.8cm,trim={10.0cm 9.0cm 10.0cm 11.0cm},clip]{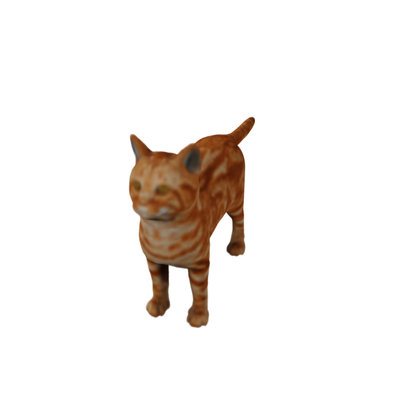}}{Input}\hfill
\jsubfig{\includegraphics[height=1.8cm]{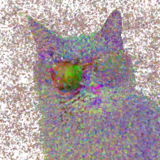}}{w/o $\mathcal{L}_{reg3D}$} \hfill 
\jsubfig{\includegraphics[height=1.8cm,trim={10.0cm 9.0cm 10.0cm 11.0cm},clip] {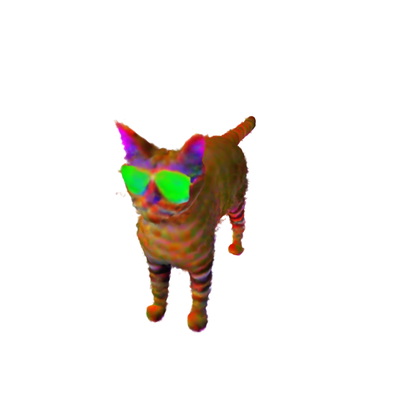}}{w/o SR} \hfill
\jsubfig{\includegraphics[height=1.8cm,trim={10.0cm 9.0cm 10.0cm 11.0cm},clip]{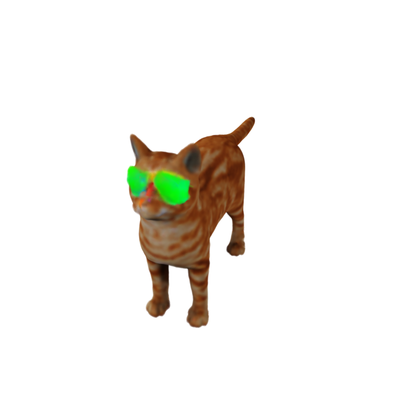}}{Output}
}

\jsubfig{\includegraphics[height=1.8cm,trim={2.125cm 1.75cm 2.5cm 2.5cm},clip]{images/ablations/kangaroo_input.png}}{}\hspace{9pt}
\jsubfig{\includegraphics[height=1.8cm]{images/ablations/kangaroo_uncoupled.png}}{} \hfill 
\jsubfig{\includegraphics[height=1.8cm,trim={2.5cm 1.75cm 2.5cm 2.5cm},clip] {images/ablations/kangaroo_unrefined.png}}{} \hfill
\jsubfig{\includegraphics[height=1.8cm,trim={2.5cm 1.75cm 2.5cm 2.5cm},clip]{images/ablations/kangaroo_refined.png}}{}
\\ 
\jsubfig{\includegraphics[height=1.8cm,trim={2.5cm 2.25cm 2.5cm 2.75cm},clip] {images/ablations/dog_input.png}}{}\hfill
\jsubfig{\includegraphics[height=1.8cm]{images/ablations/dog_uncoupled.png}}{} \hfill 
\jsubfig{\includegraphics[height=1.8cm,trim={2.5cm 2.25cm 2.5cm 2.75cm},clip] {images/ablations/dog_unrefined.png}}{} \hfill
\jsubfig{\includegraphics[height=1.8cm,trim={2.5cm 2.25cm 2.5cm 2.75cm},clip]{images/ablations/dog_refined.png}}{}
\\
\jsubfig{\includegraphics[height=1.8cm,trim={2.5cm 2.25cm 2.5cm 2.75cm},clip]{images/ablations/cat_input.png}}{Input}\hfill
\jsubfig{\includegraphics[height=1.8cm]{images/ablations/cat_uncoupled.png}}{w/o $\mathcal{L}_{reg3D}$} \hfill 
\jsubfig{\includegraphics[height=1.8cm,trim={2.5cm 2.25cm 2.5cm 2.75cm},clip] {images/ablations/cat_unrefined.png}}{w/o SR} \hfill
\jsubfig{\includegraphics[height=1.8cm,trim={2.5cm 2.25cm 2.5cm 2.75cm},clip]{images/ablations/cat_refined.png}}{Output}

\vspace{5pt} 
\caption{\textbf{Qualitative ablation results}, obtained for the target prompt ``A <object> wearing sunglasses" over three different objects. Image-space regularization (denoted by ``w/o $\mathcal{L}_\text{reg3D}$") leads to extremely noisy results. The edited grid before refinement (denoted by ``w/o SR") respects the target prompt, but some of the fidelity to the geometry and appearance of the input object is lost. In contrast, our refined grid successfully combines the edited and input regions to output a result that complies with the target text and also preserves the input object.
}
\label{fig:ablations}
\end{figure}

\begin{table}[t]
\centering
\resizebox{\linewidth}{!}{
\begin{tabular}{ll|ccccc}
\hline
 $\mathcal{L}_\text{reg3D}$ & SR  & $\text{CLIP}_{Sim}\uparrow$ & $\text{CLIP}_{Dir}\uparrow$ & $\text{FID}_{Rec}\downarrow$  & $\text{FID}_{Input}\downarrow$ \\
\hline
$\times$ & $\times$ & 0.29 & 0.05 & 367.53 & 384.55 \\
\checkmark & $\times$  & \textbf{0.37} & \textbf{0.08} & 240.37 & 288.26  \\ 
\checkmark& \checkmark & 0.36
  & 0.06 & \textbf{119.44}  & \textbf{236.32} \\  
\hline
\end{tabular}
}
\caption{\textbf{Ablation study}, evaluating the effect of the volumetric regularizer between our coupled grids ($\mathcal{L}_\text{reg3D}$, Section~\ref{sec:volume_reg}) and the 3D cross-attention-based spatial refinement module (SR, Section~\ref{sec:refine}) over a set of metrics (detailed in Section \ref{sec:results}).}
\label{tab:ablations}
\end{table}
We provide an ablation study in Table~\ref{tab:ablations} and Figure~\ref{fig:ablations}. Specifically, we ablate our volumetric regularization ($\mathcal{L}_{reg3D}$) and our 3D cross-attention-based spatial refinement module (SR). When ablating our volumetric regularization, we use a single volumetric grid and regularize the SDS objective with an image-based L2 regularization loss. More details and additional ablations are provided in the supplementary material, including alternative regularization objectives (such as image-based L1 loss, or volumetric regularization over RGB features) and results using higher order spherical harmonics
coefficients.

The benefit of using our volumetric regularization is further illustrated in Figure~\ref{fig:ablations}, which shows that image-space regularization leads to very noisy results, and often complete failures (see, for instance, the cat result, where the output is not at all correlated with the input object). Quantitatively, we can also observe that images rendered from these models are of significantly different appearance (as measured using the FID metrics). 

Regarding the SR module, as expected, it increases  similarity to the inputs (reflected in lower FID scores). This is also clearly visible in Figure \ref{fig:ablations}---for example, geometric differences are apparent by looking at the animals' paws. The output textures after refinement also are more similar to the input textures. However, we also see that this module slightly hinders CLIP similarity to the edit and text prompt. This is also somewhat expected as we are further constraining the output to stay similar to the input, sometimes at the expense of the editing signal. 



\subsection{Limitations}
\label{sec:limitations}
\begin{figure} %
\ignorethis{
\centering 
\jsubfig{\includegraphics[height=2.2cm,trim={0.0cm 3.0cm 1.0cm 3.0cm},clip]{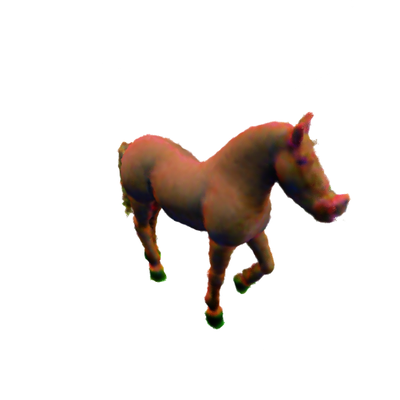}}{\footnotesize {``A horse with a pig tail"}}\hfill
\jsubfig{\includegraphics[height=2.2cm,trim={2.0cm 3.0cm 1.0cm 3.0cm},clip]{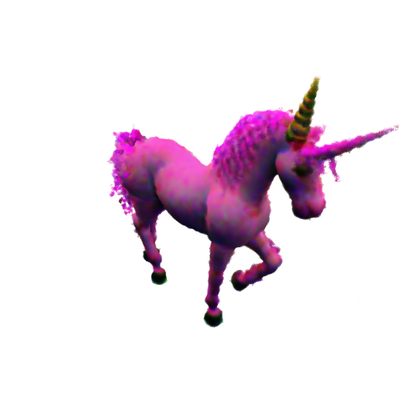}}{\footnotesize {``A pink unicorn"}}\hfill
\jsubfig{\includegraphics[height=2.2cm,trim={2.0cm 3.0cm 1.0cm 3.0cm},clip]{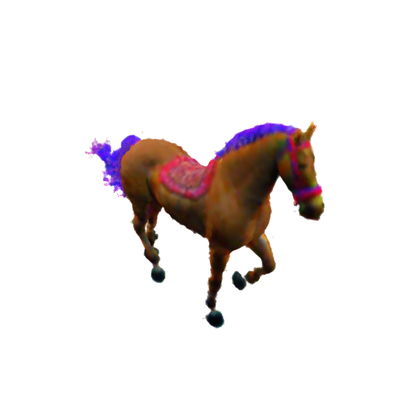}}{\footnotesize {``A horse riding on a magic carpet"}}
}
\centering 
\jsubfig{\includegraphics[height=2.2cm,trim={1.6cm 2.4cm 0.8cm 2.4cm},clip]{images/limitations/pig.png}}{\footnotesize {``A horse with a pig tail"}}\hfill
\jsubfig{\includegraphics[height=2.2cm,trim={1.6cm 2.4cm 0.8cm 2.4cm},clip]{images/limitations/unicorn2.png}}{\footnotesize {``A pink unicorn"}}\hfill
\jsubfig{\includegraphics[height=2.2cm,trim={1.6cm 2.4cm 0.8cm 2.4cm},clip]{images/limitations/carpet2.png}}{\footnotesize {``A horse riding on a magic carpet"}}

\vspace{1.5pt}
\caption{\textbf{Limitations}. Above, we present several failure cases (when provided with rendered images of the uncolored mesh displayed in Figure \ref{fig:comparisons}, top row). These likely result from incorrect attribute binding (the horse's nose turning into a pig's nose), inconsistencies across views (two horns on the unicorn) or excessive regularization to the input object (carpet on the horse, not below).
}
\label{fig:limitations}
\end{figure}

Our method applies a wide range of edits with high fidelity to 3D objects, however, there are several limitations to consider. As shown in Figure \ref{fig:limitations}, since we optimize over different views, our method attempts to edit the same object in differing spatial locations, thus failing on certain prompts. Moreover, the figure shows that some of our edits fail due to incorrect attribute binding, where the model binds attributes to the wrong subjects, which is a common challenge in large-scale diffusion-based models \cite{chefer2023attend}. 
Finally, we inherit the limitations of our volumetric representation. Thus, the quality of real scenes, for instance, could be significantly improved by borrowing ideas from works such as \cite{barron2022mip} (e.g. scene contraction to model the background).

\section{Conclusion}

In this work, we presented Vox-E, a new framework that leverages the expressive power of diffusion models for text-guided voxel editing of 3D objects.
Technically, we demonstrated that by combining a diffusion-based image-space objective with volumetric regularization we can achieve fidelity to the target prompt and to the input 3D object. We also illustrated that 2D cross-attention maps can be elevated for performing localization in 3D space.
We showed that our approach can generate both local and global edits, which are challenging for existing techniques. 
Our work makes it easy for non-experts to modify 3D objects using just text prompts as input, bringing us closer to the goal of democratizing 3D content creating and editing.
\section{Acknowledgments}
We thank Rinon Gal, Gal Metzer and Elad Richardson for their insightful feedback. This work was supported by a research gift from Meta, the Alon Fellowship and the Yandex Initiative in AI.

{\small
\bibliographystyle{ieee_fullname}
\bibliography{egbib}

\begin{thebibliography}{10}\itemsep=-1pt

\bibitem{styleflow}
Rameen Abdal, Peihao Zhu, Niloy~J. Mitra, and Peter Wonka.
\newblock Styleflow: Attribute-conditioned exploration of stylegan-generated
  images using conditional continuous normalizing flows.
\newblock {\em ACM Trans. Graph.}, 40(3), May 2021.

\bibitem{achlioptas2022changeit3d}
Panos Achlioptas, Ian Huang, Minhyuk Sung, Sergey Tulyakov, and Leonidas
  Guibas.
\newblock Changeit3d: Language-assisted 3d shape edits and deformations, 2022.

\bibitem{agarwala2004interactive}
Aseem Agarwala, Mira Dontcheva, Maneesh Agrawala, Steven Drucker, Alex Colburn,
  Brian Curless, David Salesin, and Michael Cohen.
\newblock Interactive digital photomontage.
\newblock {\em ACM Trans. Graph.}, 23(3):294–302, 2004.

\bibitem{avrahami2022blended}
Omri Avrahami, Dani Lischinski, and Ohad Fried.
\newblock Blended diffusion for text-driven editing of natural images.
\newblock In {\em Proceedings of the IEEE/CVF Conference on Computer Vision and
  Pattern Recognition}, pages 18208--18218, 2022.

\bibitem{bar2022text2live}
Omer Bar-Tal, Dolev Ofri-Amar, Rafail Fridman, Yoni Kasten, and Tali Dekel.
\newblock Text2live: Text-driven layered image and video editing.
\newblock In {\em ECCV}, 2022.

\bibitem{barron2022mip}
Jonathan~T Barron, Ben Mildenhall, Dor Verbin, Pratul~P Srinivasan, and Peter
  Hedman.
\newblock Mip-nerf 360: Unbounded anti-aliased neural radiance fields.
\newblock In {\em Proceedings of the IEEE/CVF Conference on Computer Vision and
  Pattern Recognition}, pages 5470--5479, 2022.

\bibitem{boykov2001fast}
Yuri Boykov, Olga Veksler, and Ramin Zabih.
\newblock Fast approximate energy minimization via graph cuts.
\newblock {\em IEEE Transactions on pattern analysis and machine intelligence},
  23(11):1222--1239, 2001.

\bibitem{brooks2022instructpix2pix}
Tim Brooks, Aleksander Holynski, and Alexei~A. Efros.
\newblock Instructpix2pix: Learning to follow image editing instructions.
\newblock November 2022.

\bibitem{chefer2023attend}
Hila Chefer, Yuval Alaluf, Yael Vinker, Lior Wolf, and Daniel Cohen-Or.
\newblock Attend-and-excite: Attention-based semantic guidance for
  text-to-image diffusion models.
\newblock {\em arXiv preprint arXiv:2301.13826}, 2023.

\bibitem{chen2022tango}
Yongwei Chen, Rui Chen, Jiabao Lei, Yabin Zhang, and Kui Jia.
\newblock Tango: Text-driven photorealistic and robust 3d stylization via
  lighting decomposition.
\newblock {\em arXiv preprint arXiv:2210.11277}, 2022.

\bibitem{couairon2022diffedit}
Guillaume Couairon, Jakob Verbeek, Holger Schwenk, and Matthieu Cord.
\newblock Diffedit: Diffusion-based semantic image editing with mask guidance.
\newblock {\em arXiv preprint arXiv:2210.11427}, 2022.

\bibitem{gal2021stylegannada}
Rinon Gal, Or Patashnik, Haggai Maron, Gal Chechik, and Daniel Cohen-Or.
\newblock Stylegan-nada: Clip-guided domain adaptation of image generators,
  2021.

\bibitem{garbin2022voltemorph}
Stephan~J Garbin, Marek Kowalski, Virginia Estellers, Stanislaw Szymanowicz,
  Shideh Rezaeifar, Jingjing Shen, Matthew Johnson, and Julien Valentin.
\newblock Voltemorph: Realtime, controllable and generalisable animation of
  volumetric representations.
\newblock {\em arXiv preprint arXiv:2208.00949}, 2022.

\bibitem{haque2023instruct}
Ayaan Haque, Matthew Tancik, Alexei~A Efros, Aleksander Holynski, and Angjoo
  Kanazawa.
\newblock Instruct-nerf2nerf: Editing 3d scenes with instructions.
\newblock {\em arXiv preprint arXiv:2303.12789}, 2023.

\bibitem{hertz2022prompt}
Amir Hertz, Ron Mokady, Jay Tenenbaum, Kfir Aberman, Yael Pritch, and Daniel
  Cohen-Or.
\newblock Prompt-to-prompt image editing with cross attention control.
\newblock 2022.

\bibitem{Heusel2017GANsTB}
Martin Heusel, Hubert Ramsauer, Thomas Unterthiner, Bernhard Nessler, and Sepp
  Hochreiter.
\newblock Gans trained by a two time-scale update rule converge to a local nash
  equilibrium.
\newblock In {\em NeurIPS}, 2017.

\bibitem{ho2020denoising}
Jonathan Ho, Ajay Jain, and Pieter Abbeel.
\newblock Denoising diffusion probabilistic models.
\newblock {\em NeurIPS}, 2020.

\bibitem{huang2022ladis}
Ian Huang, Panos Achlioptas, Tianyi Zhang, Sergey Tulyakov, Minhyuk Sung, and
  Leonidas Guibas.
\newblock Ladis: Language disentanglement for 3d shape editing.
\newblock {\em arXiv preprint arXiv:2212.05011}, 2022.

\bibitem{igarashi2005rigid}
Takeo Igarashi, Tomer Moscovich, and John~F Hughes.
\newblock As-rigid-as-possible shape manipulation.
\newblock {\em ACM transactions on Graphics (TOG)}, 24(3):1134--1141, 2005.

\bibitem{jain2022zero}
Ajay Jain, Ben Mildenhall, Jonathan~T Barron, Pieter Abbeel, and Ben Poole.
\newblock Zero-shot text-guided object generation with dream fields.
\newblock In {\em CVPR}, 2022.

\bibitem{karnewar2022relu}
Animesh Karnewar, Tobias Ritschel, Oliver Wang, and Niloy Mitra.
\newblock Relu fields: The little non-linearity that could.
\newblock In {\em ACM SIGGRAPH 2022 Conference Proceedings}, pages 1--9, 2022.

\bibitem{kawar2022imagic}
Bahjat Kawar, Shiran Zada, Oran Lang, Omer Tov, Huiwen Chang, Tali Dekel, Inbar
  Mosseri, and Michal Irani.
\newblock Imagic: Text-based real image editing with diffusion models.
\newblock {\em arXiv preprint arXiv:2210.09276}, 2022.

\bibitem{adamoptimizer}
Diederik~P. Kingma and Jimmy Ba.
\newblock Adam: A method for stochastic optimization.
\newblock {\em CoRR}, abs/1412.6980, 2014.

\bibitem{kobayashi2022decomposing}
Sosuke Kobayashi, Eiichi Matsumoto, and Vincent Sitzmann.
\newblock Decomposing nerf for editing via feature field distillation.
\newblock {\em Advances in Neural Information Processing Systems},
  35:23311--23330, 2022.

\bibitem{lin2022magic3d}
Chen-Hsuan Lin, Jun Gao, Luming Tang, Towaki Takikawa, Xiaohui Zeng, Xun Huang,
  Karsten Kreis, Sanja Fidler, Ming-Yu Liu, and Tsung-Yi Lin.
\newblock Magic3d: High-resolution text-to-3d content creation.
\newblock {\em arXiv preprint arXiv:2211.10440}, 2022.

\bibitem{liu2021editing}
Steven Liu, Xiuming Zhang, Zhoutong Zhang, Richard Zhang, Jun-Yan Zhu, and
  Bryan Russell.
\newblock Editing conditional radiance fields.
\newblock In {\em Proceedings of the IEEE/CVF International Conference on
  Computer Vision}, pages 5773--5783, 2021.

\bibitem{meng2022sdedit}
Chenlin Meng, Yutong He, Yang Song, Jiaming Song, Jiajun Wu, Jun-Yan Zhu, and
  Stefano Ermon.
\newblock {SDE}dit: Guided image synthesis and editing with stochastic
  differential equations.
\newblock In {\em International Conference on Learning Representations}, 2022.

\bibitem{metzer2022latent}
Gal Metzer, Elad Richardson, Or Patashnik, Raja Giryes, and Daniel Cohen-Or.
\newblock Latent-nerf for shape-guided generation of 3d shapes and textures.
\newblock {\em arXiv preprint arXiv:2211.07600}, 2022.

\bibitem{michel2022text2mesh}
Oscar Michel, Roi Bar-On, Richard Liu, Sagie Benaim, and Rana Hanocka.
\newblock Text2mesh: Text-driven neural stylization for meshes.
\newblock In {\em CVPR}, 2022.

\bibitem{mildenhall2021nerf}
Ben Mildenhall, Pratul~P Srinivasan, Matthew Tancik, Jonathan~T Barron, Ravi
  Ramamoorthi, and Ren Ng.
\newblock Nerf: Representing scenes as neural radiance fields for view
  synthesis.
\newblock {\em Communications of the ACM}, 65(1):99--106, 2021.

\bibitem{nichol2021glide}
Alexander~Quinn Nichol, Prafulla Dhariwal, Aditya Ramesh, Pranav Shyam, Pamela
  Mishkin, Bob McGrew, Ilya Sutskever, and Mark Chen.
\newblock {GLIDE:} towards photorealistic image generation and editing with
  text-guided diffusion models.
\newblock In {\em ICML}, 2022.

\bibitem{Parmar2023ZeroshotIT}
Gaurav Parmar, Krishna~Kumar Singh, Richard Zhang, Yijun Li, Jingwan Lu, and
  Jun-Yan Zhu.
\newblock Zero-shot image-to-image translation.
\newblock {\em ArXiv}, abs/2302.03027, 2023.

\bibitem{patashnik2021styleclip}
Or Patashnik, Zongze Wu, Eli Shechtman, Daniel Cohen-Or, and Dani Lischinski.
\newblock Styleclip: Text-driven manipulation of stylegan imagery.
\newblock In {\em ICCV}, 2021.

\bibitem{poole2022dreamfusion}
Ben Poole, Ajay Jain, Jonathan~T Barron, and Ben Mildenhall.
\newblock Dreamfusion: Text-to-3d using 2d diffusion.
\newblock {\em arXiv preprint arXiv:2209.14988}, 2022.

\bibitem{radford2021learning}
Alec Radford, Jong~Wook Kim, Chris Hallacy, Aditya Ramesh, Gabriel Goh,
  Sandhini Agarwal, Girish Sastry, Amanda Askell, Pamela Mishkin, Jack Clark,
  et~al.
\newblock Learning transferable visual models from natural language
  supervision.
\newblock In {\em ICML}, 2021.

\bibitem{richardson2023texture}
Elad Richardson, Gal Metzer, Yuval Alaluf, Raja Giryes, and Daniel Cohen-Or.
\newblock Texture: Text-guided texturing of 3d shapes.
\newblock {\em arXiv preprint arXiv:2302.01721}, 2023.

\bibitem{saharia2022photorealistic}
Chitwan Saharia, William Chan, Saurabh Saxena, Lala Li, Jay Whang, Emily
  Denton, Seyed Kamyar~Seyed Ghasemipour, Burcu~Karagol Ayan, S~Sara Mahdavi,
  Rapha~Gontijo Lopes, et~al.
\newblock Photorealistic text-to-image diffusion models with deep language
  understanding.
\newblock {\em arXiv preprint arXiv:2205.11487}, 2022.

\bibitem{sanghi2022clip}
Aditya Sanghi, Hang Chu, Joseph~G Lambourne, Ye Wang, Chin-Yi Cheng, Marco
  Fumero, and Kamal~Rahimi Malekshan.
\newblock Clip-forge: Towards zero-shot text-to-shape generation.
\newblock In {\em CVPR}, 2022.

\bibitem{Seitzer2020FID}
Maximilian Seitzer.
\newblock {pytorch-fid: FID Score for PyTorch}.
\newblock \url{https://github.com/mseitzer/pytorch-fid}, 08 2020.
\newblock Version 0.2.1.

\bibitem{shen2020interpreting}
Yujun Shen, Jinjin Gu, Xiaoou Tang, and Bolei Zhou.
\newblock Interpreting the latent space of gans for semantic face editing.
\newblock In {\em CVPR}, 2020.

\bibitem{sorkine2007rigid}
Olga Sorkine and Marc Alexa.
\newblock As-rigid-as-possible surface modeling.
\newblock In {\em Symposium on Geometry processing}, volume~4, pages 109--116,
  2007.

\bibitem{sun2022direct}
Cheng Sun, Min Sun, and Hwann-Tzong Chen.
\newblock Direct voxel grid optimization: Super-fast convergence for radiance
  fields reconstruction.
\newblock In {\em Proceedings of the IEEE/CVF Conference on Computer Vision and
  Pattern Recognition}, pages 5459--5469, 2022.

\bibitem{pnpDiffusion2022}
Narek Tumanyan, Michal Geyer, Shai Bagon, and Tali Dekel.
\newblock Plug-and-play diffusion features for text-driven image-to-image
  translation.
\newblock {\em arXiv preprint arXiv:2211.12572}, 2022.

\bibitem{wang2022clip}
Can Wang, Menglei Chai, Mingming He, Dongdong Chen, and Jing Liao.
\newblock Clip-nerf: Text-and-image driven manipulation of neural radiance
  fields.
\newblock In {\em Proceedings of the IEEE/CVF Conference on Computer Vision and
  Pattern Recognition}, pages 3835--3844, 2022.

\bibitem{wang2022nerf}
Can Wang, Ruixiang Jiang, Menglei Chai, Mingming He, Dongdong Chen, and Jing
  Liao.
\newblock Nerf-art: Text-driven neural radiance fields stylization.
\newblock {\em arXiv preprint arXiv:2212.08070}, 2022.

\bibitem{haochen2022score}
Haochen Wang, Xiaodan Du, Jiahao Li, Raymond~A. Yeh, and Greg Shakhnarovich.
\newblock Score jacobian chaining: Lifting pretrained 2d diffusion models for
  3d generation.
\newblock {\em ArXiv}, abs/2212.00774, 2022.

\bibitem{xiang2021neutex}
Fanbo Xiang, Zexiang Xu, Milos Hasan, Yannick Hold-Geoffroy, Kalyan Sunkavalli,
  and Hao Su.
\newblock Neutex: Neural texture mapping for volumetric neural rendering.
\newblock In {\em Proceedings of the IEEE/CVF Conference on Computer Vision and
  Pattern Recognition}, pages 7119--7128, 2021.

\bibitem{xu2022deforming}
Tianhan Xu and Tatsuya Harada.
\newblock Deforming radiance fields with cages.
\newblock In {\em Computer Vision--ECCV 2022: 17th European Conference, Tel
  Aviv, Israel, October 23--27, 2022, Proceedings, Part XXXIII}, pages
  159--175. Springer, 2022.

\bibitem{yang2022neumesh}
Bangbang Yang, Chong Bao, Junyi Zeng, Hujun Bao, Yinda Zhang, Zhaopeng Cui, and
  Guofeng Zhang.
\newblock Neumesh: Learning disentangled neural mesh-based implicit field for
  geometry and texture editing.
\newblock In {\em Computer Vision--ECCV 2022: 17th European Conference, Tel
  Aviv, Israel, October 23--27, 2022, Proceedings, Part XVI}, pages 597--614.
  Springer, 2022.

\bibitem{yuan2022nerf}
Yu-Jie Yuan, Yang-Tian Sun, Yu-Kun Lai, Yuewen Ma, Rongfei Jia, and Lin Gao.
\newblock Nerf-editing: geometry editing of neural radiance fields.
\newblock In {\em Proceedings of the IEEE/CVF Conference on Computer Vision and
  Pattern Recognition}, pages 18353--18364, 2022.

\bibitem{zhang2022arf}
Kai Zhang, Nick Kolkin, Sai Bi, Fujun Luan, Zexiang Xu, Eli Shechtman, and Noah
  Snavely.
\newblock Arf: Artistic radiance fields, 2022.

\end{thebibliography}
}

\clearpage
\appendix
{\LARGE\textbf{Appendix}}

\section{Additional Details}
\label{sec:details}

\subsection{Implementation Details}
\label{sec:imp}
Below we provide all the implementation details of our method, detailed in Section 3 in the main paper.

\subsubsection*{Grid-Based Volumetric Representation}
We use 100 images uniformly sampled from upper hemisphere poses along with corresponding camera intrinsic and extrinsic parameters to train our initial grid. We follow the standard ReLU Fields~\cite{karnewar2022relu} training process using their default settings aside from two modifications: 
\begin{enumerate}
    \item We change the result grid size from the standard $128^3$ to $160^3$ to increase the output render quality.
    \item As detailed in the main paper, we limit the order of spherical harmonics to be zero order only to avoid undesirable view-dependent effects (we further illustrate these effects in Section \ref{sec:sh}). 
\end{enumerate}

\subsubsection*{Text-guided Object Editing}
We perform 8000 training iterations during the object editing optimization stage. During each iteration, a random pose is uniformly sampled from an upper hemisphere and an image is rendered from our edited grid $G_e$ according to the sampled pose and the rendering process described in ReLU Fields \cite{karnewar2022relu}. Noise is then added to the rendered image according to the time-step sampled from the fitting distribution. 

We use an annealed SDS loss which gradually decreases the maximal time-step we draw $t$ from. Formally, this annealed SDS loss introduces three additional hyper-parameters to our system: a starting iteration $i_{start}$, an annealing frequency $f_a$ and an annealing factor $\gamma_a$. With these hyper-parameters set, we change our time-step distribution to be:
\begin{equation}
    t \sim U[t_0 + \varepsilon, t_{final}*k_i + \varepsilon],
\end{equation}
\begin{equation}
    k_i = 
    \begin{cases}
    1, & \text{if } i < i_{start} \\
    k_{i-1}*\gamma_a, & \text{else if } i\ \% \ f_a = 0 \\
    k_{i-1}, & \text{otherwise}
    \end{cases}
\end{equation}
In all our experiments, the values we use for $\varepsilon$, $i_{start}$, $f_a$ and $\gamma_a$ are 0.02, 4000, 600, and 0.75. Additionally, we stop annealing the time-step when it reaches a value of 0.35. 
The latent diffusion model we use in our experiments is "StableDiffusion 2.1" by Stability AI\href{https://huggingface.co/stabilityai/stable-diffusion-2-1}. 

We use a weight of $200$ to balance the two terms (multiplying $\mathcal{L}_\text{reg3D}$ by this weight value). The volumetric regularization term operates only on the density features of the editing grid. The optimizer we used in this (and all other stages) is the Adam optimizer~\cite{adamoptimizer} with a learning rate of 0.03 and betas 0.9, 0.999. The resolution of the images rendered from our grid is 266$\times$266. We add a "a render of" prefix to all of our editing prompts as we found that this produced more coherent results (and the images the LDM receives are indeed renders).


\subsubsection*{Spatial Refinement via 3D Cross-Attention}
The diffusion model we use for this stage is \href{ https://huggingface.co/CompVis/stable-diffusion-v1-4 }{"StableDiffusion 1.4" by CompVis} and it consists of several 
cross-attention layers at resolutions 32, 16, and 8. To extract a single attention map for each token we interpolate each cross attention map from each layer and attention head to our image resolution (266x266) and take an average per each token. 
The time-step we use to generate the attention maps is 0.2 (the actual step being 0.2 * $N_{steps}$ = 200). 

The cross-attention grids $A_{e}$ and $A_{obj}$ contain a density feature and an additional one-dimensional feature $a$, which represents the cross-attention value at a given voxel and can be interpreted and rendered as a grayscale luma value. We initialize the density features in these grids to the density features of the editing grid's (the former stage's output) and freeze them.
At each refinement iteration we generate two 2D cross-attention maps from the LDM, one for the object and one for the edit. After obtaining the 2D cross-attention maps, we render gray-scale heatmaps 
from $A_{e}$ and $A_{obj}$ and use $L1$ loss to encourage similarity between the rendered attention images and their corresponding attention maps extracted from the diffusion model. We repeat this process for 1500 iterations, sampling a random upper-hemisphere pose each time. As in the former optimization stage, we use the Adam optimizer with a learning rate of 0.03 and betas 0.9 and 0.999 and generate images in 266$\times$266 resolution.

After obtaining the two grids $A_{e}$ and $A_{obj}$, we perform element-wise softmax on their $a$ values to obtain probabilities for each voxel belonging to either the object, denoted by $P_{obj}(v)$, or the edit, denoted by $P_e(v)$. 
We then proceed to calculate the binary refinement volumetric mask. To do this we define a graph in which each non-zero density voxel in our edited grid $G_e$ is a node. We define "edit" and "object" labels as the \emph{source} and \emph{drain} nodes, such that a node connected to the source node is marked as an "edit" node and a node connected to the drain node is marked as an "object" node. We rank the nodes according to their $P_e(v)$ values and connect the top $N_{init-edit}$ nodes to the source node. We then rank the nodes according to their $P_{obj}(v)$ value and connect the top $N_{init-object}$ nodes to the drain node.
We then connect the non-terminal nodes to each-other in a 6-neighborhood with the capacity of each edge being $w_{pq}$ as detailed in the main paper.


We set the hyper-parameters $N_{init-edit}$ and $N_{init-object}$ to be 300 and 200. 
To perform graph-cut~\cite{boykov2001fast}, we used the \href{ https://github.com/pmneila/PyMaxflow }{PyMaxflow} implementation of the max-flow / min-cut algorithm. 

\subsection{Evaluation Protocol}
To evaluate our results quantitatively, we constructed a test set composed of eight scenes: 'White Dog', 'Grey Dog', 'White Cat', 'Ginger Cat', 'Kangaroo', 'Alien', 'Duck' and 'Horse', and six editing prompts: (1) A $\left<object\right>$ wearing big sunglasses, (2) A $\left<object\right>$ wearing a Christmas sweater, (3) A $\left<object\right>$ wearing a birthday party hat, (4) A yarn doll of a $\left<object\right>$, (5) A wood carving of a $\left<object\right>$, (6) A claymation $\left<object\right>$. This yields 18 edited scenes in total. We render each edited scene from 100 different poses distributed evenly along a $360^{\circ}$ ring. In addition to these 18 scenes we also render 100 images from the same poses on the initial (reconstruction) grid $G_i$ for each input scene. When comparing our result with other 3D textual editing papers we evaluate our results using two CLIP-based metrics. 
The CLIP model we used for both of these metrics is \href{https://github.com/openai/CLIP}{ViT-B/32} and the input image text prompts used to calculate the directional CLIP metric is ``A render of a $\left<object\right>$". $CLIP_{Dir}$ is calculated for each edited image in relation to the corresponding image in the reconstruction scene.
To quantitatively evaluate ablations we use two additional metrics using FID \cite{Seitzer2020FID}. For this we use the \href{ https://github.com/mseitzer/pytorch-fid }{pytorch implementation} given by the authors with the standard settings. 

\subsubsection*{$360^\circ$ \emph{Real Scenes}} For the $360^\circ$ \emph{Real Scenes} edits we follow the same implementation details as outlined previously, with three modifications:

\begin{enumerate}
    \item We alternate between using the DVGO model or the ReLU-Fields model as our 3D representation. Results for both models are presented in Figure 6 of the main paper.
    
    \item Our input poses are  created in a spherical manner and when rendering we sample linearly in inverse depth rather than in depth as seen in the official implementation of NeRF \href{https://github.com/bmild/nerf}.
    
    \item We perform 5000 training iterations during the object editing optimization stage and the values we use for $\varepsilon$, $i_{start}$, $f_a$ and $\gamma_a$ are 0.02, 3000, 400, and 0.75.
\end{enumerate}

\subsection{3D Object Editing Techniques} 
Below we provide additional details on the alternative 3D object editing techniques we compare against. All of the techniques we compare against use only an un-textured mesh and an editing prompt as input. As such, we used the meshes our inputs were rendered from as input for the editing methods. Additionally, we tested an additional scenario in which we imported the 'horse' mesh from the \href{https://github.com/threedle/text2mesh/blob/main/data/source_meshes/horse.obj}{Text2Mesh GitHub repository} to blender, added a grey-matte material to it and rendered images of it to use as input for our system. This scenario used four prompts: (1) A wood carving of a horse, (2) A horse wearing a Santa hat, (3) A donkey, (4) A carousel horse, and was used for qualitative comparisons only.

\subsubsection*{Text2Mesh} 
When comparing to Text2Mesh we used the \href{https://github.com/threedle/text2mesh}{code provided by the authors} and the input settings given in the "run\_horse.sh" demo file.



\subsubsection*{SketchShape}
 In this comparison we again use the \href{https://github.com/eladrich/latent-nerf}{code provided by the authors}. And the input parameters used are the default parameters in the 'train\_latent\_nerf.py' script  \href{https://github.com/eladrich/latent-nerf/tree/main/scripts}{'train\_latent\_nerf.py' script} with 10,000 training steps (as opposed to the default 5,000).

\subsubsection*{Latent-Paint}
We compared our method to Latent-Paint only qualitatively as this method outputs edits that transform only the appearance of the input mesh, rather than appearance and geometry. As in SketchShape we used the code provided by the authors and used the default input settings provided for latent paint, which are given in the  \href{https://github.com/eladrich/latent-nerf/tree/main/scripts}{'train\_latent\_paint.py' script}.

\subsubsection*{DFF + CN}
 In this comparison we use the \href{https://github.com/pfnet-research/distilled-feature-fields} {code provided by the authors} and the default input parameters provided for this method.

\subsection{2D Image Editing Techniques}

When comparing to InstructPix2Pix and SDEdit we constructed two image sets for each scene / prompt combination we wanted to test. Both sets were created by rendering one of our inputs in evenly spaced poses along a $360^{\circ}$ ring, one set was rendered over a white background and the other over a 'realistic' image of a brick wall. We used these sets as input for each 2D editing method along with an editing prompt and compared the results to rendered outputs from our result grids. When comparing to InstructPix2Pix we used the standard \href{https://huggingface.co/docs/diffusers/api/pipelines/stable_diffusion/pix2pix}{InstructPix2Pix pipeline} with 16bit floating point precision and 20 inference steps. We used the default guidance scale (1.0) for the images rendered over the 'realistic' background and increased the guidance scale to 3.0 for the images rendered over a white background, as we found it to produce higher quality results specifically for these more 'synthetic' images. When giving prompts to InstructPix2Pix we rephrased our prompts as instructions, for example turning "a dog wearing sunglasses" to "put sunglasses on this dog". When comparing to SDEdit we used the \href{https://huggingface.co/docs/diffusers/using-diffusers/img2img}{standard SDEdit pipeline} with guidance scale of 0.75 and a strength of 0.6.


\ignorethis{
\begin{enumerate}
    \item Real-scenes?
    \item cross-attention maps - \textbf{maybe a correction figure?} also, what timesteps are we using, any other important details?
    \item explain ablations and comparisons (all details needed to reproduce experiments)
    \item all hyperparameters
    \item Dataset details - the meshes, all the prompts
\end{enumerate}
}

\section{Ablations}
\label{sec:ab}
In this section, we show a more detailed ablation study which evaluates the effect of our volumetric regularization loss (Section \ref{sec:ablation-reg}) and an additional experiment, demonstrating the effect of using high order spherical harmonics coefficients (Section \ref{sec:sh}).

\subsection{Alternative Regularization Objectives}
\label{sec:ablation-reg}
Table \ref{tab:grid_losses} shows a quantitative comparison over different image-space and volumetric regularizations. Only the image-space L1 loss also appears in the main paper. 
Below we provide additional details on these ablations.

\begin{figure}[h!] %
\centering 
  \jsubfig{\includegraphics[height=1.9cm]{images/comparisons/image_editing/dog_white.png}}{}
\jsubfig{\includegraphics[height=1.9cm]{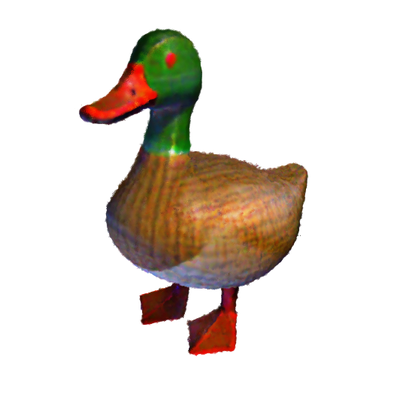}}{} 
\jsubfig{\includegraphics[height=1.9cm] {images/comparisons/image_editing/dog_white.png}}{} 
\jsubfig{\includegraphics[height=1.9cm]{images/ablations/fcl/dog_nofcl_cropped.png}}{}
\rotatebox[origin=tc]{-90}{$\mathcal{L}_{reg3D}$ \whitetxt{xxxxx}}
\vspace{-10pt}
\\ 
\jsubfig{\includegraphics[height=1.93cm]{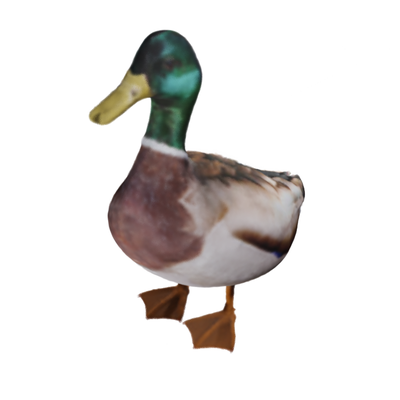}}{\footnotesize {Input}}
\jsubfig{\includegraphics[height=1.9cm]{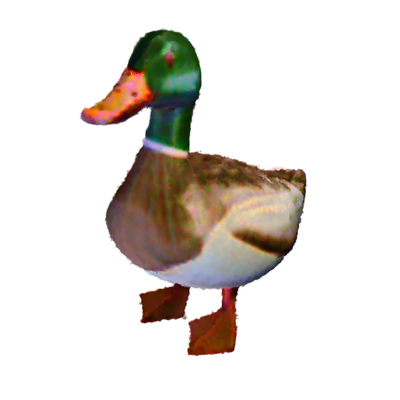}}{\footnotesize {''A wood carving of a duck"}} 
\jsubfig{\includegraphics[height=1.9cm] {images/ablations/fcl/dog_ref_cropped3.png}}{\footnotesize {Input}} 
\jsubfig{\includegraphics[height=1.9cm]{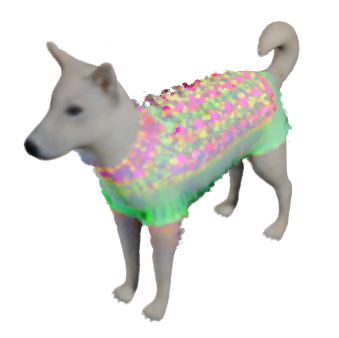}}{\footnotesize {''A dog wearing a christmas sweater"}}
\rotatebox[origin=tc]{-90}{$\mathcal{L}_{reg3D++}$ \whitetxt{xxxxxx}}
\vspace{3.2pt}

\caption{\textbf{Regularizing RGB colors in addition to volumetric densities}. We show results obtained when using our default regularization objective $\mathcal{L}_{reg3D}$ (top-row) compared against results obtained when using $\mathcal{L}_{reg3D++}$- an alternative version of $\mathcal{L}_{reg3D}$ (bottom-row) in which we penalize the miscorrelation between both density and color features. These results show that regularizing both density and RGB can be limiting, especially when the edit requires a drastic change in color, such as changing the white fur of the dog into a vibrant christmas sweater. }

\label{fig:supp_fcl}
\end{figure}

\paragraph{Image-space Regularization}
In this setting we render images from our editing grid $G_e$ in the poses corresponding to the input images during each iteration of the optimization stage. Rather than using a volumetric regularization, we incur a loss between the images rendered from $G_e$ and the corresponding input image while using the same weight used to balance $\mathcal{L}_\text{reg3D}$ with the annealed SDS loss (this weight is set to $200$, as detailed in Section \ref{sec:imp}). We evaluate this ablation using $L_1$ and $L_2$ image space loss functions. 

\begin{table}[t]
\centering
\resizebox{\linewidth}{!}{
\begin{tabular}{llcccc}
\toprule
 &Loss Function & $\text{CLIP}_{Sim}\uparrow$ & $\text{CLIP}_{Dir}\uparrow$ & $\text{FID}_{Rec}\downarrow$  & $\text{FID}_{Input}\downarrow$ \\
\midrule
\multirow{2}{*}{2D}&$L_1$ & 0.26 & 0.02 & 415.96 & 437.09  \\ 
&$L_2$ & 0.25 & 0.02 & 437.68 & 467.14 \\
\midrule
\multirow{3}{*}{3D}&$L_1$ & 0.36 & 0.05 & 222.91 & 284.86  \\ 
&$L_2$ & 0.35 & 0.05 & 240.50 & 284.83 \\
&$\mathcal{L}_{reg3D++}$ & 0.34 & 0.02 & \textbf{210.46} & \textbf{242.73} 
\\
&$\mathcal{L}_\text{reg3D}$ & \textbf{0.36} & \textbf{0.06} & 223.89 & 272.73 \\
\bottomrule
\end{tabular}
}
\\

\caption{\textbf{Detailed ablation study}, evaluating the effect of different  regularization objectives. We compare the performance using $\mathcal{L}_\text{reg3D}$, with image-space (top rows) and volumetric (bottom rows) $L_1$ and $L_2$ losses, as well as  $\mathcal{L}_\text{reg3D++}$, which also penalizes miscorrelations between color features.}
\label{tab:grid_losses}
\end{table}

\paragraph{Alternative Volumetric Regularization Functions}
In this setting we replace our correlation-based regularization with
other functions that encourage similarity between the density features of the grids using the same balancing weight. 
Namely we compare against $L1$ and $L2$ volumetric loss functions, both penalizing the distance between the density features of $G_i$ and those of $G_e$. We additionally compare against an alternative version of $\mathcal{L}_\text{reg3D}$ in which we penalize the miscorrelation between both density and color features, formally:

\begin{equation}
    \mathcal{L}_{reg3D++}  = \mathcal{L}_{reg3D} + (
1 - \frac{Cov(f^{rgb}_i, f^{rgb}_e)}
{\sqrt{Var(f^{rgb}_i)Var(f^{rgb}_e)}}
)
\end{equation}

We find that using this loss yields better reconstruction scores, at the expense of significantly lower CLIP-based scores (e.g., $\text{CLIP}_{Dir}$ scores drop from 0.08 to 0.02). Qualitatively, constraining RGB values as well as density features appears too limiting for our purposes. This can be seen in Figure \ref{fig:supp_fcl}, where we compare results obtained when using $\mathcal{L}_{reg3D++}$ against results obtained when using $\mathcal{L}_{reg3D}$. When observing these results, we can see that the edit integrity is reduced at the expense of the preservation of the origin object's color. This is evident in the duck, for instance, where the brown wooden color of the body is only clearly visible in the $\mathcal{L}_{reg3D}$ example. Furthermore, the colors of the sweater on the dog are significantly faded when regularized with $\mathcal{L}_{reg3D++}$ as the colors of a standard christmas sweater are typically much more vibrant than the white fur of the dog.

\begin{figure*} %
\centering
\rotatebox{90}{2D cross-attention}\hfill
\jsubfig{\includegraphics[height=2.72cm]{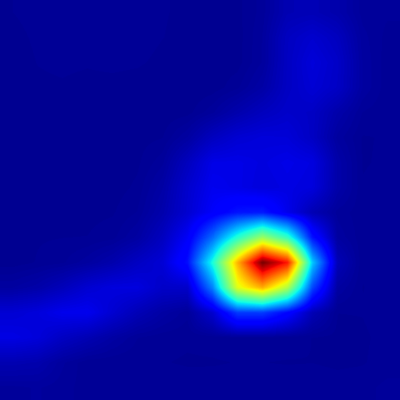}
\includegraphics[height=2.72cm]{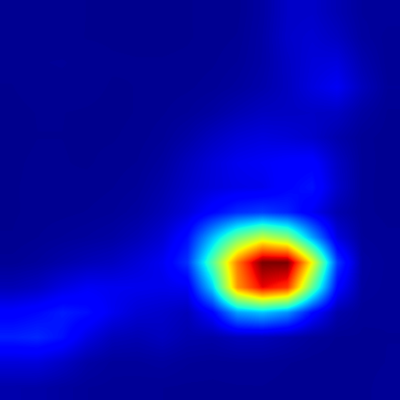}
\includegraphics[height=2.72cm]{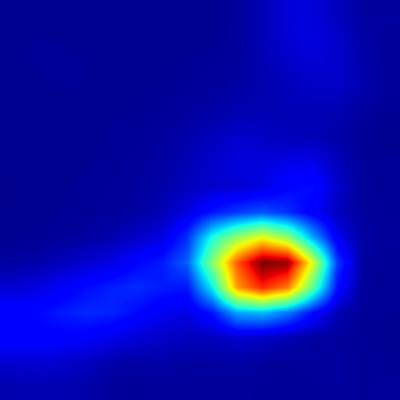}
\includegraphics[height=2.72cm]{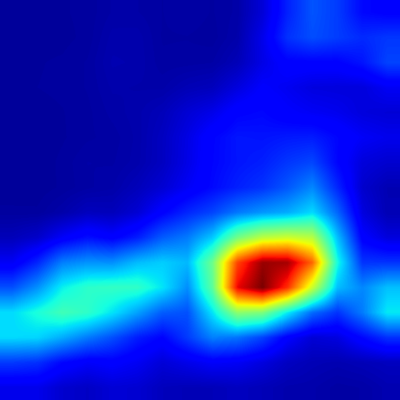}
\includegraphics[height=2.72cm]{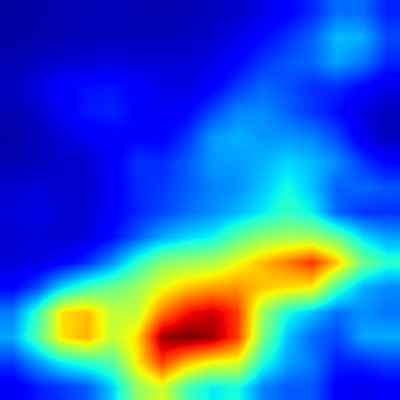}
\includegraphics[height=2.72cm]{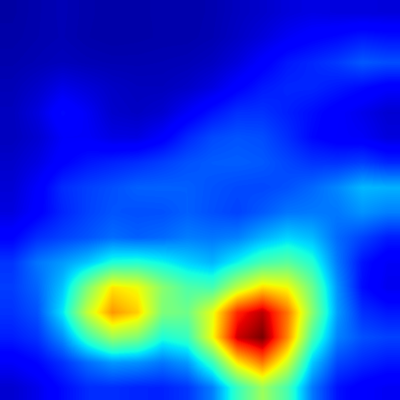}}{}
\\
\rotatebox{90}{\whitetxt{xx}3D grid ($A_e$)} \hfill 
\jsubfig{\includegraphics[height=2.72cm]{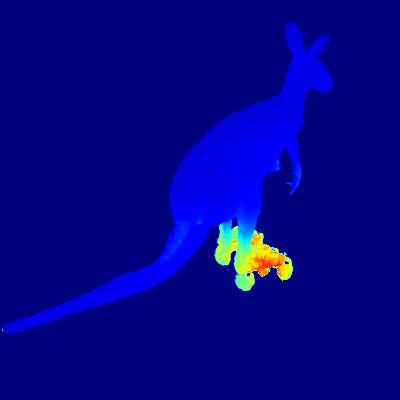}
\includegraphics[height=2.72cm]{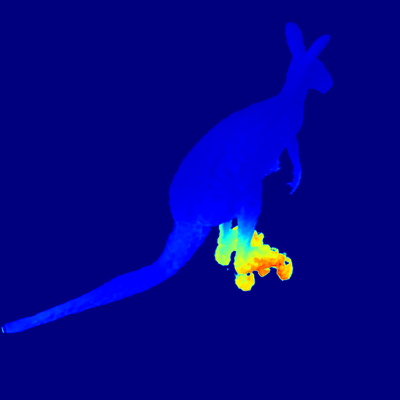}
\includegraphics[height=2.72cm]{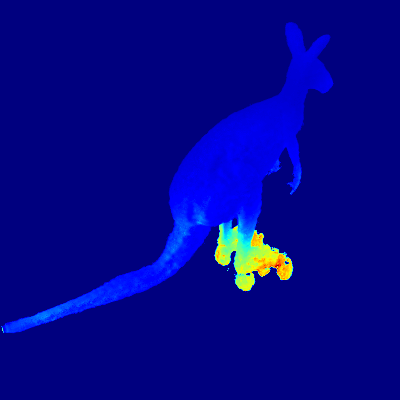}
\includegraphics[height=2.72cm]{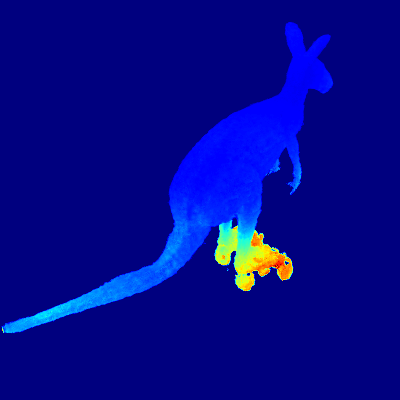}
\includegraphics[height=2.72cm]{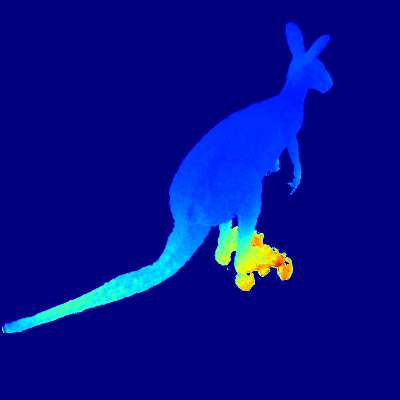}
\includegraphics[height=2.72cm]{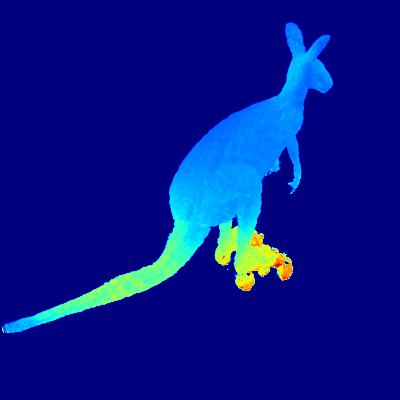}}{}
\\
\rotatebox{90}{2D cross-attention}\hfill
\jsubfig{\includegraphics[height=2.72cm]{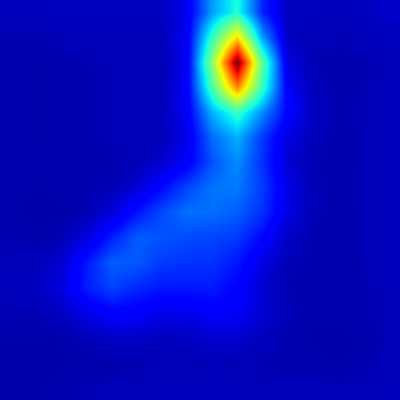}
\includegraphics[height=2.72cm]{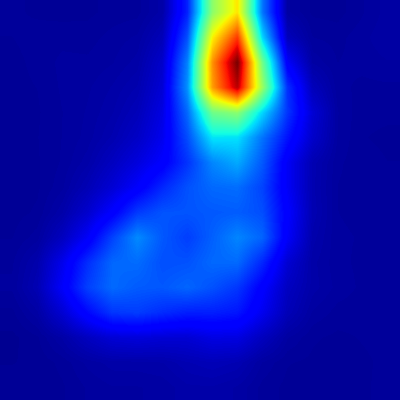}
\includegraphics[height=2.72cm]{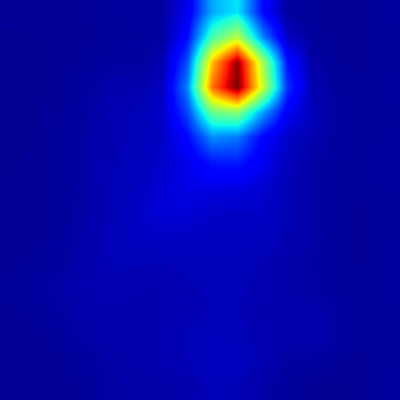}
\includegraphics[height=2.72cm]{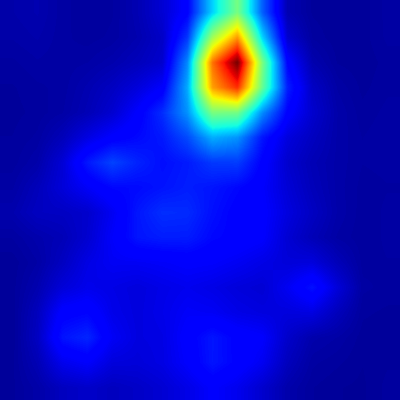}
\includegraphics[height=2.72cm]{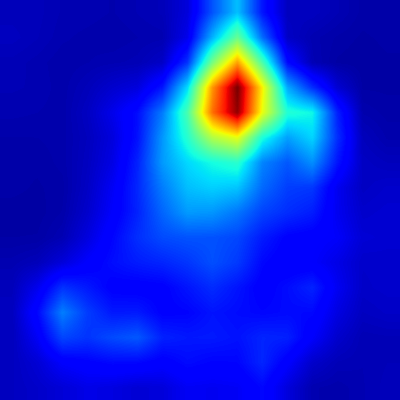}
\includegraphics[height=2.72cm]{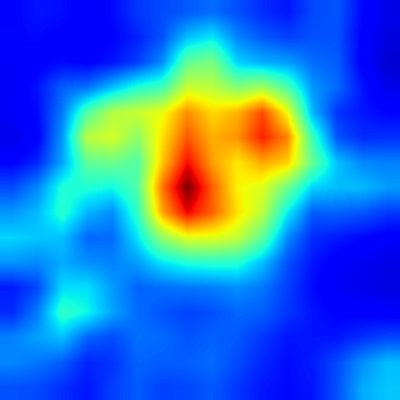}}{}
\\
\rotatebox{90}{\whitetxt{xx}3D grid ($A_e$)}\hfill \hspace{8pt}
\jsubfig{\includegraphics[height=2.72cm]{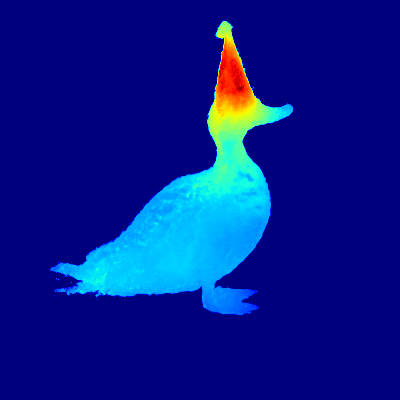}}{\footnotesize {t=1}} 
\jsubfig{\includegraphics[height=2.72cm]{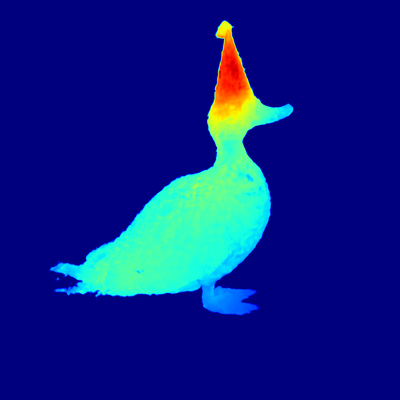}}{\footnotesize {t=200}} 
\jsubfig{\includegraphics[height=2.72cm]{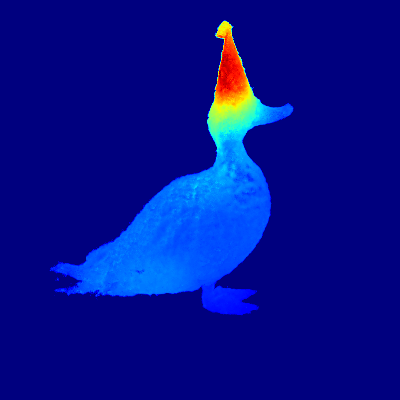}}{\footnotesize {t=400}} 
\jsubfig{\includegraphics[height=2.72cm]{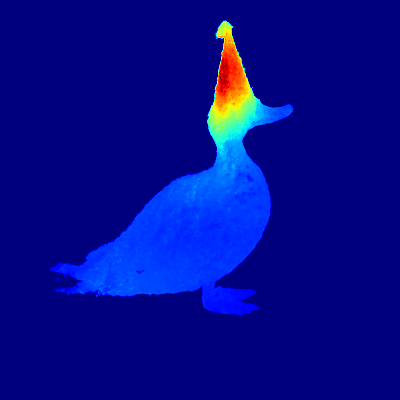}}{\footnotesize {t=600}}
\jsubfig{\includegraphics[height=2.72cm]{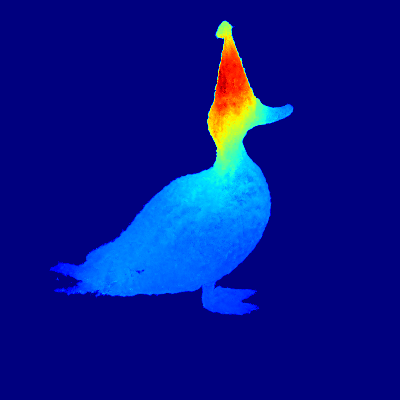}}{\footnotesize {t=800}}
\jsubfig{\includegraphics[height=2.72cm]{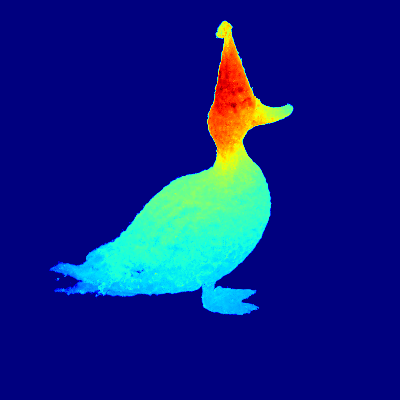}}{\footnotesize {t=999}}
\vspace{5pt} 
\caption{\textbf{Visualizing 2D cross-attention maps and 3d cross-attention grids over different diffusion timestamps}. We visualize the trained 3d cross-attention grids and the corresponding 2D cross-attention maps used as supervision across different diffusion timestamps.  We show them for the edit region corresponding to the token associated with the word ``rollerskates" (top two rows) and ``hat" (bottom two rows).}
\label{fig:supp_timestamps}
\end{figure*}

\ignorethis{
\begin{figure} %
\centering 
\jsubfig{\includegraphics[height=1.860cm,trim={1.2cm 0.8cm 1.2cm 1.6cm},clip] {images/t23dablation/input.png}}{\footnotesize {Input}}\hfill
\jsubfig{\includegraphics[height=1.860cm]{images/t23dablation/latent_nerf_skates_.png}\includegraphics[height=1.860cm,trim={1.2cm 0.8cm 1.2cm 1.6cm},clip]{images/t23dablation/ours_skates.png}}{\footnotesize {``Kangaroo on rollerskates"}} \hfill
\jsubfig{\includegraphics[height=1.860cm]{images/t23dablation/latent_nerf_skis_.png}\includegraphics[height=1.860cm,trim={0cm 0cm 0cm 1.75cm},clip]{images/t23dablation/ours_skis.png}}{\footnotesize {``Kangaroo on skis"}}
\vspace{-1pt} 
\caption{\textbf{Comparison to unconditional text-to-3D generation}. We compare to unconditional text-to-3D methods by comparing to Latent-NeRF ~\cite{metzer2022latent}, providing it with the two target prompts displayed above. We display these alongside our results (LatentNeRF on the left, ours on the right). As illustrated above, unconditional methods cannot easily match an input object, and are also not guaranteed to generate a consistent object over different prompts.  }
\label{fig:t23dablation}
\end{figure}

}

\subsection{Ablating the Color Representation}
\label{sec:sh}
As mentioned in Section 3.1 of the main paper, we  do not model view dependent effects using higher order spherical harmonics as that leads to undesirable effects. We demonstrate this by observing these effects in examples rendered with 1st and 2nd order spherical harmonic coefficients as color features. These results can be seen in videos available on our project page. 

When observing these results we can clearly see how view-dependent colors yield undesirable effects such as the feet of the ``yarn kangaroo" varying from green to yellow across views or the head of the dog becoming a birthday party hat when it faces away from the camera. We additionally see the colors become over-saturated, especially when using second-order spherical harmonic coefficients. It is also evident that the added expressive capabilities of the model allow it to over-fit more easily to specific views, creating unrealistic results such as the ``cat wearing glasses" in the first and second order coefficient models, where glasses are scattered along various parts of its body. We note that while this expressive power currently produces undesirable effects it does potentially enable higher quality and more realistic renders, and therefore, we believe that constraining this power is an interesting topic for future research.

\subsection{Cross-attention Grid Supervision}
As explained in Section \ref{sec:imp}, we use a constant time-stamp of 0.2 when extracting attention maps for training our attention grids $A_e$ and $A_{obj}$. This value was chosen empirically as we found that higher time-steps tend to be noisier and less focused, while lower time-steps varied largely from pose to pose producing inferior attention grids. This can be seen qualitatively in Figure \ref{fig:supp_timestamps}. As illustrated in the figure, the attention values for the edit region get gradually more smeared and unfocused as the time-steps increase. This is evident, for instance, in warmer regions around the kangaroo's tail or the head of the duck. While perhaps less visually distinct, we can also observe that in lower timestamps the warm regions denoting high attention values cover a smaller area of the region which should be edited. We empirically find that this makes it more challenging for separating the object and edit regions.


\ignorethis{

\begin{figure*}[t] %
\centering 
 \jsubfig{\includegraphics[height=2cm]{images/comparisons/image_editing/dog_white.png}}{}
\jsubfig{\includegraphics[height=2cm]{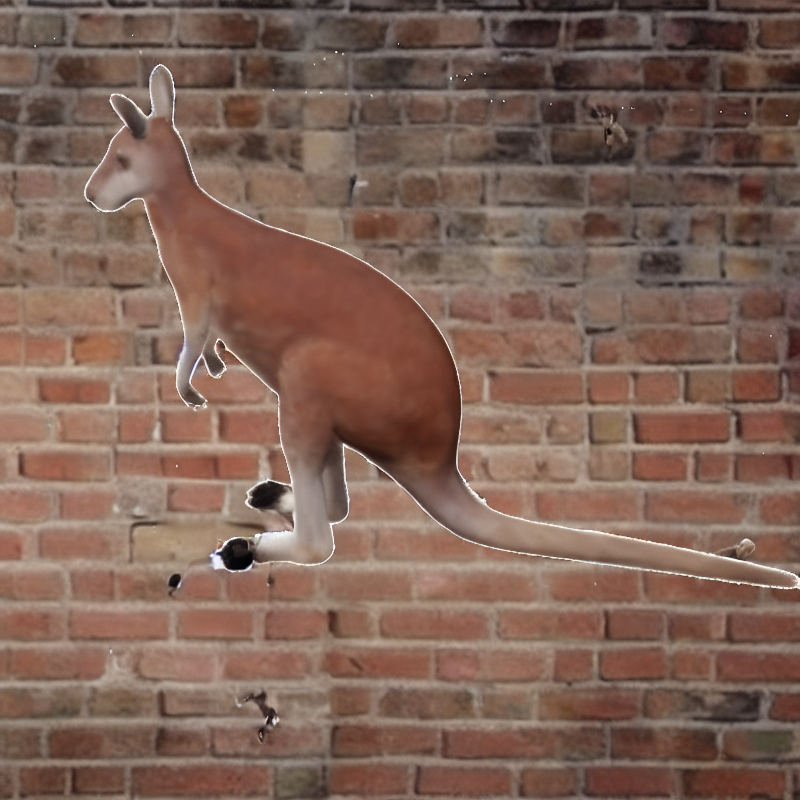}}{} 
\jsubfig{\includegraphics[height=2cm] {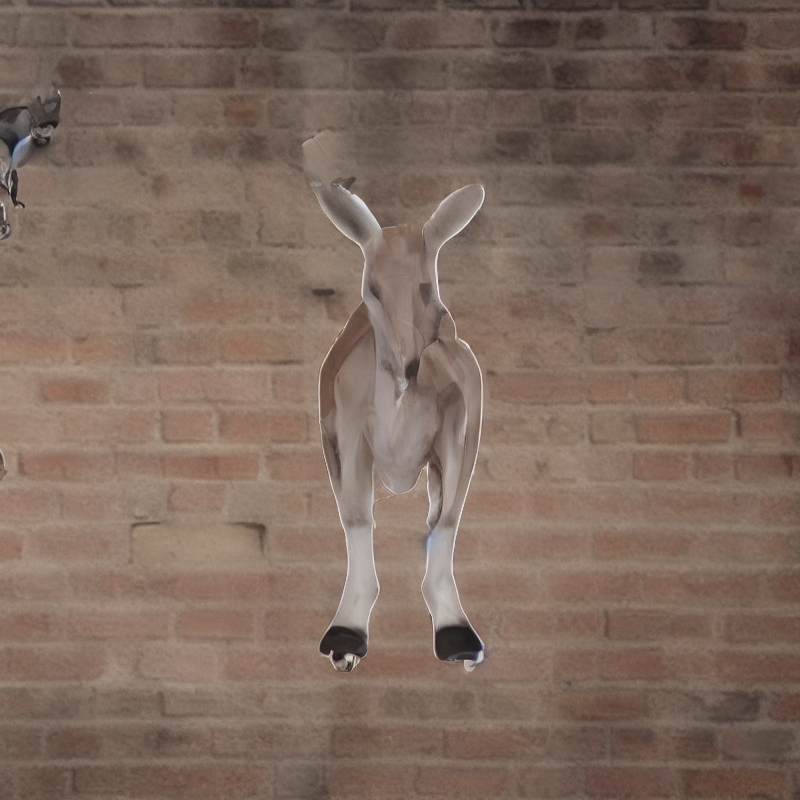}}{} 
\jsubfig{\includegraphics[height=2cm]{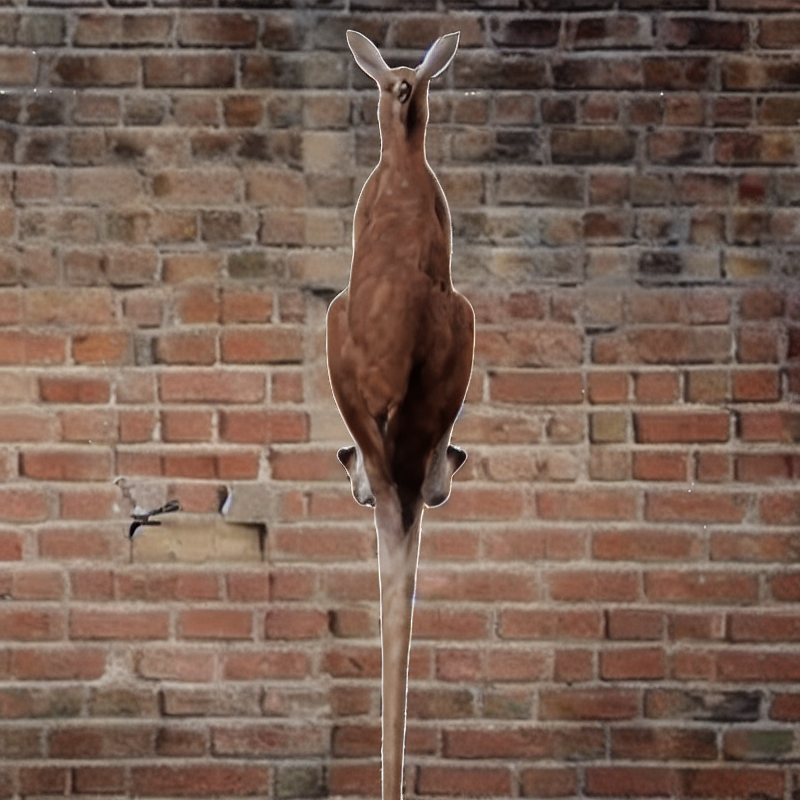}}{}
\hfill
\jsubfig{\includegraphics[height=2cm]{images/comparisons/image_editing/dog_white.png}}{}
\jsubfig{\includegraphics[height=2cm]{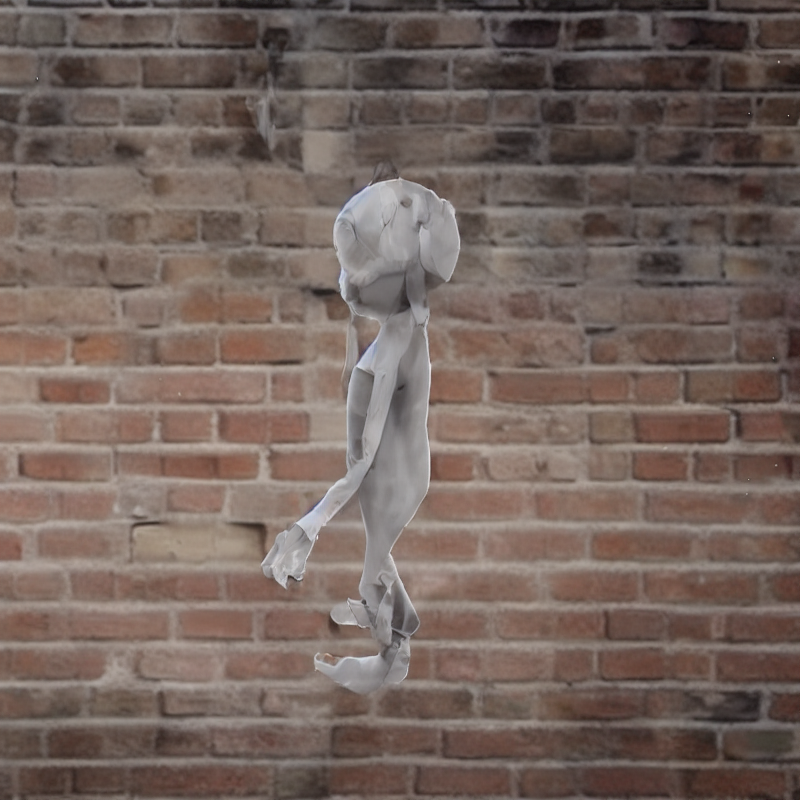}}{} 
\jsubfig{\includegraphics[height=2cm] {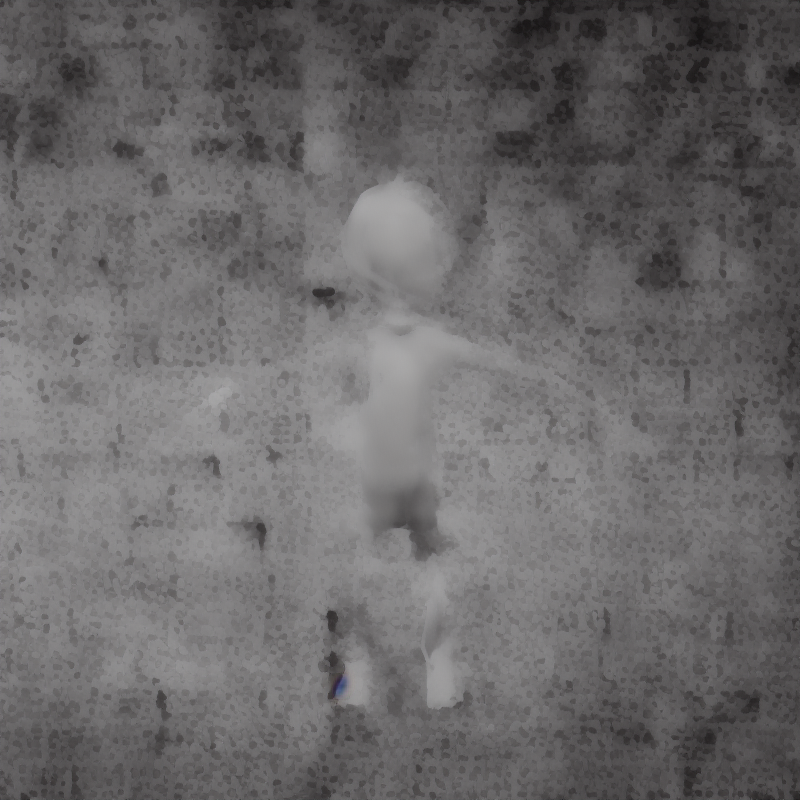}}{} 
\jsubfig{\includegraphics[height=2cm]{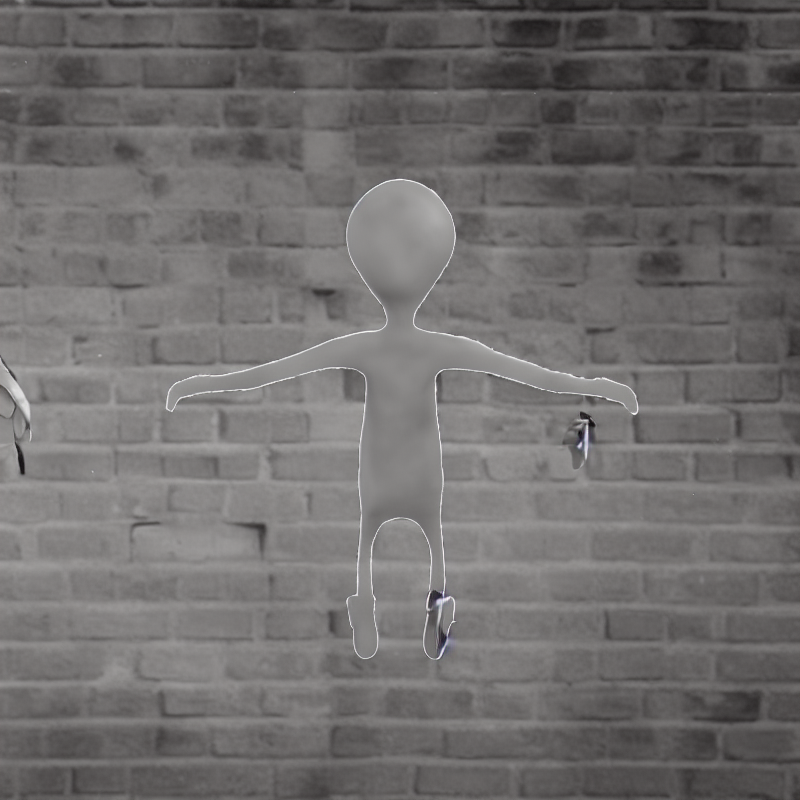}}{} 
\rotatebox[origin=tc]{-90}{SDEdit\whitetxt{x}\includegraphics[height=0.24cm]{images/bg.png}\whitetxt{xxxxxxxx}}
\\[-30pt] 
\jsubfig{\includegraphics[height=2cm]{images/comparisons/image_editing/dog_white.png}}{}
\jsubfig{\includegraphics[height=2cm]{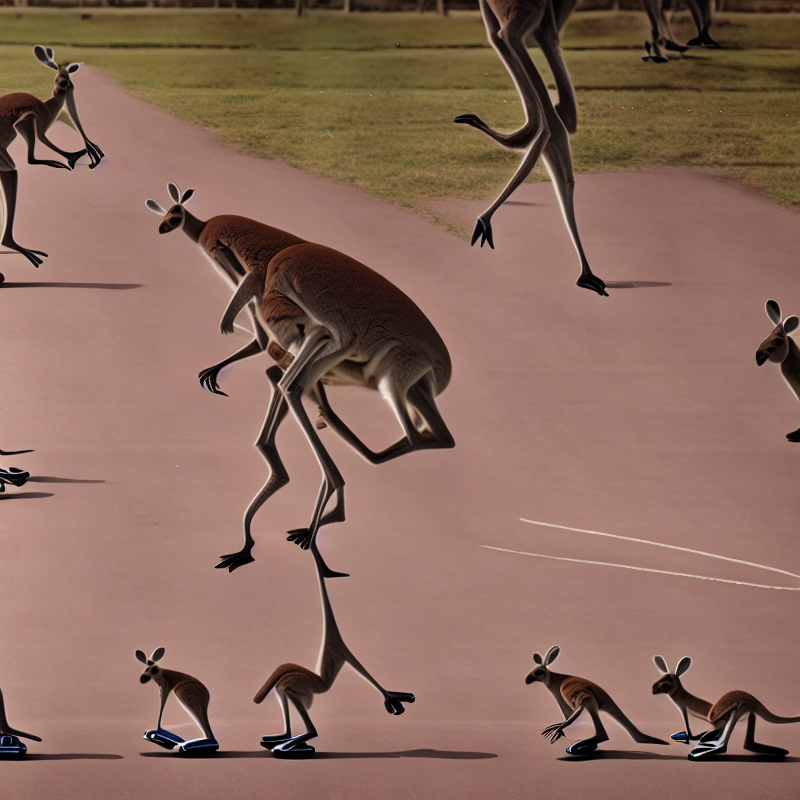}}{} 
\jsubfig{\includegraphics[height=2cm] {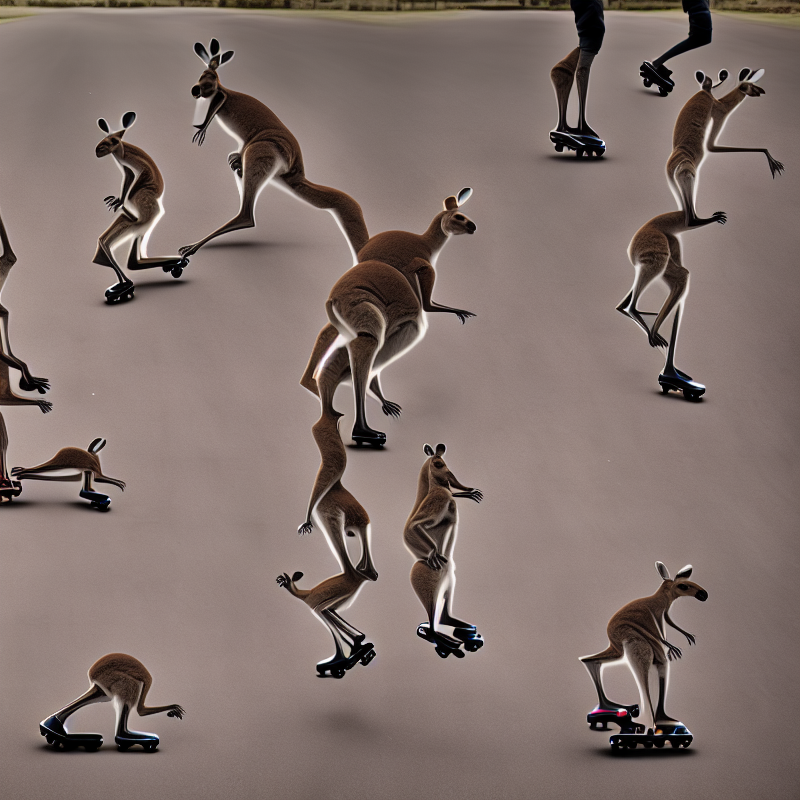}}{} 
\jsubfig{\includegraphics[height=2cm]{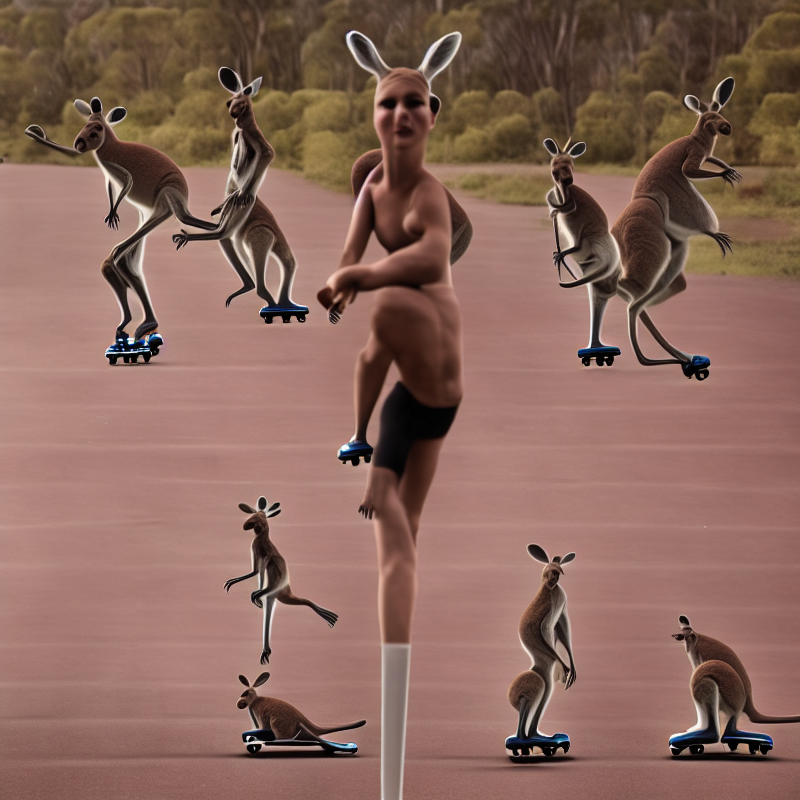}}{}
\hfill
\jsubfig{\includegraphics[height=2cm]{images/comparisons/image_editing/dog_white.png}}{}
\jsubfig{\includegraphics[height=2cm]{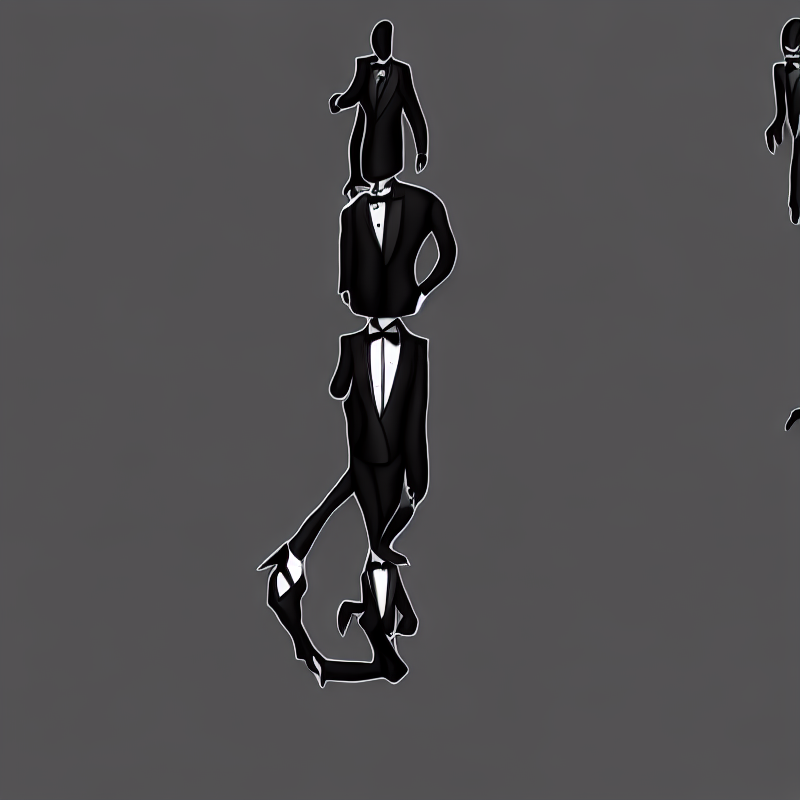}}{} 
\jsubfig{\includegraphics[height=2cm] {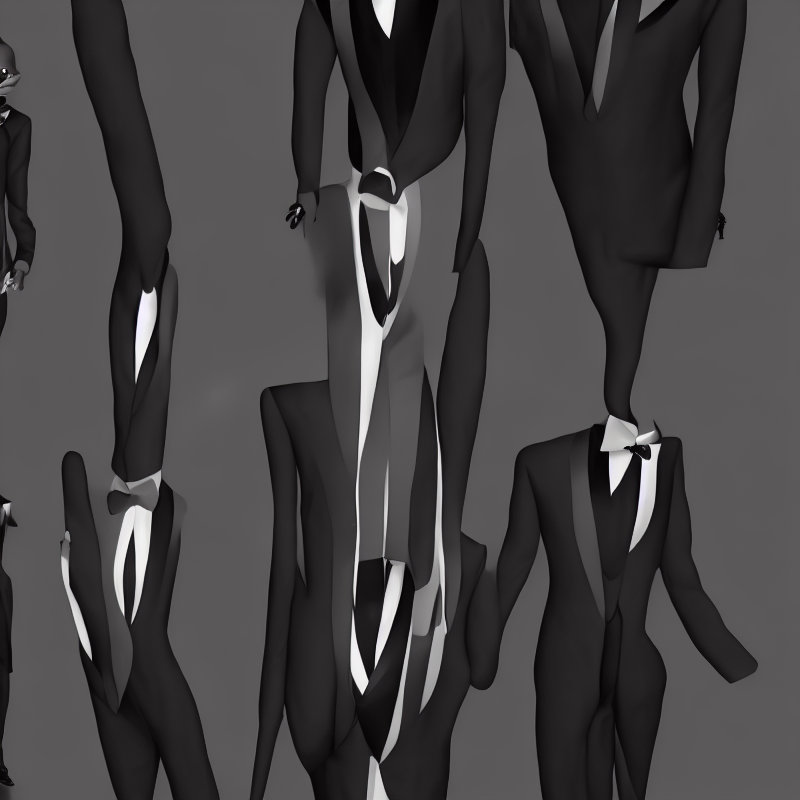}}{} 
\jsubfig{\includegraphics[height=2cm]{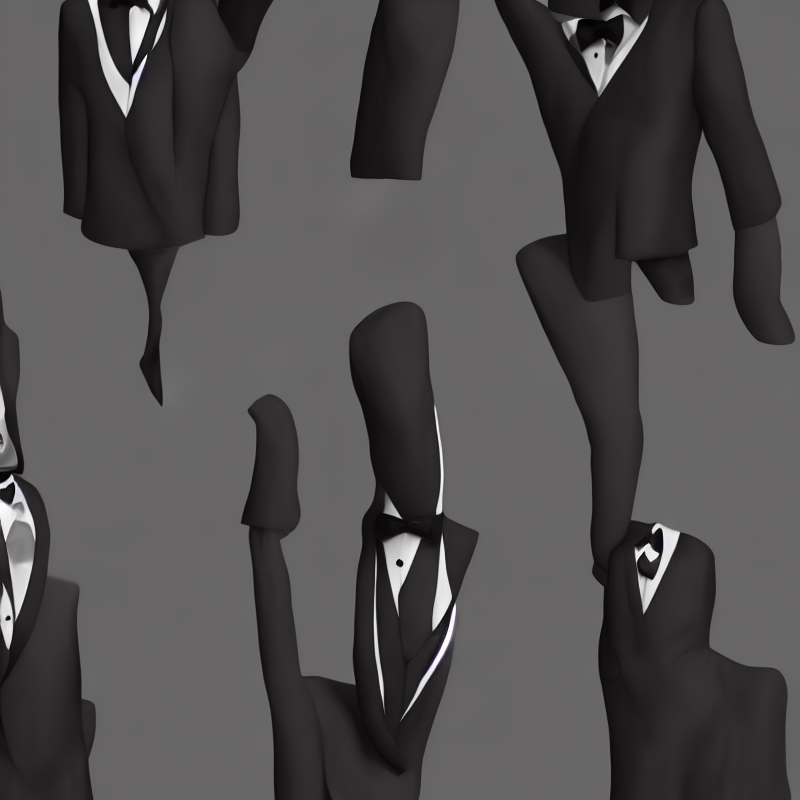}}{} 
\rotatebox[origin=tc]{-90}{SDEdit\whitetxt{x}\includegraphics[height=0.24cm]{images/bg.png}\whitetxt{xxxxxxxx}}
\\[-30pt] 
\jsubfig{\includegraphics[height=2cm]{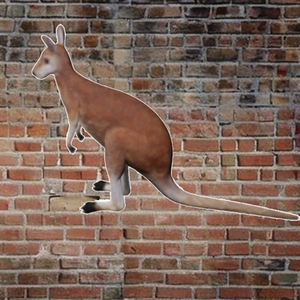}}{}
\jsubfig{\includegraphics[height=2cm]{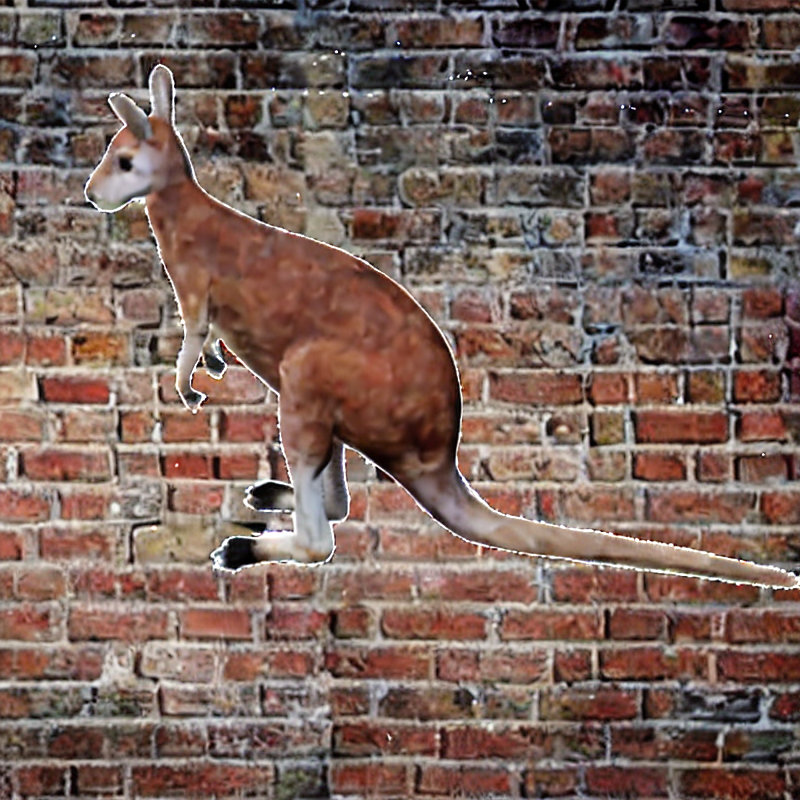}}{} 
\jsubfig{\includegraphics[height=2cm] {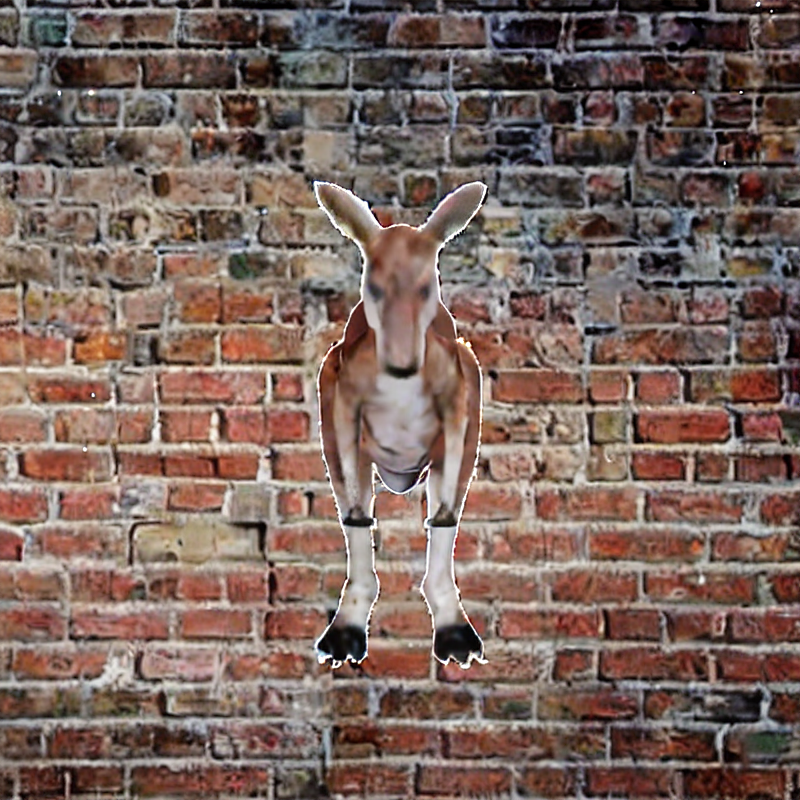}}{} 
\jsubfig{\includegraphics[height=2cm]{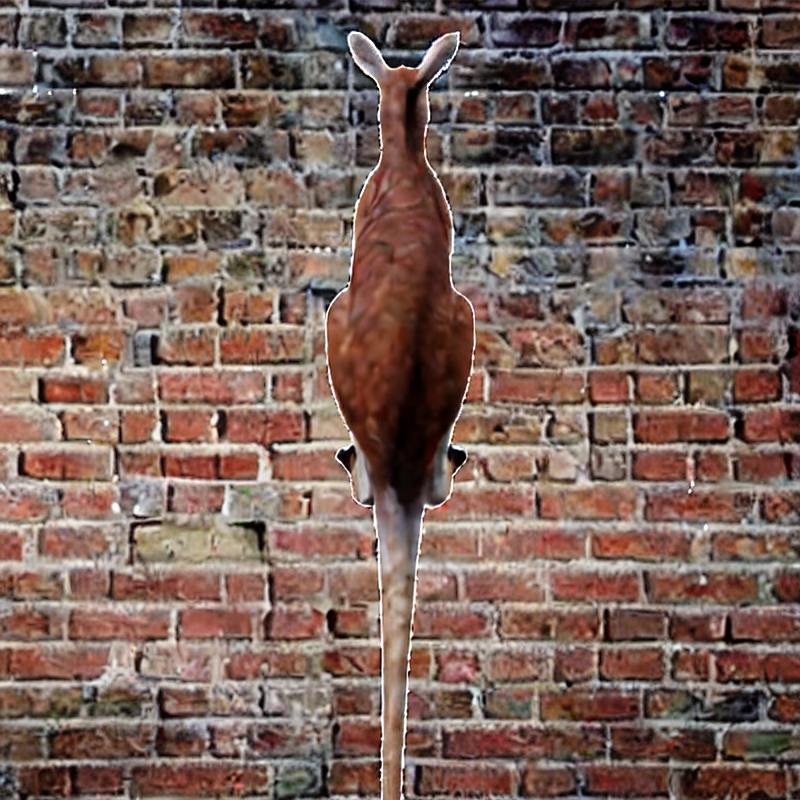}}{} 
\hfill
\jsubfig{\includegraphics[height=2cm]{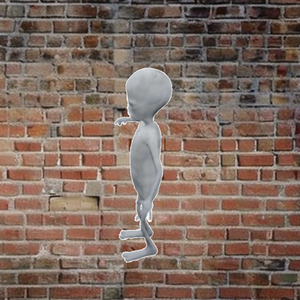}}{}
\jsubfig{\includegraphics[height=2cm]{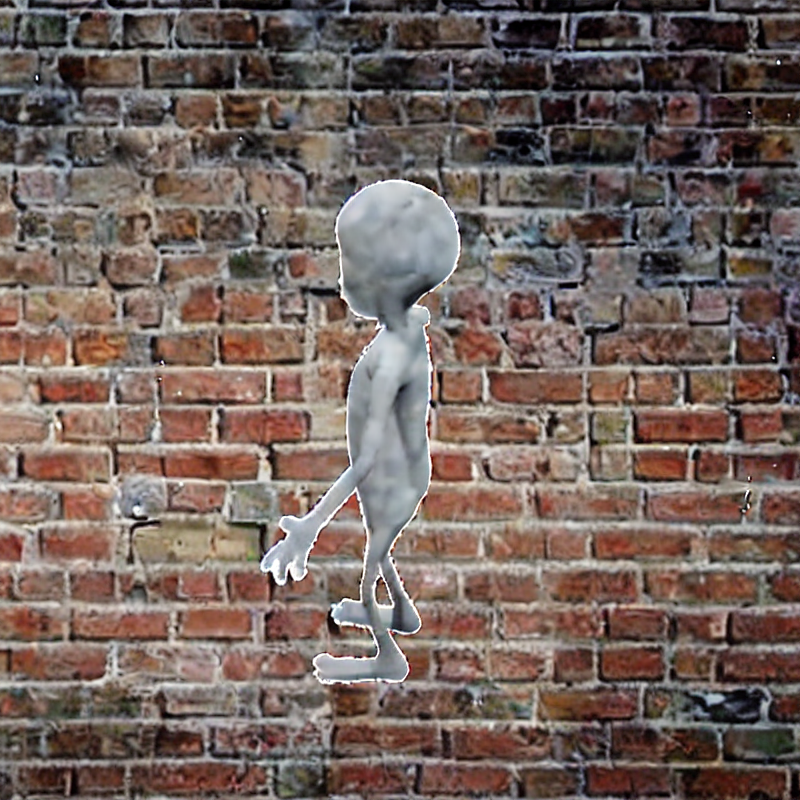}}{} 
\jsubfig{\includegraphics[height=2cm] {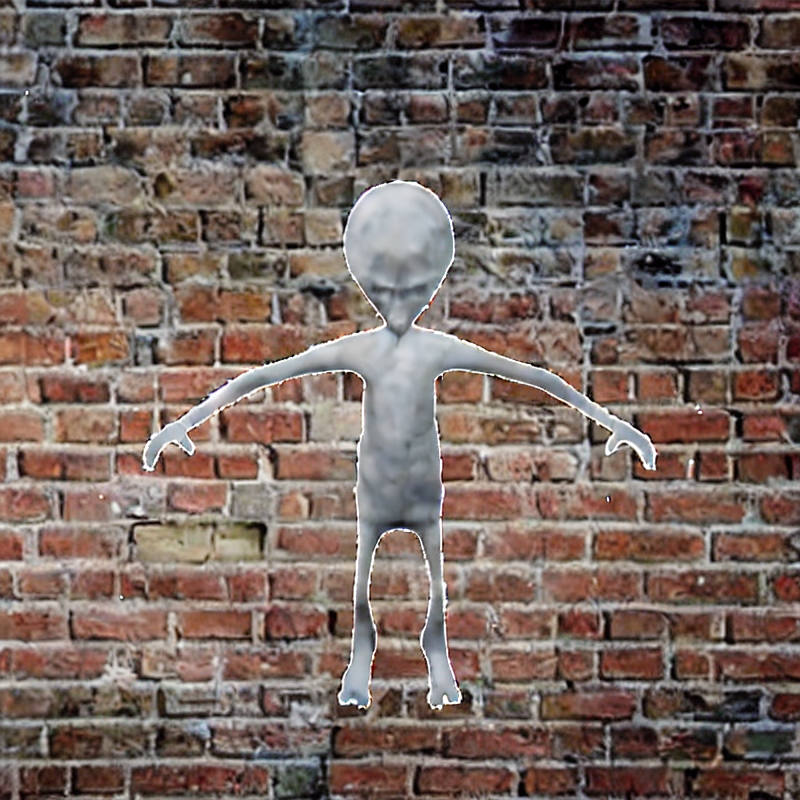}}{} 
\jsubfig{\includegraphics[height=2cm]{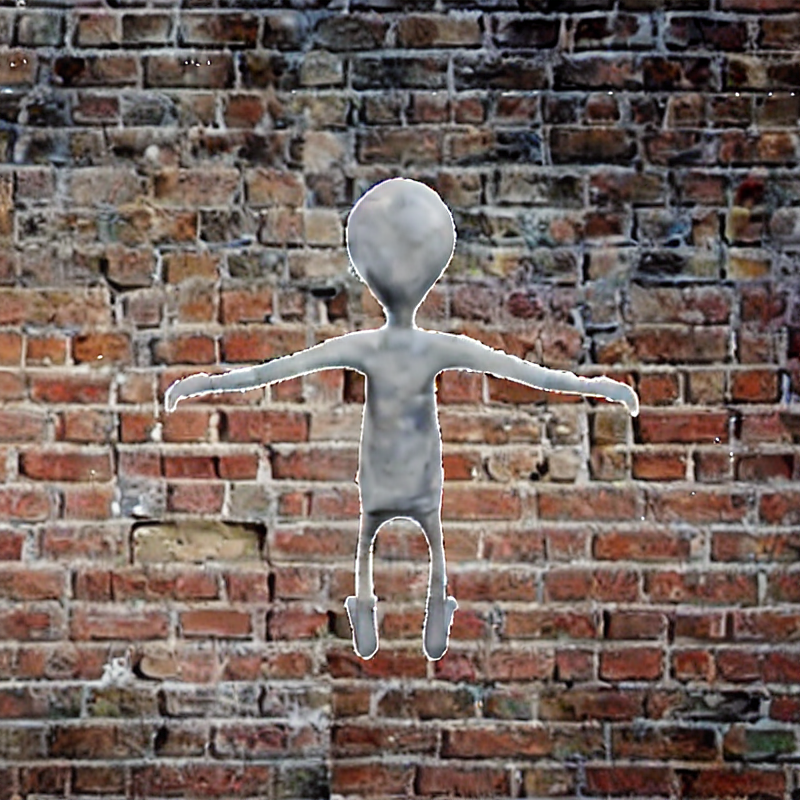}}{} 
\rotatebox[origin=tc]{-90}{IPix2Pix\whitetxt{x}\includegraphics[height=0.24cm]{images/bg.png}\whitetxt{xxxxxxxx}}
\\[-30pt]
\jsubfig{\includegraphics[height=2cm] {images/comparisons/image_editing/dog_white.png}}{}
\jsubfig{\includegraphics[height=2cm]{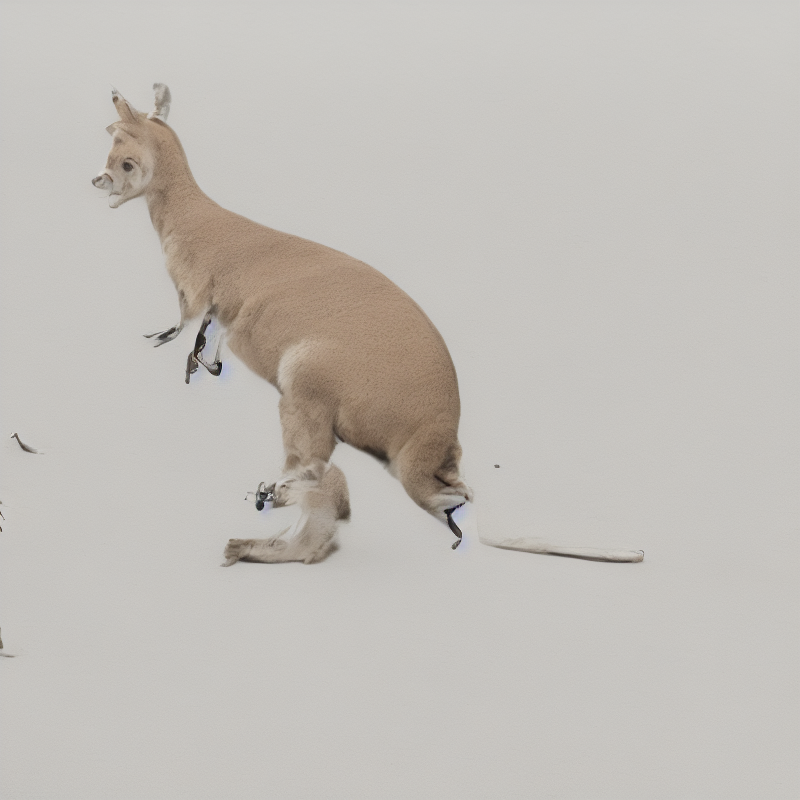}}{} 
\jsubfig{\includegraphics[height=2cm] {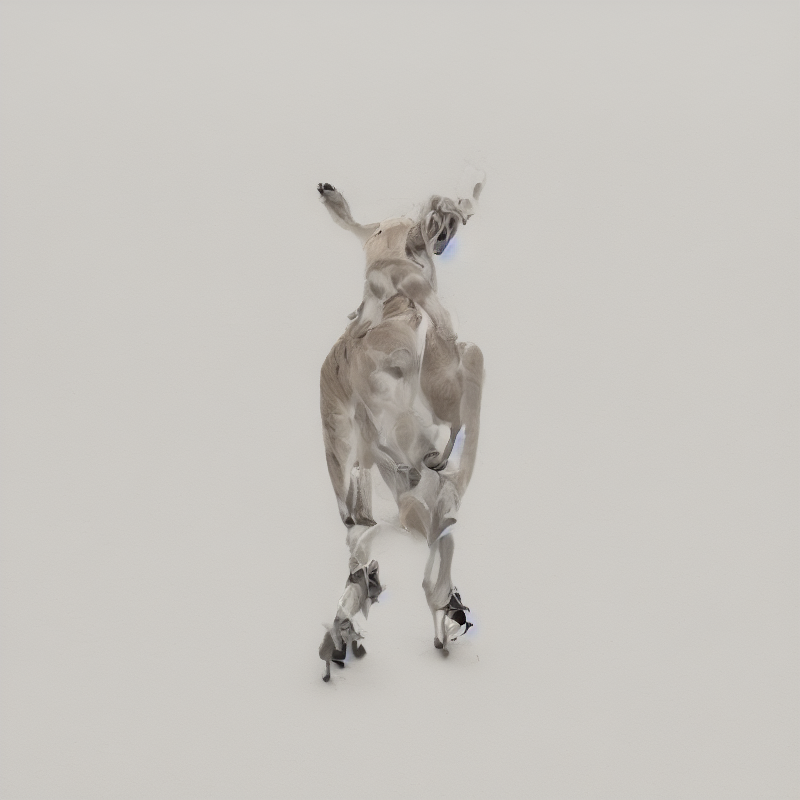}}{} 
\jsubfig{\includegraphics[height=2cm]{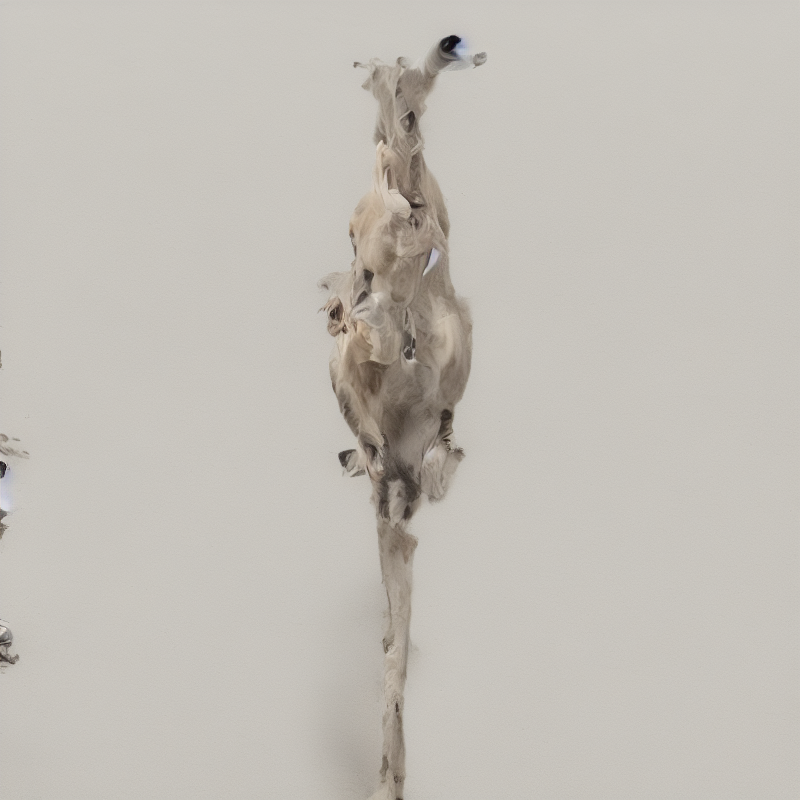}}{} 
\hfill
\jsubfig{\includegraphics[height=2cm] {images/comparisons/image_editing/dog_white.png}}{}
\jsubfig{\includegraphics[height=2cm]{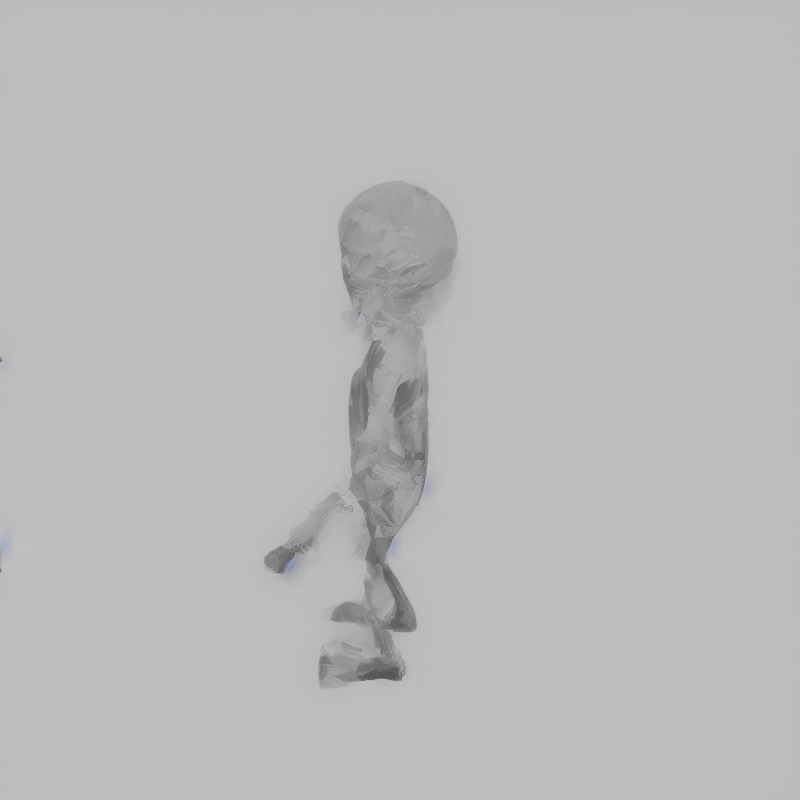}}{} 
\jsubfig{\includegraphics[height=2cm] {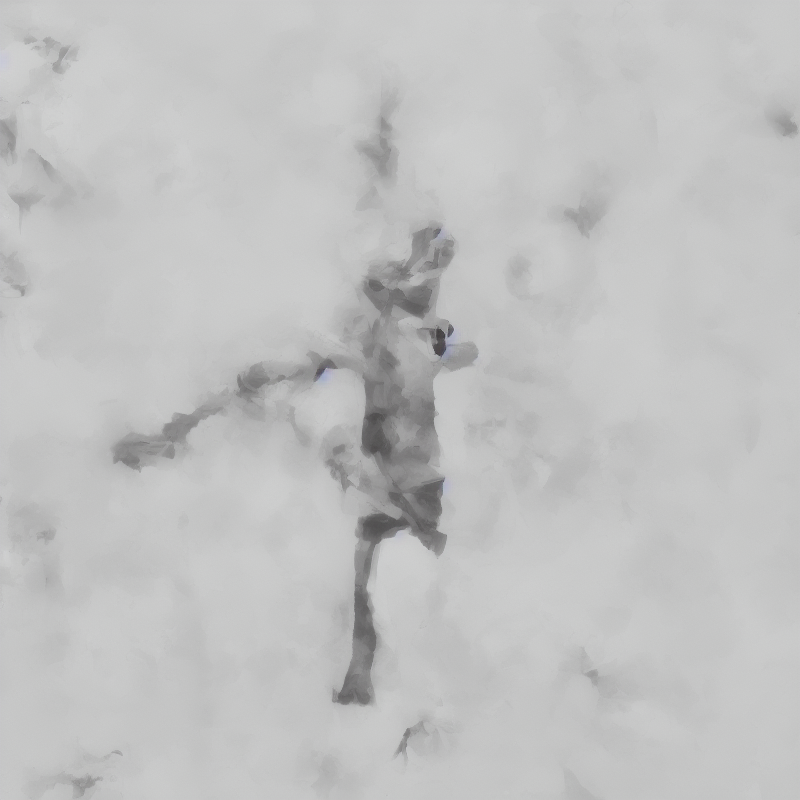}}{} 
\jsubfig{\includegraphics[height=2cm]{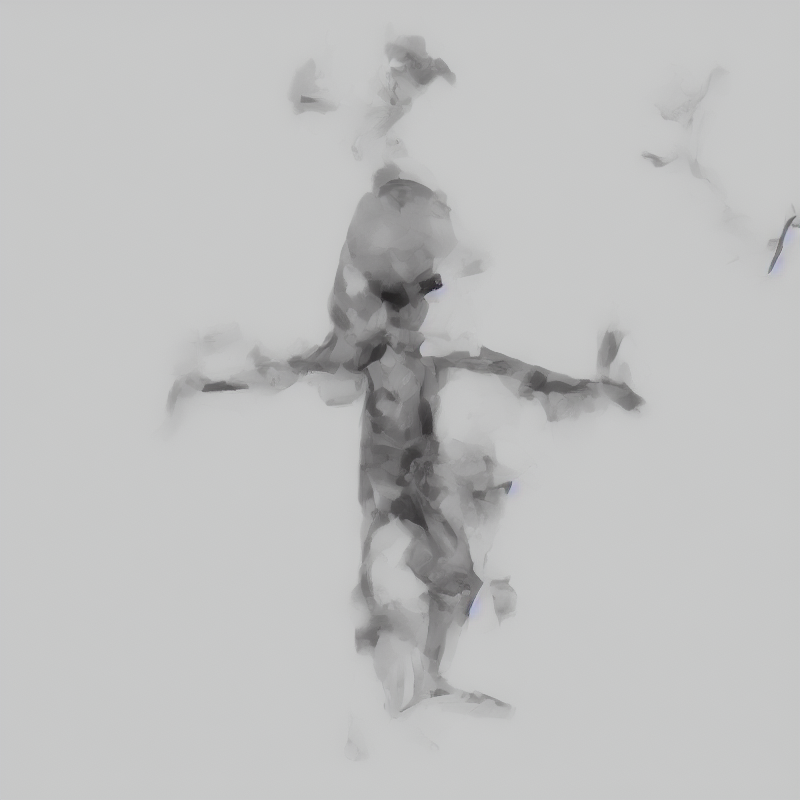}}{} 
\rotatebox[origin=tc]{-90}{SDEdit\whitetxt{xxxxxxxxxx}}
\\[-30pt] 
\jsubfig{\includegraphics[height=2cm] {images/comparisons/image_editing/dog_white.png}}{}
\jsubfig{\includegraphics[height=2cm]{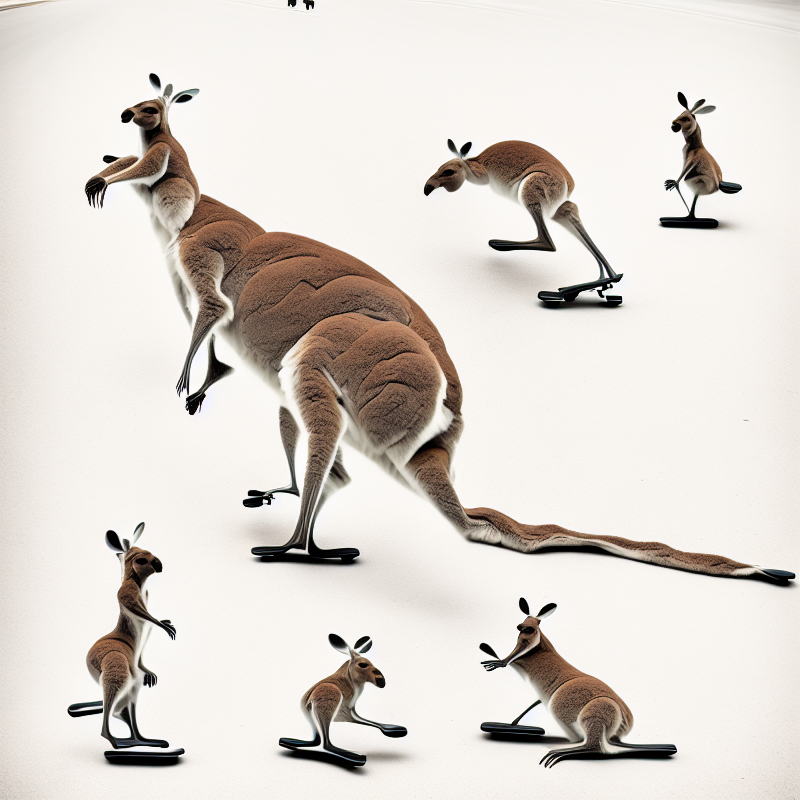}}{} 
\jsubfig{\includegraphics[height=2cm] {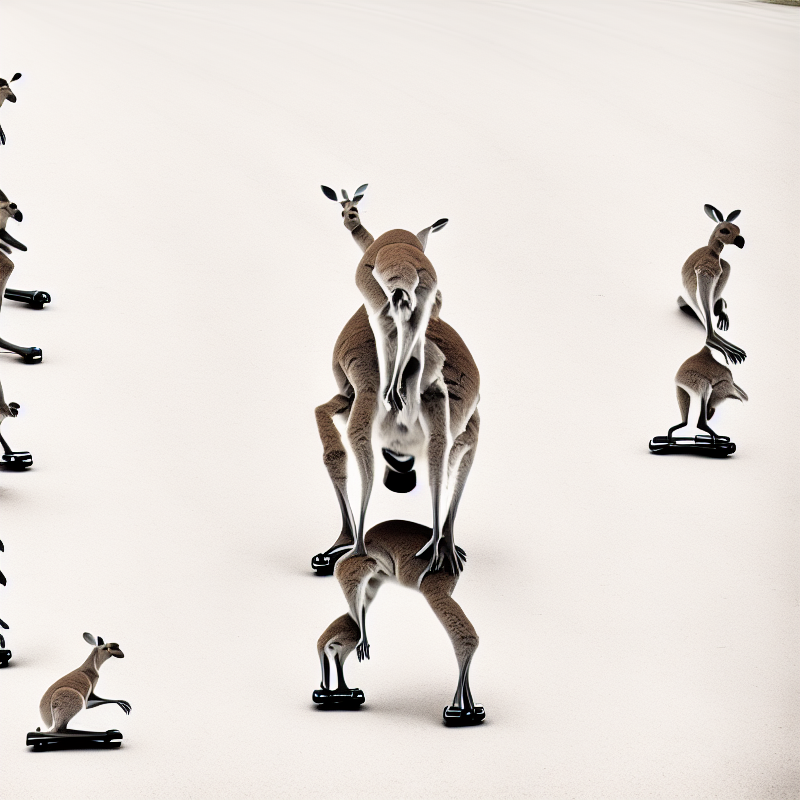}}{} 
\jsubfig{\includegraphics[height=2cm]{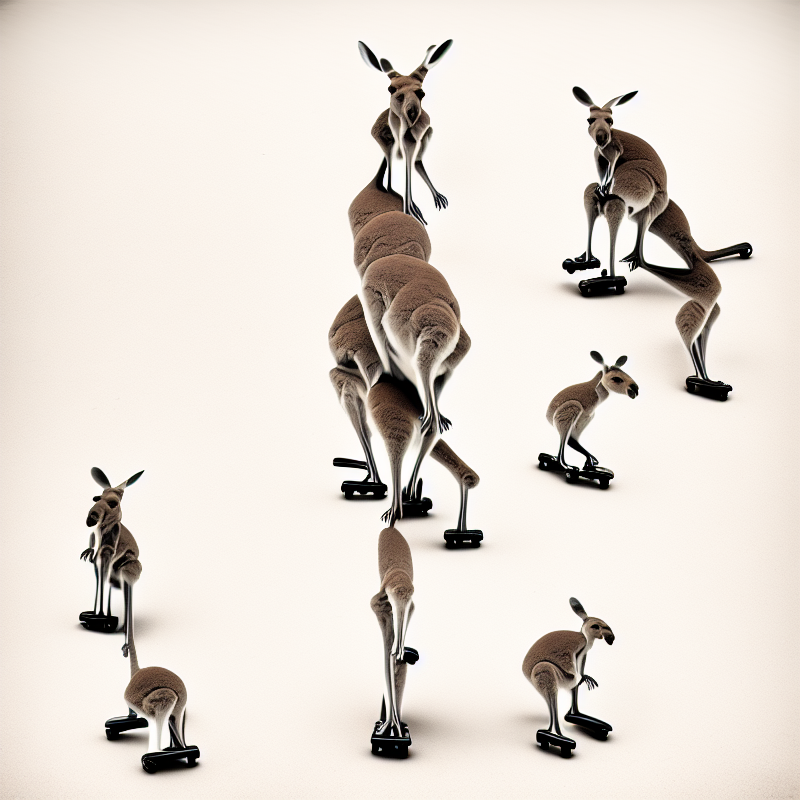}}{} 
\hfill
\jsubfig{\includegraphics[height=2cm] {images/comparisons/image_editing/dog_white.png}}{}
\jsubfig{\includegraphics[height=2cm]{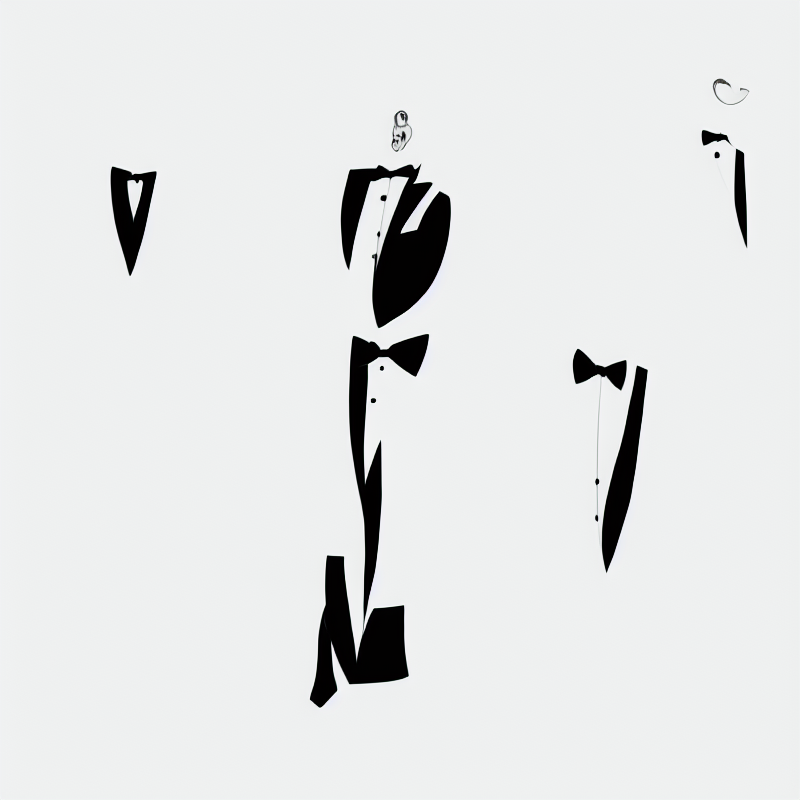}}{} 
\jsubfig{\includegraphics[height=2cm] {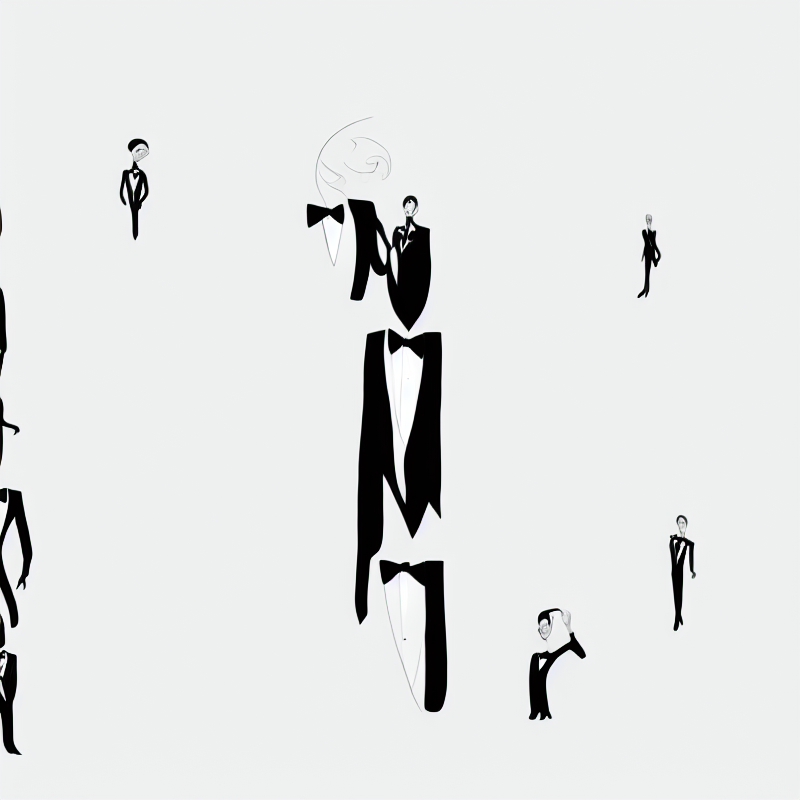}}{} 
\jsubfig{\includegraphics[height=2cm]{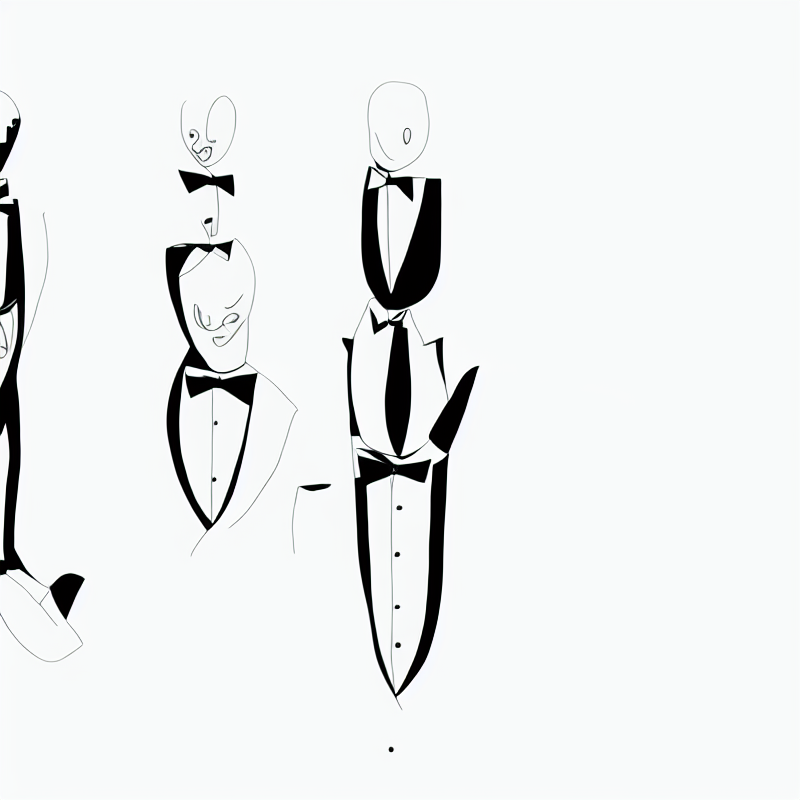}}{} 
\rotatebox[origin=tc]{-90}{IPix2Pix\whitetxt{xxxxxxxxxx}}
\\[-30pt] 
\jsubfig{\includegraphics[height=2cm] {images/comparisons/image_editing/dog_white.png}}{}
\jsubfig{\includegraphics[height=2cm]{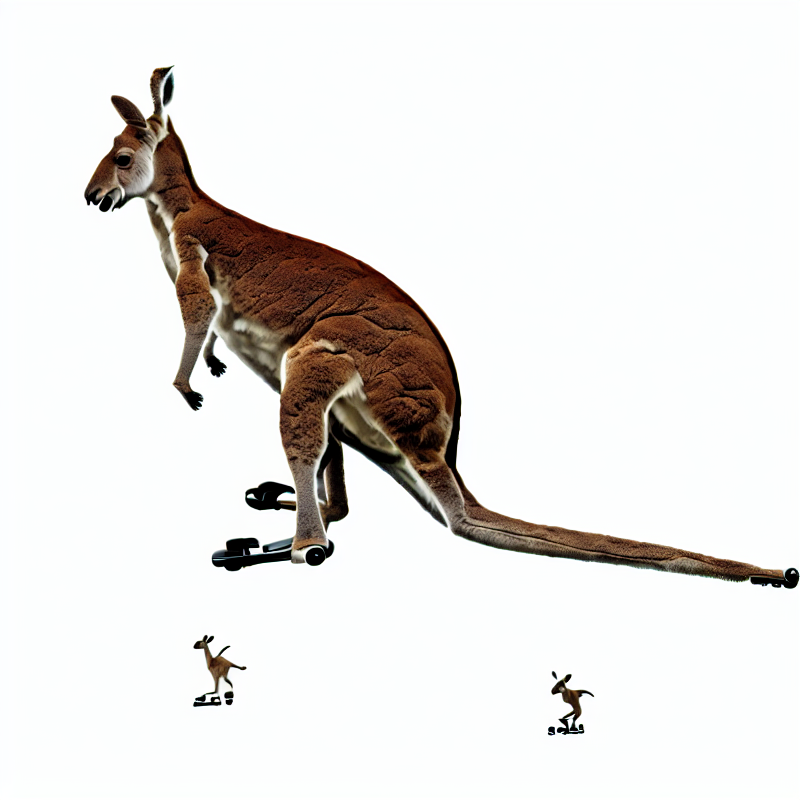}}{} 
\jsubfig{\includegraphics[height=2cm] {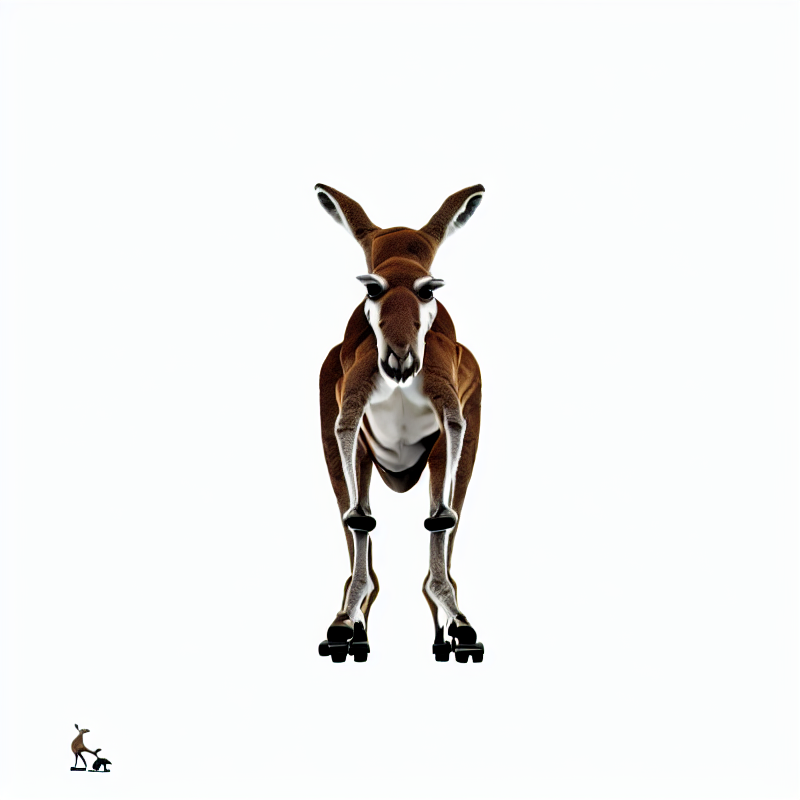}}{} 
\jsubfig{\includegraphics[height=2cm]{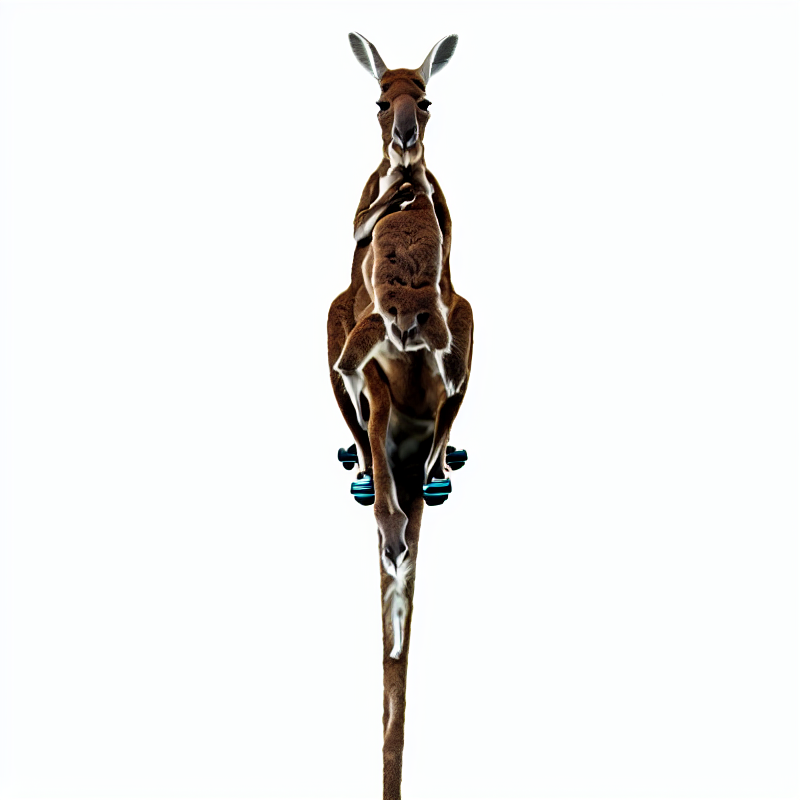}}{} 
\hfill
\jsubfig{\includegraphics[height=2cm] {images/comparisons/image_editing/dog_white.png}}{}
\jsubfig{\includegraphics[height=2cm]{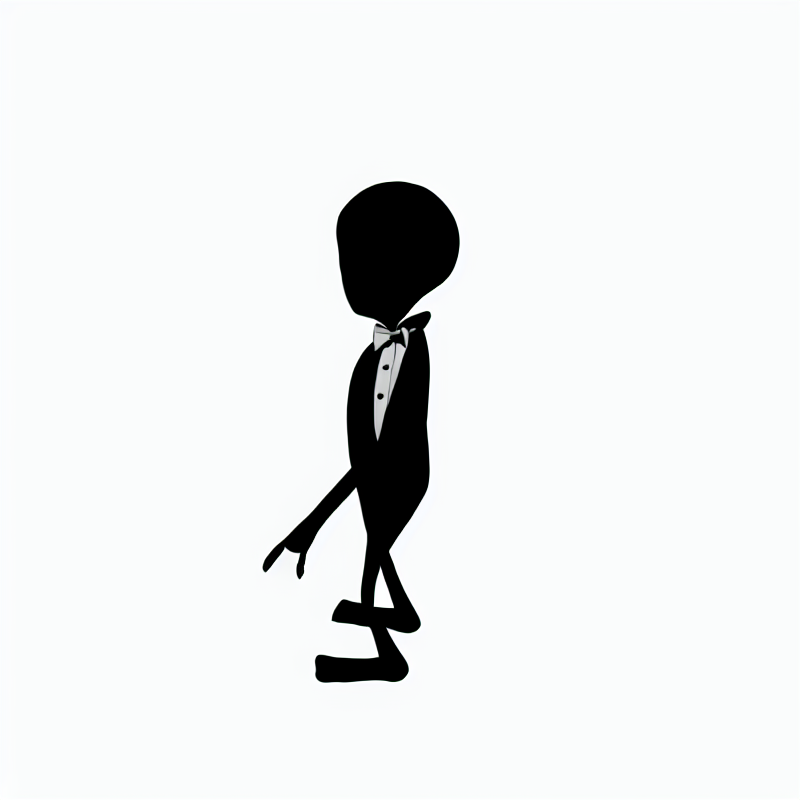}}{} 
\jsubfig{\includegraphics[height=2cm] {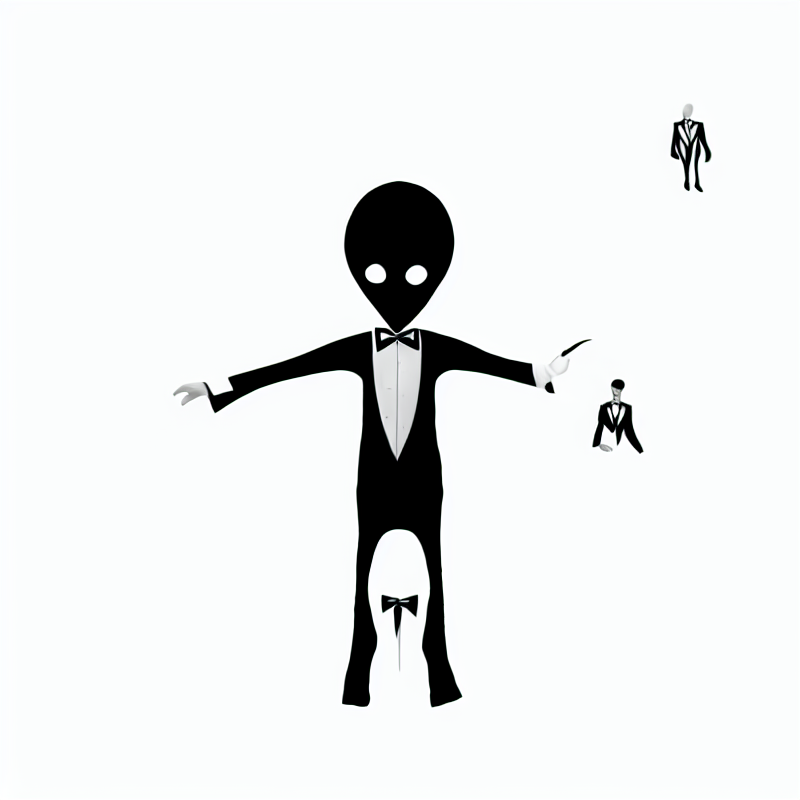}}{} 
\jsubfig{\includegraphics[height=2cm]{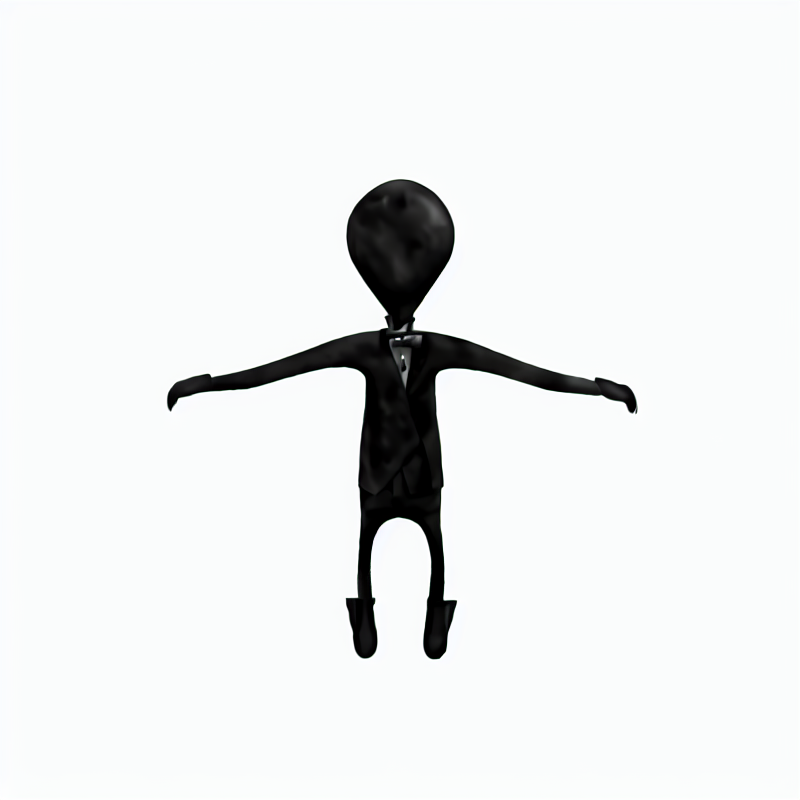}}{} 
\rotatebox[origin=tc]{-90}{IPix2Pix\whitetxt{xxxxxxxxxx}}
\\[-30pt] 
\jsubfig{\includegraphics[height=2cm]{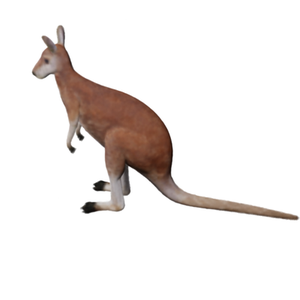}}{\footnotesize {Input}}
\jsubfig{\includegraphics[height=2cm]{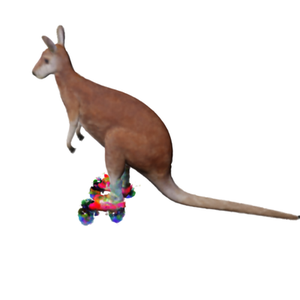} 
\includegraphics[height=2cm] {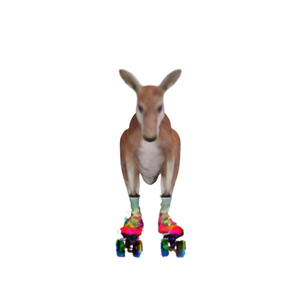} 
\includegraphics[height=2cm] {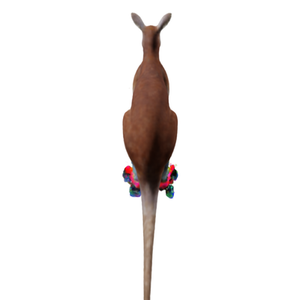}}{\footnotesize {''A kangaroo on rollerskates"}} 
\hfill
\jsubfig{\includegraphics[height=2cm]{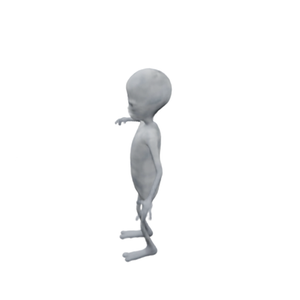}}{\footnotesize {''Input"}}
\jsubfig{\includegraphics[height=2cm]{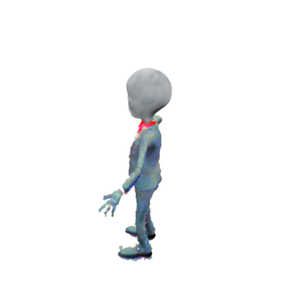}
\includegraphics[height=2cm] {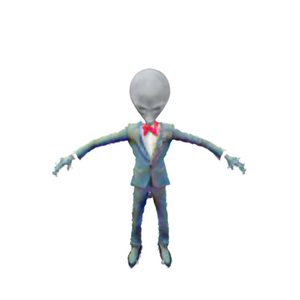} 
\includegraphics[height=2cm] {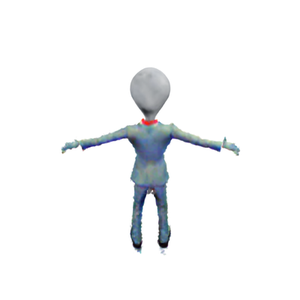}}{\footnotesize {''An alien wearing a tuxedo"}}
\rotatebox[origin=tc]{-90}{Ours\whitetxt{xxxxxxxxxxx}}
\\[-15pt] 

\ignorethis{
\jsubfig{\includegraphics[height=1.9cm]{images/comparisons/image_editing/dog_white.png}}{}
\rotatebox{90}{\whitetxt{x}SDEdit\whitetxt{x}\includegraphics[height=0.24cm]{images/bg.png}}
\hfill
\jsubfig{\includegraphics[height=1.9cm]{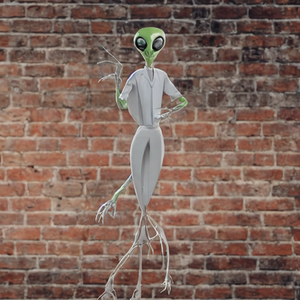}}{} 
\jsubfig{\includegraphics[height=1.9cm] {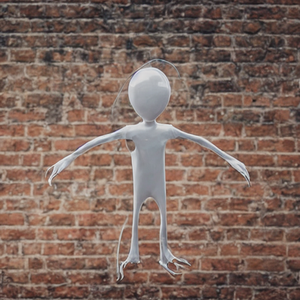}}{} 
\jsubfig{\includegraphics[height=1.9cm]{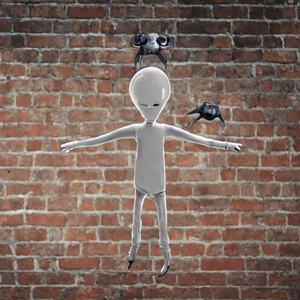}}{} 
\\ 
\jsubfig{\includegraphics[height=1.9cm]{images/comparisons/image_editing/alien/pix2pixin_40_bg.png}}{}
\rotatebox{90}{\whitetxt{x}IPix2Pix\whitetxt{x}\includegraphics[height=0.24cm]{images/bg.png}}
\hfill 
\hfill 
\jsubfig{\includegraphics[height=1.9cm]{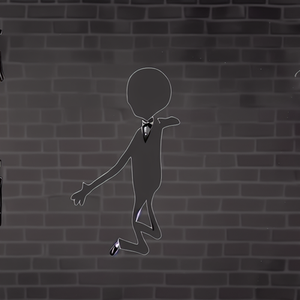}}{} 
\jsubfig{\includegraphics[height=1.9cm] {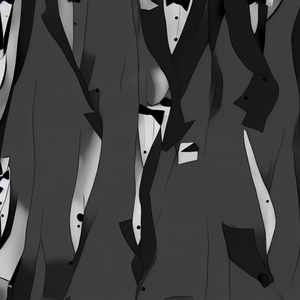}}{} 
\jsubfig{\includegraphics[height=1.9cm]{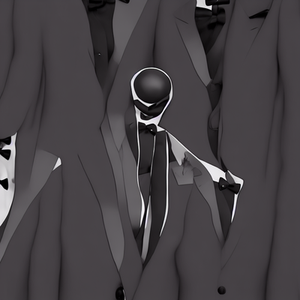}}{} 
\\
\jsubfig{\includegraphics[height=1.9cm] {images/comparisons/image_editing/dog_white.png}}{}
\rotatebox{90}{\whitetxt{xxx}SDEdit}
\hfill
\jsubfig{\includegraphics[height=1.9cm]{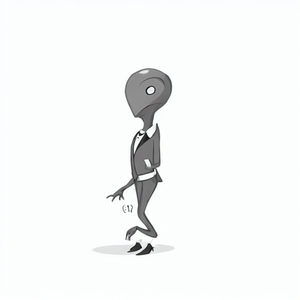}}{} 
\jsubfig{\includegraphics[height=1.9cm] {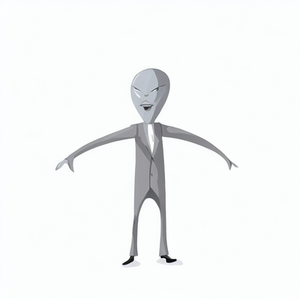}}{} 
\jsubfig{\includegraphics[height=1.9cm]{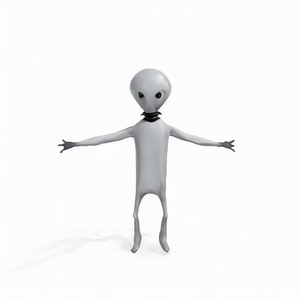}}{} 
\\ 
\jsubfig{\includegraphics[height=1.9cm] {images/comparisons/image_editing/dog_white.png}}{}
\rotatebox{90}{\whitetxt{xx}IPix2Pix}
\hfill
\jsubfig{\includegraphics[height=1.9cm]{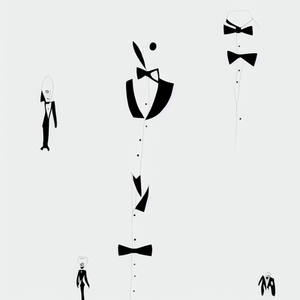}}{} 
\jsubfig{\includegraphics[height=1.9cm] {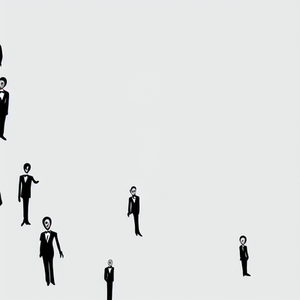}}{} 
\jsubfig{\includegraphics[height=1.9cm]{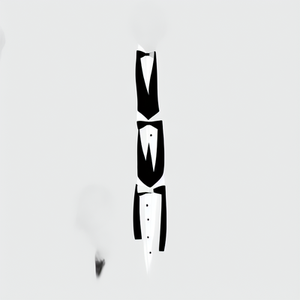}}{} 
\\
\jsubfig{\includegraphics[height=1.9cm]{images/comparisons/image_editing/alien/pix2pixin_40.png}}{}\rotatebox{90}{\whitetxt{xxx}Ours}
\hfill
\jsubfig{\includegraphics[height=1.9cm]{images/comparisons/image_editing/alien/alien_tux_ours_50.png}}{} 
\jsubfig{\includegraphics[height=1.9cm] {images/comparisons/image_editing/alien/alien_tux_ours_0.png}}{} 
\jsubfig{\includegraphics[height=1.9cm] {images/comparisons/image_editing/alien/alien_tux_ours_90.png}}{} 
}

\ignorethis{
\begin{figure*}[t] %
\centering 
 \jsubfig{\includegraphics[height=1.9cm]{images/comparisons/image_editing/dog_white.png}}{}
 \rotatebox{90}{\whitetxt{x}SDEdit\whitetxt{x}\includegraphics[height=0.24cm]{images/bg.png}}
 \hfill
\jsubfig{\includegraphics[height=1.9cm]{images/comparisons/image_editing/sde_dog_1.jpg}}{} 
\jsubfig{\includegraphics[height=1.9cm] {images/comparisons/image_editing/sde_dog_2.jpg}}{} 
\jsubfig{\includegraphics[height=1.9cm]{images/comparisons/image_editing/sde_dog_3.jpg}}{} 
\\ 
\jsubfig{\includegraphics[height=1.9cm]{images/comparisons/image_editing/dogpix2pixin_000004b.png}}{}
\rotatebox{90}{\whitetxt{x}IPix2Pix\whitetxt{x}\includegraphics[height=0.24cm]{images/bg.png}}
\hfill 
\hfill 
\jsubfig{\includegraphics[height=1.9cm]{images/comparisons/image_editing/dog_postpix2pix_000004b.png}}{} 
\jsubfig{\includegraphics[height=1.9cm] {images/comparisons/image_editing/dog_postpix2pix_000087b.png}}{} 
\jsubfig{\includegraphics[height=1.9cm]{images/comparisons/image_editing/dog_postpix2pix_000058b.png}}{} 
\\
\jsubfig{\includegraphics[height=1.9cm] {images/comparisons/image_editing/dog_white.png}}{}
\rotatebox{90}{\whitetxt{xxx}SDEdit}
\hfill
\jsubfig{\includegraphics[height=1.9cm]{images/comparisons/image_editing/sde_dog_1_nobg.jpg}}{} 
\jsubfig{\includegraphics[height=1.9cm] {images/comparisons/image_editing/sde_dog_2_nobg.jpg}}{} 
\jsubfig{\includegraphics[height=1.9cm]{images/comparisons/image_editing/sde_dog_3_nobg.jpg}}{} 
\\ 
\jsubfig{\includegraphics[height=1.9cm] {images/comparisons/image_editing/dog_white.png}}{}
\rotatebox{90}{\whitetxt{xx}IPix2Pix}
\hfill
\jsubfig{\includegraphics[height=1.9cm]{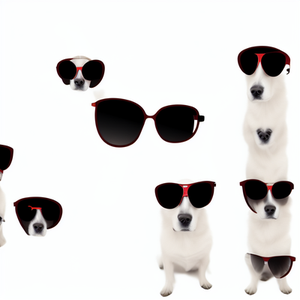}}{} 
\jsubfig{\includegraphics[height=1.9cm] {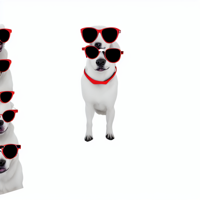}}{} 
\jsubfig{\includegraphics[height=1.9cm]{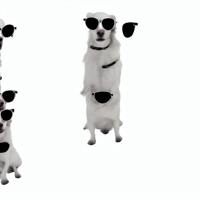}}{} 
\\
\jsubfig{\includegraphics[height=1.9cm]{images/comparisons/image_editing/dog_pix2pixin_000004.png}}{}\rotatebox{90}{\whitetxt{xxxxx}Ours}
\hfill
\jsubfig{\includegraphics[height=1.9cm]{images/comparisons/image_editing/color_50_dog.png}}{} 
\jsubfig{\includegraphics[height=1.9cm] {images/comparisons/image_editing/color_0_dog.png}}{} 
\jsubfig{\includegraphics[height=1.9cm] {images/comparisons/image_editing/color_90_dog.png}}{} 
\\
\jsubfig{\includegraphics[height=1.9cm]{images/comparisons/image_editing/dog_white.png}}{}
\rotatebox{90}{\whitetxt{x}SDEdit\whitetxt{x}\includegraphics[height=0.24cm]{images/bg.png}}
\hfill
\jsubfig{\includegraphics[height=1.9cm]{images/comparisons/image_editing/alien/postsde_pix2pixin_50.png}}{} 
\jsubfig{\includegraphics[height=1.9cm] {images/comparisons/image_editing/alien/postsde_pix2pixin_0.png}}{} 
\jsubfig{\includegraphics[height=1.9cm]{images/comparisons/image_editing/alien/postsde_pix2pixin_90.png}}{} 
\\ 
\jsubfig{\includegraphics[height=1.9cm]{images/comparisons/image_editing/alien/pix2pixin_40_bg.png}}{}
\rotatebox{90}{\whitetxt{x}IPix2Pix\whitetxt{x}\includegraphics[height=0.24cm]{images/bg.png}}
\hfill 
\hfill 
\jsubfig{\includegraphics[height=1.9cm]{images/comparisons/image_editing/alien/postpix2pix_60_bg.png}}{} 
\jsubfig{\includegraphics[height=1.9cm] {images/comparisons/image_editing/alien/postpix2pix_90_bg.png}}{} 
\jsubfig{\includegraphics[height=1.9cm]{images/comparisons/image_editing/alien/postpix2pix_150_bg.png}}{} 
\\
\jsubfig{\includegraphics[height=1.9cm] {images/comparisons/image_editing/dog_white.png}}{}
\rotatebox{90}{\whitetxt{xxx}SDEdit}
\hfill
\jsubfig{\includegraphics[height=1.9cm]{images/comparisons/image_editing/alien/postsde_50.png}}{} 
\jsubfig{\includegraphics[height=1.9cm] {images/comparisons/image_editing/alien/postsde_0.png}}{} 
\jsubfig{\includegraphics[height=1.9cm]{images/comparisons/image_editing/alien/postsde_90.png}}{} 
\\ 
\jsubfig{\includegraphics[height=1.9cm] {images/comparisons/image_editing/dog_white.png}}{}
\rotatebox{90}{\whitetxt{xx}IPix2Pix}
\hfill
\jsubfig{\includegraphics[height=1.9cm]{images/comparisons/image_editing/alien/postpix2pix_50.png}}{} 
\jsubfig{\includegraphics[height=1.9cm] {images/comparisons/image_editing/alien/postpix2pix_0.png}}{} 
\jsubfig{\includegraphics[height=1.9cm]{images/comparisons/image_editing/alien/postpix2pix_90.png}}{} 
\\
\jsubfig{\includegraphics[height=1.9cm]{images/comparisons/image_editing/alien/pix2pixin_40.png}}{}\rotatebox{90}{\whitetxt{xxx}Ours}
\hfill
\jsubfig{\includegraphics[height=1.9cm]{images/comparisons/image_editing/alien/alien_tux_ours_50.png}}{} 
\jsubfig{\includegraphics[height=1.9cm] {images/comparisons/image_editing/alien/alien_tux_ours_0.png}}{} 
\jsubfig{\includegraphics[height=1.9cm] {images/comparisons/image_editing/alien/alien_tux_ours_90.png}}{} 
}

\ignorethis{
  \jsubfig{\includegraphics[height=1.9cm]{images/comparisons/image_editing/dog_white.png}}{}
 \rotatebox{90}{\whitetxt{x}SDEdit\whitetxt{x}\includegraphics[height=0.24cm]{images/bg.png}}
 \hfill
\jsubfig{\includegraphics[height=1.9cm]{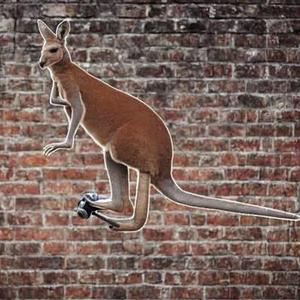}}{} 
\jsubfig{\includegraphics[height=1.9cm] {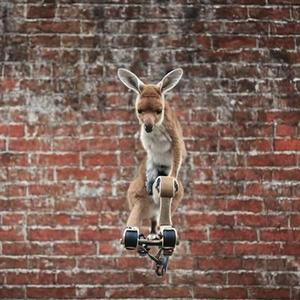}}{} 
\jsubfig{\includegraphics[height=1.9cm]{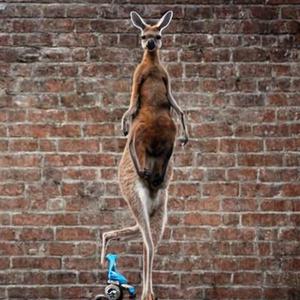}}{} 
\\ 
\jsubfig{\includegraphics[height=1.9cm]{images/comparisons/image_editing/kangaroo/pix2pixin_50b.png}}{}
\rotatebox{90}{\whitetxt{x}IPix2Pix\whitetxt{x}\includegraphics[height=0.24cm]{images/bg.png}}
\hfill 
\hfill 
\jsubfig{\includegraphics[height=1.9cm]{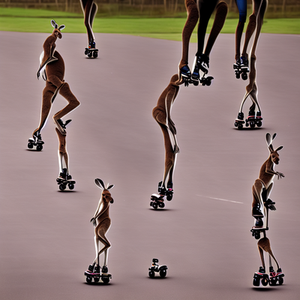}}{} 
\jsubfig{\includegraphics[height=1.9cm] {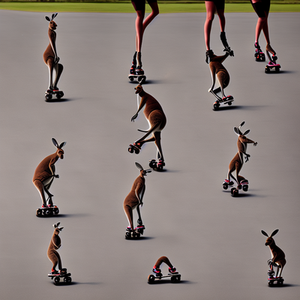}}{} 
\jsubfig{\includegraphics[height=1.9cm]{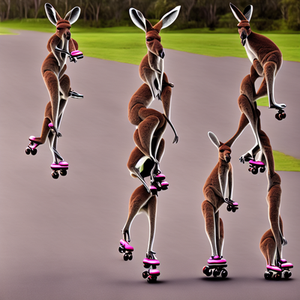}}{} 
\\
\jsubfig{\includegraphics[height=1.9cm] {images/comparisons/image_editing/dog_white.png}}{}
\rotatebox{90}{\whitetxt{xxx}SDEdit}
\hfill
\jsubfig{\includegraphics[height=1.9cm]{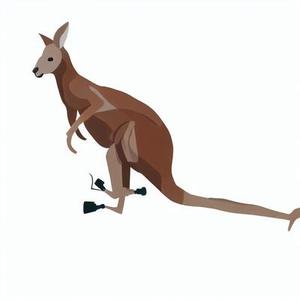}}{} 
\jsubfig{\includegraphics[height=1.9cm] {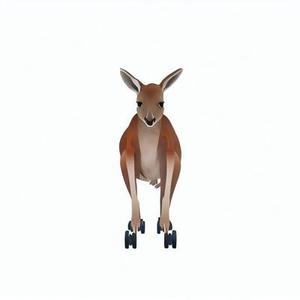}}{} 
\jsubfig{\includegraphics[height=1.9cm]{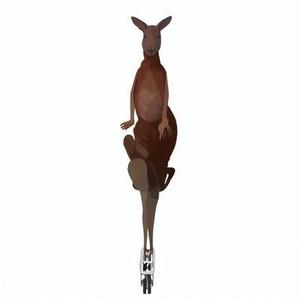}}{} 
\\ 
\jsubfig{\includegraphics[height=1.9cm] {images/comparisons/image_editing/dog_white.png}}{}
\rotatebox{90}{\whitetxt{xx}IPix2Pix}
\hfill
\jsubfig{\includegraphics[height=1.9cm]{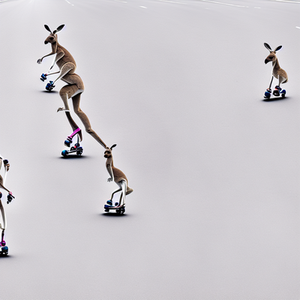}}{} 
\jsubfig{\includegraphics[height=1.9cm] {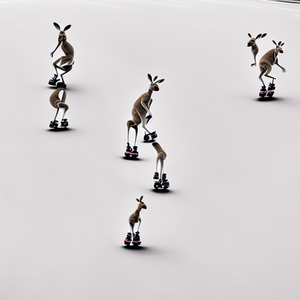}}{} 
\jsubfig{\includegraphics[height=1.9cm]{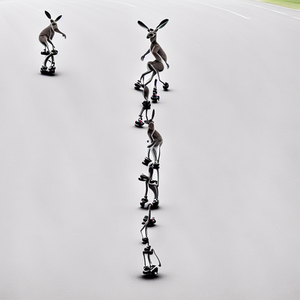}}{} 
\\
\jsubfig{\includegraphics[height=1.9cm]{images/comparisons/image_editing/kangaroo/pix2pixin_50.png}}{}\rotatebox{90}{\whitetxt{xxx}Ours}
\hfill
\jsubfig{\includegraphics[height=1.9cm]{images/comparisons/image_editing/kangaroo/color_50.png}}{} 
\jsubfig{\includegraphics[height=1.9cm] {images/comparisons/image_editing/kangaroo/color_0.png}}{} 
\jsubfig{\includegraphics[height=1.9cm] {images/comparisons/image_editing/kangaroo/color_90.png}}{}
}

\caption{\esc{Need to change this?} We compare to the text-guided image editing techniques InstructPix2Pix (IPix2Pix)~\cite{brooks2022instructpix2pix} and SDEdit~\cite{meng2022sdedit} by providing it with images from different viewpoints and text prompts. For IPix2Pix, prompts are modified (e.g., we use ``put sunglasses on the dog" instead of ``a dog with sunglasses"). 
We show one input image on the left, and three outputs on the right (side, front and back views), where the leftmost output corresponds to the input viewpoint. We show two variants, one with added backgrounds (top rows), as we observe that the white backgrounds are generally more challenging for these methods. As illustrated, 2D techniques cannot easily achieve 3D-consistent edit results. 
}
\label{fig:comparison2d_supp}
\end{figure*}

}

\section{Additional Visualizations and Results}
\label{sec:res}

\paragraph{Visualizing 2D cross-attention maps and images rendered from our 3D cross-attention grids}
While the attention maps used as ground-truth are inherently unfocused (as they are up-sampled from very low resolutions) and are not guaranteed to be view consistent, we show that learning the projection of these attention maps on to our object’s density produces view-consistent heat maps for object and edit regions (Figure \ref{fig:supp_attn_kangaroo}).

\ignorethis{
\paragraph{Additional 2D image editing results} In Figure \ref{fig:comparison2d_supp} we show additional comparisons to 2D image editing techniques. These results further illustrate that 2D methods struggle to produce view consistent results. 
}

\begin{figure*} %
\centering
\rotatebox{90}{2D cross-attention} \hfill
\jsubfig{\includegraphics[height=2.72cm]{images/supp_attn/kangaroo/2d_edit/0_edit.png}
\includegraphics[height=2.72cm]{images/supp_attn/kangaroo/2d_edit/30_edit.png}
\includegraphics[height=2.72cm]{images/supp_attn/kangaroo/2d_edit/60_edit.png}
\includegraphics[height=2.72cm]{images/supp_attn/kangaroo/2d_edit/90_edit.png}
\includegraphics[height=2.72cm]{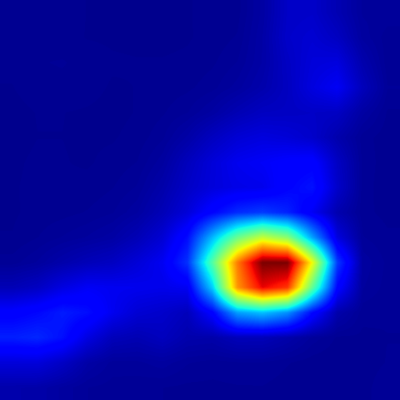}
\includegraphics[height=2.72cm]{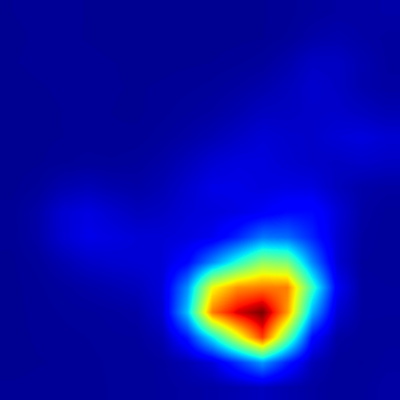}}{}
\\
\rotatebox{90}{\whitetxt{xx}3D grid ($A_e$)} \hfill
\jsubfig{\includegraphics[height=2.72cm]{images/supp_attn/kangaroo/3d_edit/0.png}
\includegraphics[height=2.72cm]{images/supp_attn/kangaroo/3d_edit/30.png}
\includegraphics[height=2.72cm]{images/supp_attn/kangaroo/3d_edit/60.png}
\includegraphics[height=2.72cm]{images/supp_attn/kangaroo/3d_edit/90.png}
\includegraphics[height=2.72cm]{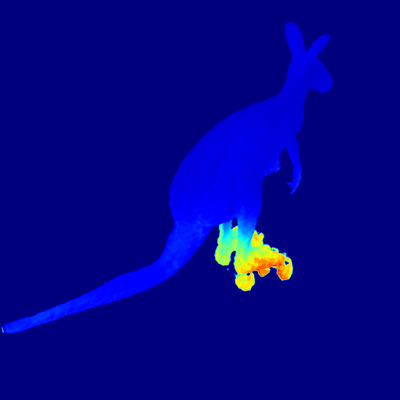}
\includegraphics[height=2.72cm]{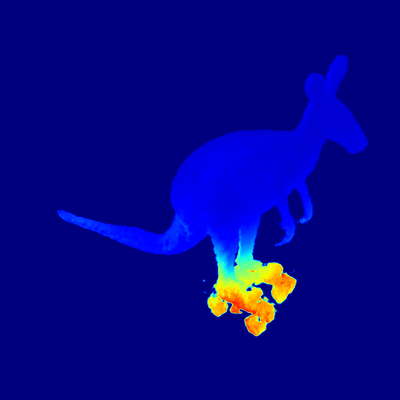}}{} 
\\ 
\rotatebox{90}{2D cross-attention} \hfill 
\jsubfig{\includegraphics[height=2.72cm]{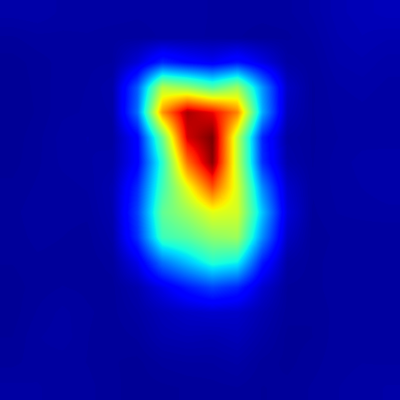}
\includegraphics[height=2.72cm]{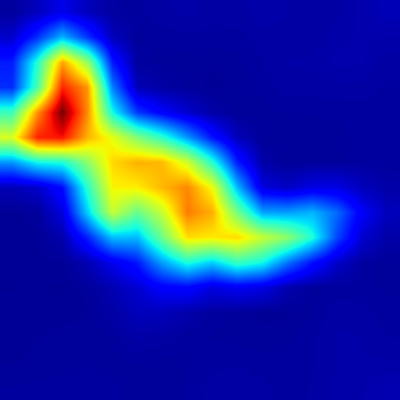}
\includegraphics[height=2.72cm]{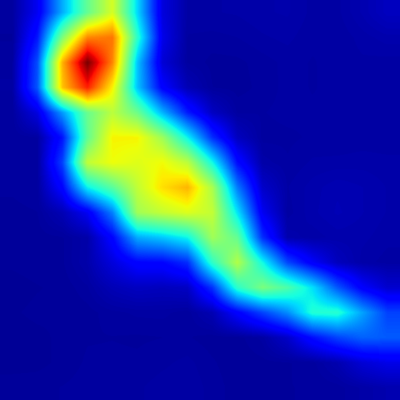}
\includegraphics[height=2.72cm]{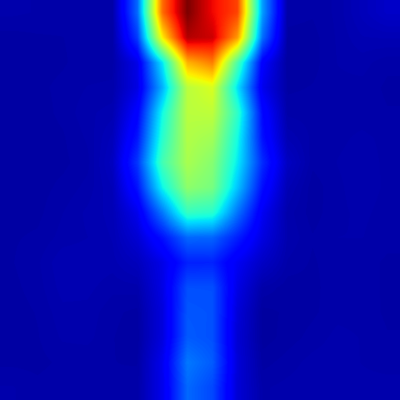}
\includegraphics[height=2.72cm]{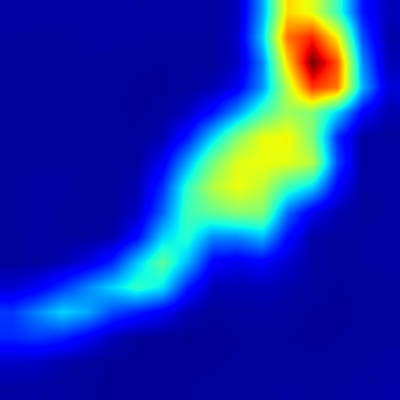}
\includegraphics[height=2.72cm]{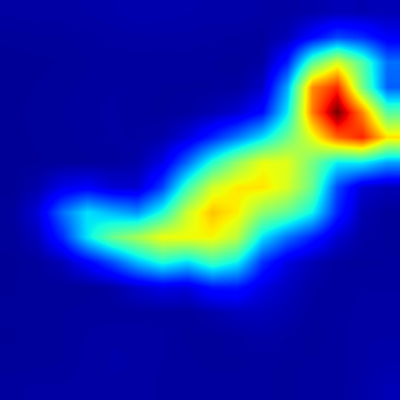}}{}
\\ 
\rotatebox{90}{\whitetxt{xx}3D grid ($A_{obj}$)} \hfill \hspace{4pt}
\jsubfig{\includegraphics[height=2.72cm]{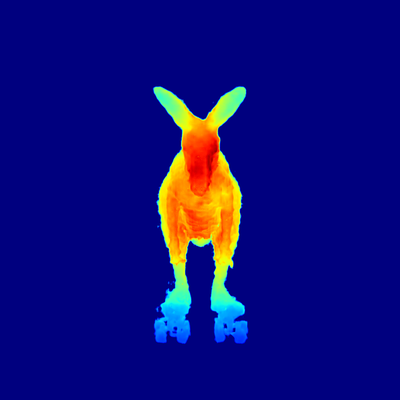}
\includegraphics[height=2.72cm]{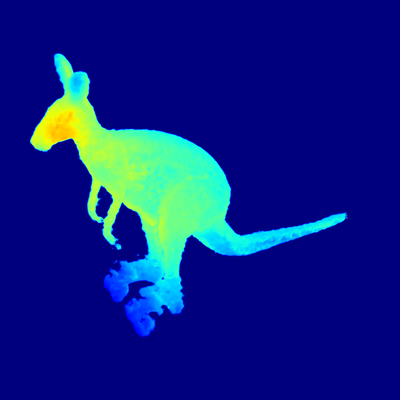}
\includegraphics[height=2.72cm]{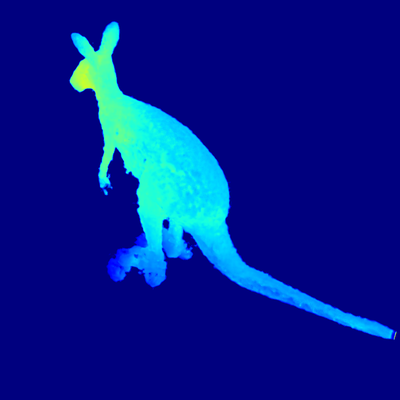}
\includegraphics[height=2.72cm]{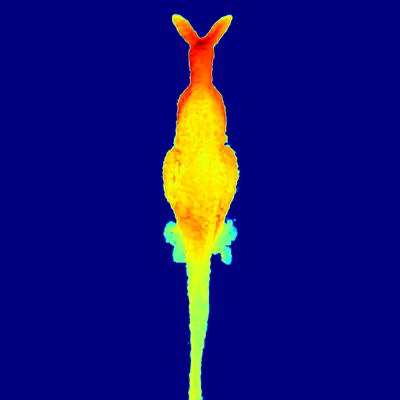}
\includegraphics[height=2.72cm]{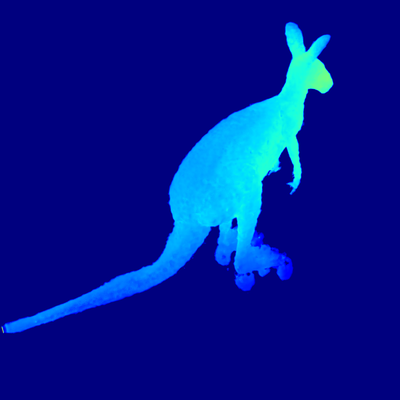}
\includegraphics[height=2.72cm]{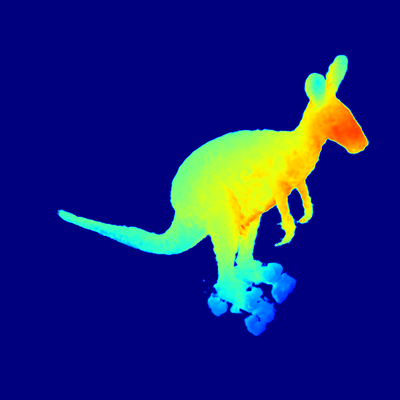}}{} \\
\rotatebox{90}{2D cross-attention} \hfill
\jsubfig{\includegraphics[height=2.72cm]{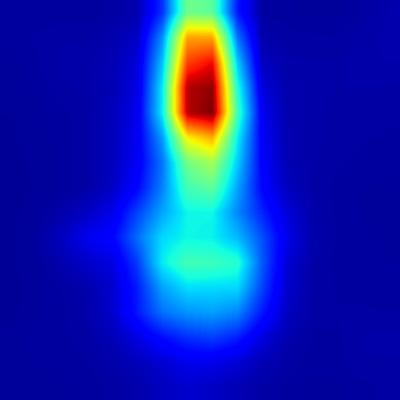}
\includegraphics[height=2.72cm]{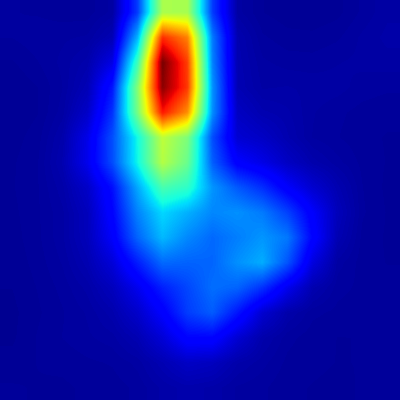}
\includegraphics[height=2.72cm]{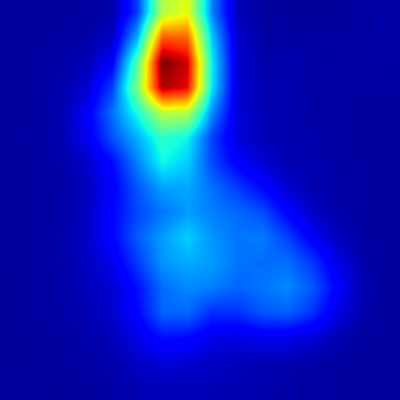}
\includegraphics[height=2.72cm]{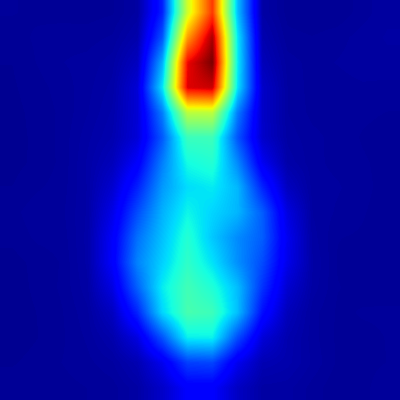}
\includegraphics[height=2.72cm]{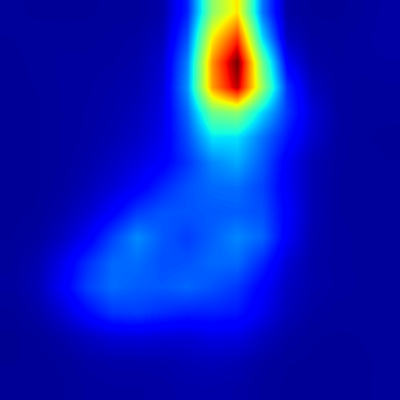}
\includegraphics[height=2.72cm]{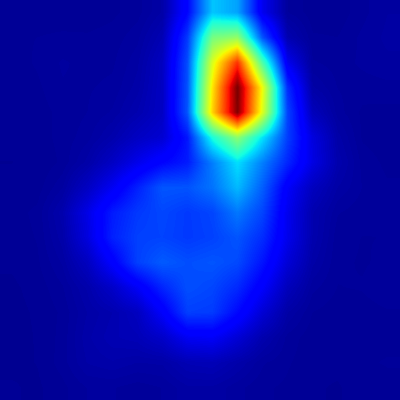}}{}
\\
\rotatebox{90}{\whitetxt{xx}3D grid ($A_e$)} \hfill
\jsubfig{\includegraphics[height=2.72cm]{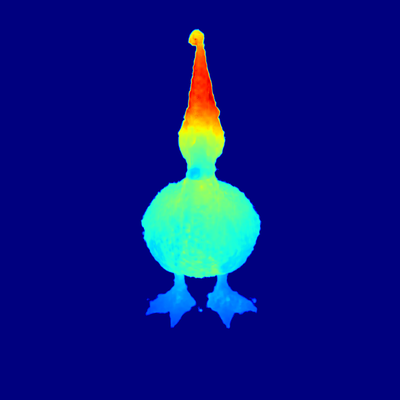}
\includegraphics[height=2.72cm]{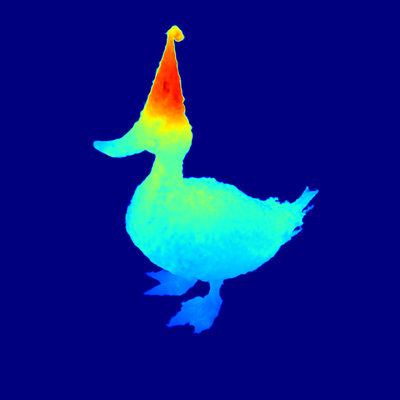}
\includegraphics[height=2.72cm]{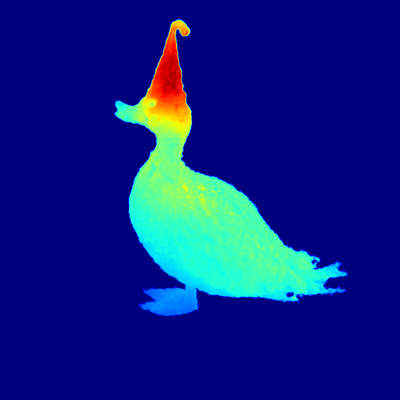}
\includegraphics[height=2.72cm]{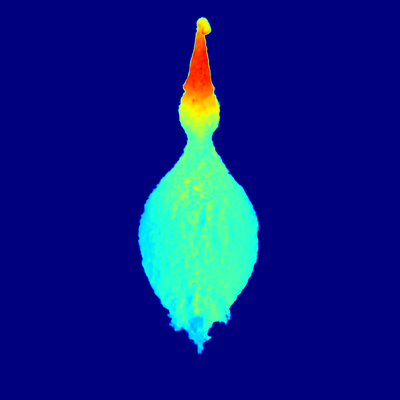}
\includegraphics[height=2.72cm]{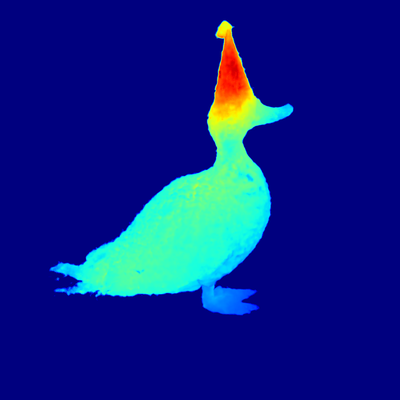}
\includegraphics[height=2.72cm]{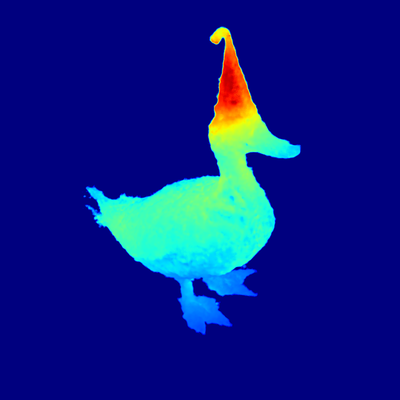}}{} 
\\ 
\rotatebox{90}{2D cross-attention} \hfill 
\jsubfig{\includegraphics[height=2.72cm]{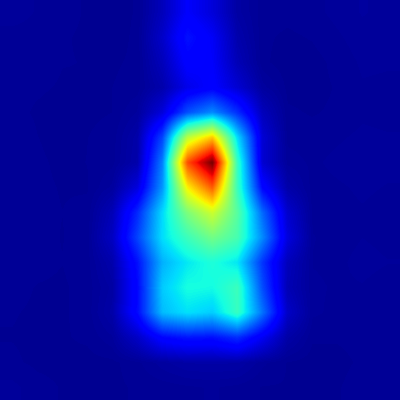}
\includegraphics[height=2.72cm]{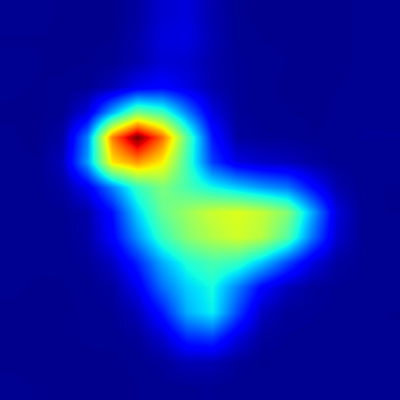}
\includegraphics[height=2.72cm]{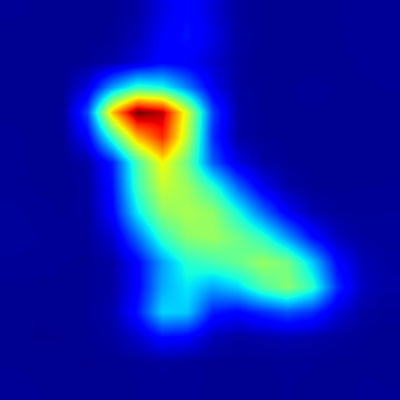}
\includegraphics[height=2.72cm]{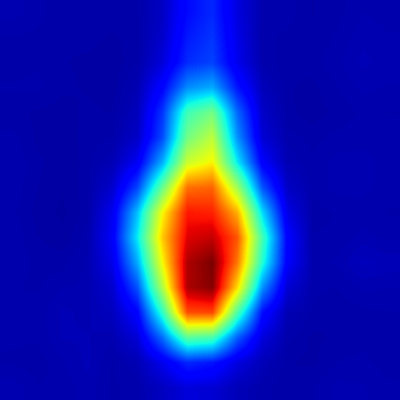}
\includegraphics[height=2.72cm]{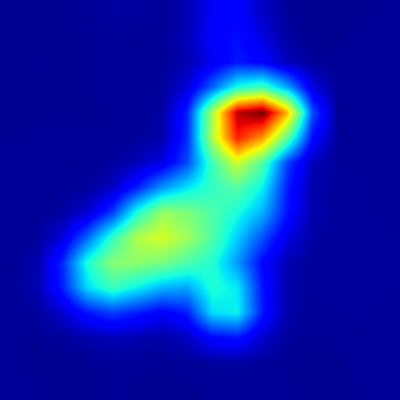}
\includegraphics[height=2.72cm]{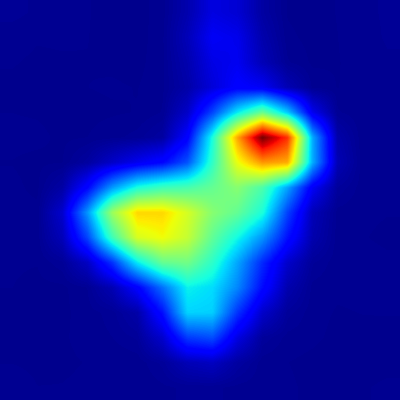}}{}
\\ 
\rotatebox{90}{\whitetxt{xx}3D grid ($A_{obj}$)} \hfill \hspace{4pt}
\jsubfig{\includegraphics[height=2.72cm]{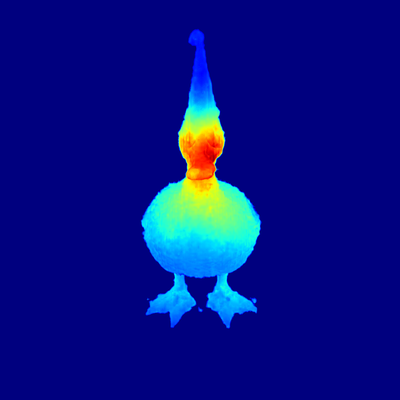}
\includegraphics[height=2.72cm]{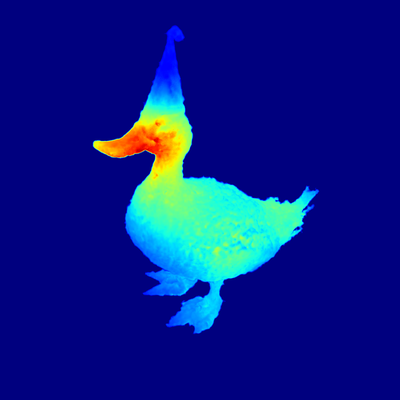}
\includegraphics[height=2.72cm]{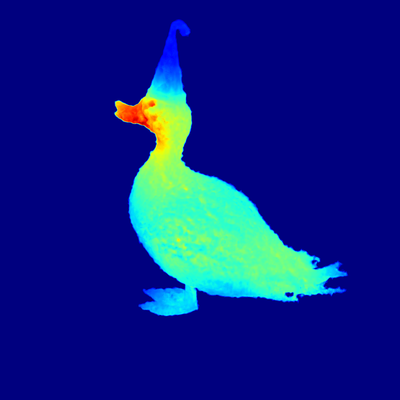}
\includegraphics[height=2.72cm]{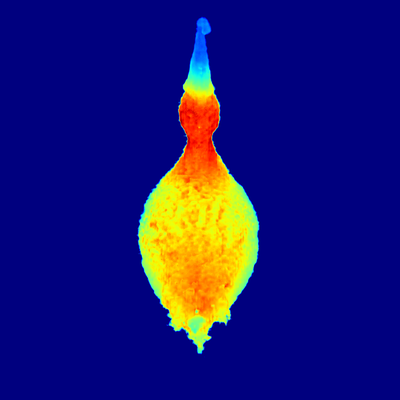}
\includegraphics[height=2.72cm]{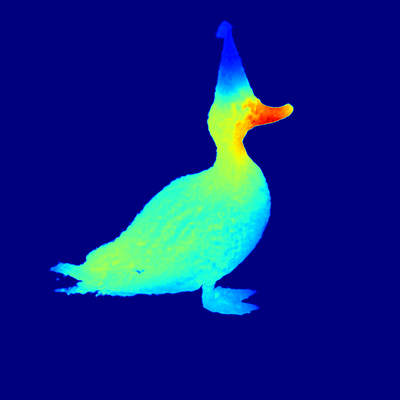}
\includegraphics[height=2.72cm]{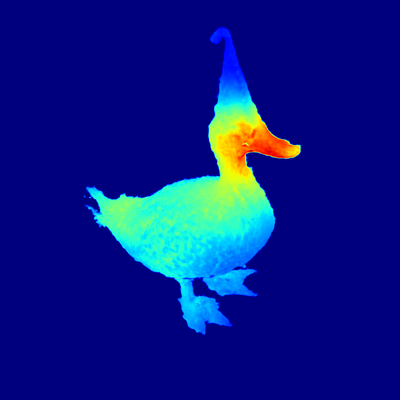}}{} 
\caption{\textbf{Visualizing 2D cross-attention maps and 3d cross-attention grids over multiple viewpoints}. We visualize the optimized 3d cross-attention grids and the corresponding 2D cross-attention maps used as supervision. We show them for the edit region corresponding to the token associated with the word ``rollerskates" (top two rows) and ``hat" (fifth and sixth rows) and the object region (third and fourth rows for the kangaroo and bottom two rows for the duck). 
}
\label{fig:supp_attn_kangaroo}
\end{figure*}

\end{document}